\documentclass[10pt,twocolumn,letterpaper]{article}

\usepackage{iccv}        
\usepackage{multirow}
\usepackage{makecell}
\usepackage{pifont}
\definecolor{iccvblue}{rgb}{0.21,0.49,0.74}
\usepackage[pagebackref,breaklinks,colorlinks,allcolors=iccvblue]{hyperref}

\def\paperID{7964} 
\def\confName{ICCV}
\def\confYear{2025}

\title{\raisebox{-0.3\height}{\includegraphics[width=1cm]{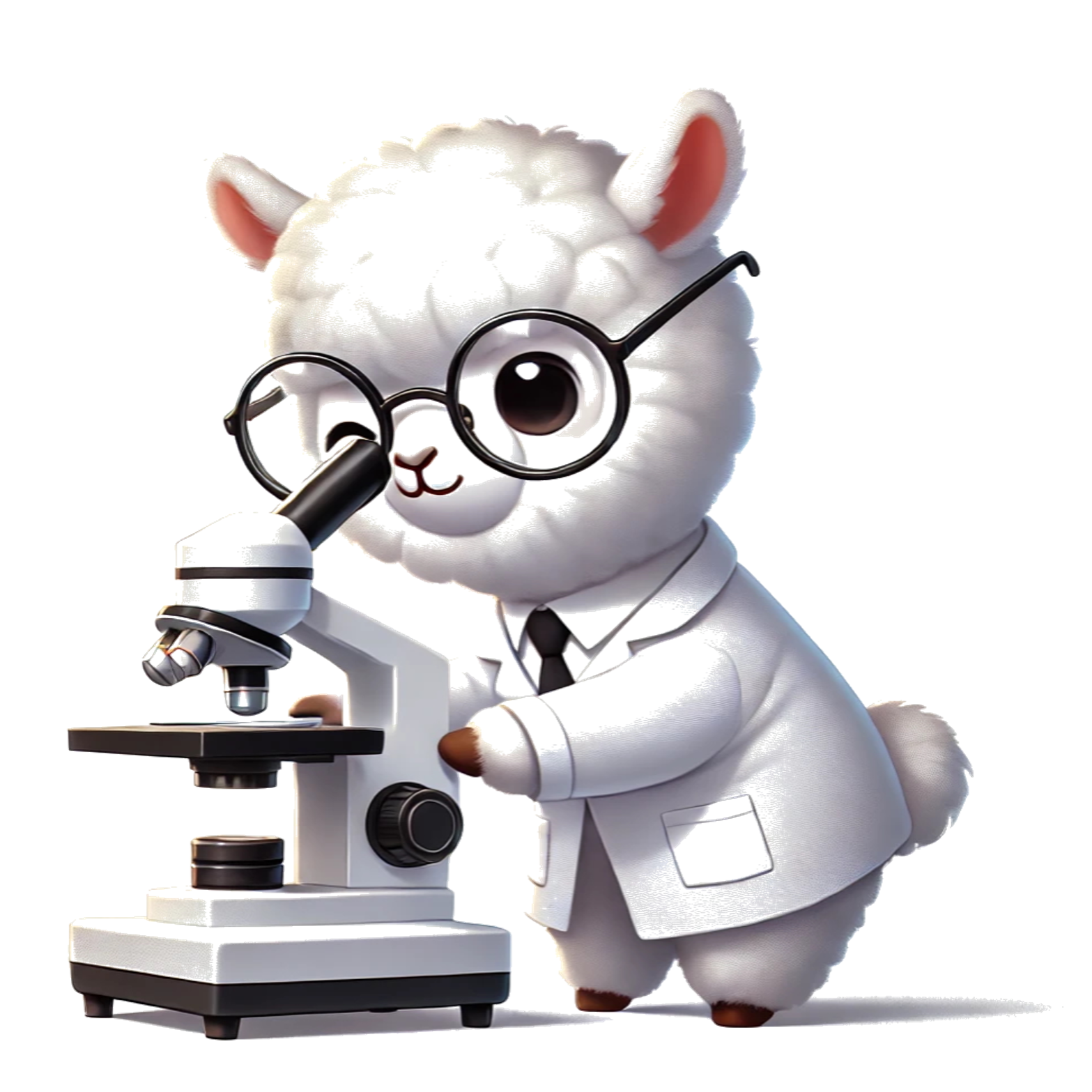}} WSI-LLaVA: A Multimodal Large Language Model for Whole Slide Image}

\author{
    Yuci Liang$^{1}$\thanks{Equal contribution.}, 
    Xinheng Lyu$^{1,2}$\footnotemark[1], 
    Wenting Chen$^{3}$\thanks{Corresponding Author.}, 
    Meidan Ding$^{1}$, 
    Jipeng Zhang$^{5}$, \\
    Xiangjian He$^{2}$, 
    Song Wu$^{1}$, 
    Xiaohan Xing$^{4}$, 
    Sen Yang$^{4}$, 
    Xiyue Wang$^{4}$, 
    Linlin Shen$^{1}$\footnotemark[2]\\
    $^{1}$Shenzhen University, 
    $^{2}$University of Nottingham Ningbo China, 
    $^{3}$City University of Hong Kong, \\
    $^{4}$Stanford University, 
    $^{5}$Hong Kong University of Science and Technology
}

\begin{document}
\maketitle
\begin{abstract}

Recent advances in computational pathology have introduced whole slide image (WSI)-level multimodal large language models (MLLMs) for automated pathological analysis. However, current WSI-level MLLMs face two critical challenges: limited explainability in their decision-making process and insufficient attention to morphological features crucial for accurate diagnosis. To address these challenges, we first introduce \textbf{WSI-Bench}, a large-scale morphology-aware benchmark containing 180k VQA pairs from 9,850 WSIs across 30 cancer types, specifically designed to evaluate MLLMs' understanding of morphological characteristics crucial for accurate diagnosis. To the best of our knowledge, WSI-Bench presents the first benchmarking systematically evaluate morphological understanding capabilities in WSI analysis. To enhance the model explainability, we present \textbf{WSI-LLaVA}, an MLLM framework for gigapixel WSI understanding with a three-stage training strategy, which can provide detailed morphological findings to explain its final answer. For more precise model assessment in pathological contexts, we develop two specialized \textbf{WSI metrics}: \textbf{WSI-Precision} and \textbf{WSI-Relevance}, focusing on clinical accuracy. Extensive evaluation on WSI-Bench reveals both the capabilities and limitations of current WSI MLLMs in morphological analysis and various pathology tasks, while demonstrating WSI-LLaVA's superior performance across all capabilities on both internal and external datasets. \href{https://github.com/XinhengLyu/WSI-LLaVA}{Source code and data} are released. 

\end{abstract}    
\section{Introduction}
\label{sec:intro}
Recent computational pathology advances have introduced both patch-level~\cite{lu2023foundational,seyfioglu2024quilt,wu2024pathinsight,xiang2025vision,Ding2024EAGLE} and whole slide image (WSI)-level~\cite{chen2024slidechat,chen2024wsicaption,guo2024histgen,Wang2024APF,chen2025wsi,lyu2025wsi} multimodal large language models (MLLMs). While patch-level models analyze localized tissue regions, WSI-level MLLMs leverage comprehensive contextual information from gigapixel images for diagnostics. WSI-level analysis is particularly valuable since histopathological diagnosis requires holistic analysis where pathologists integrate findings across extensive tissue areas~\cite{farahani2015whole,wang2023retccl,xu2024whole}. The WSI-level MLLMs fall into two categories: transformer-based models~\cite{chen2025wsi,chen2024wsicaption,guo2024histgen} for pathology report generation and simple visual question answering (VQA), and LLM-based model~\cite{chen2024slidechat} for complex VQA tasks. Despite their promising performance, these models still face significant challenges.


\begin{figure}
  \centering
  \includegraphics[width=0.95 \linewidth]{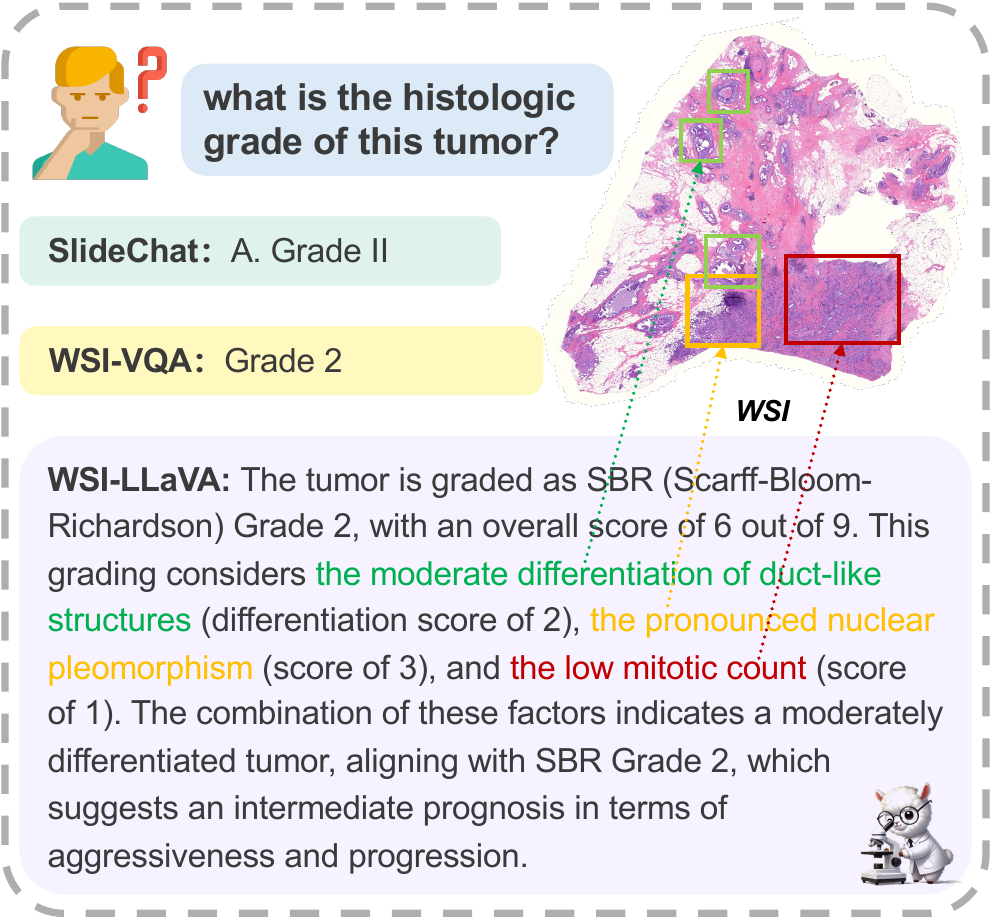}
  \caption{Comparison of WSI MLLMs. Existing frameworks offer only brief answers without explanations, while WSI-LLaVA provides comprehensive grading with detailed morphological reasoning, showing the explainability gap in WSI analysis.}
  \label{fig:different_benchmarks}
\end{figure}

The first challenge in current WSI-level MLLMs lies in their explainability: these models typically struggle to provide explanations for their final answers. Existing WSI-level models tend to offer short answers without supporting explanations. Some approaches~\cite{chen2024slidechat} are trained exclusively on closed-ended visual question answering (VQA) pairs, conditioning models to produce only final results without justification. Others~\cite{chen2025wsi} are trained on open-ended VQA pairs that primarily elicit keyword responses, as illustrated in Fig.~\ref{fig:different_benchmarks}. Unlike human pathologists who articulate detailed observations and reasoning processes, these minimal responses prevent clinicians from confidently trusting or thoroughly validating model outputs~\cite{mcduff2025towards,liu2025application,lu2023foundational}. Therefore, enhancing \textbf{\textit{explainability}} in WSI-level MLLMs is crucial for improving their clinical utility and adoption.

Another significant challenge with current pathology MLLMs is their tendency to map pathology images directly to diagnostic outcomes, overlooking crucial morphological features necessary for accurate diagnosis~\cite{lu2023foundational}. Pathologists depend on detailed observation of these morphological characteristics when interpreting pathology images. For example, distinguishing between invasive ductal carcinoma (IDC) and invasive lobular carcinoma (ILC) requires recognizing subtle differences in tumor cell arrangements~\cite{yoder2007molecular}. While some pathology MLLMs incorporate questions about morphological features, many studies~\cite{he2020pathvqa,seyfioglu2024quilt,lu2023foundational} are limited to patch-level analysis—preventing holistic assessment across the entire slide—or are trained on closed-ended VQA data that constrains their ability to produce rich, descriptive responses~\cite{chen2025wsi,chen2024slidechat}. As a result, they cannot analyze morphological features comprehensively and systematically with the detailed, nuanced descriptions characteristic of expert pathologists. Moreover, morphological observation capabilities remain unevaluated in current WSI multi-modal benchmarks~\cite{sauta2024enhancing}, which primarily assess WSI MLLMs on common pathology tasks such as diagnosis, prognosis, and report generation. Therefore, integrating detailed \textbf{\textit{morphological observations}} into WSI MLLMs is essential to better emulate human pathologists and properly evaluate their morphological understanding capabilities.



\begin{figure*}
  \centering
  \includegraphics[width=\linewidth]{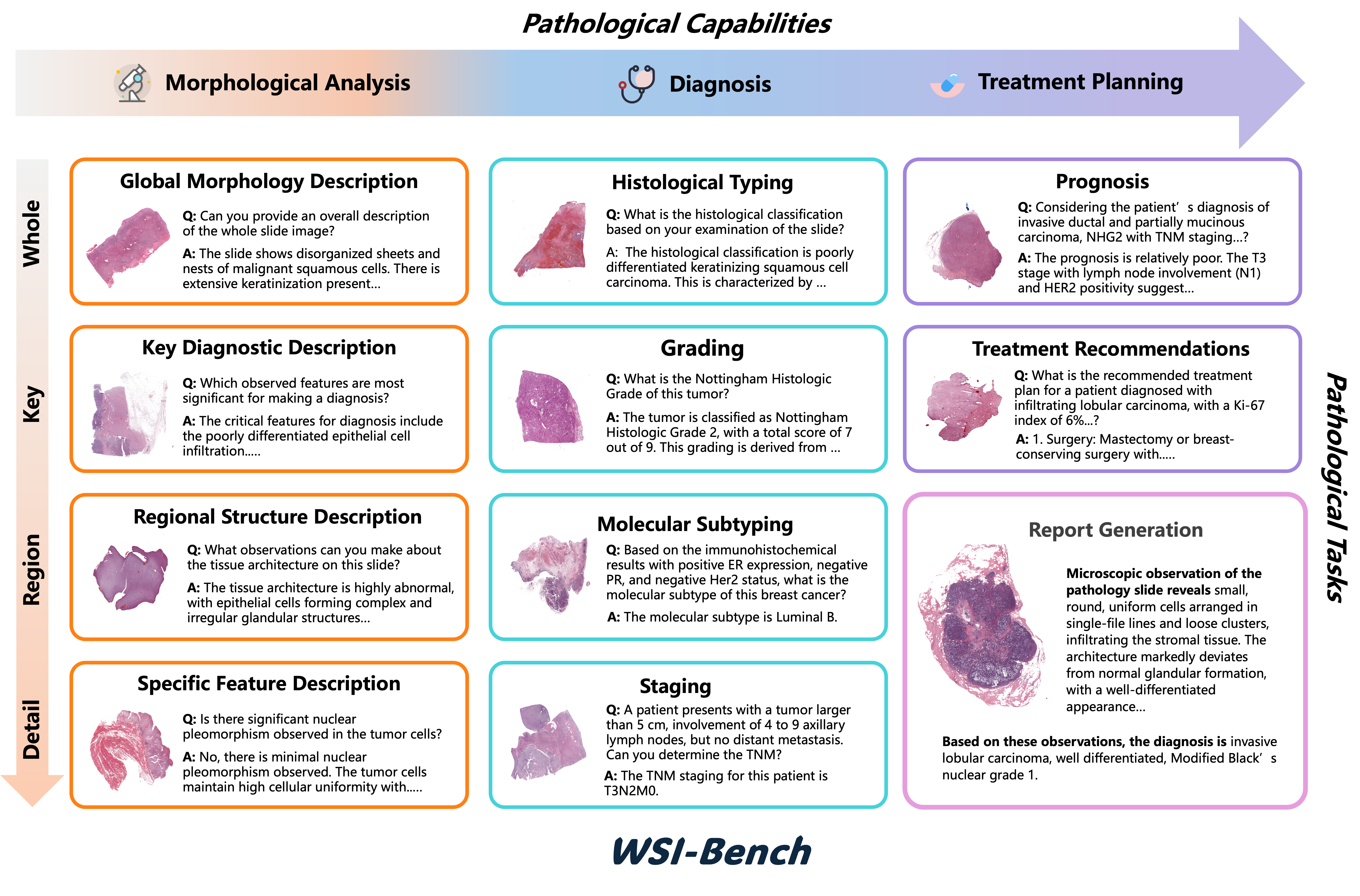}
  \vspace{-0.5cm}
  \caption{Overview of \textbf{WSI-Bench}, a morphology-aware benchmark for gigapixel WSI evaluation across 4 pathological capabilities and 11 tasks, which encompasses about 180k VQA pairs from 9,850 WSIs across 30 cancer types, sourced from 8,368 patients.}
  \label{fig:bench}
\end{figure*}

\textbf{WSI-Bench}: To address these challenges, we introduce \textbf{WSI-Bench}, a morphology-aware gigapixel WSI benchmark that systematically evaluates WSI-level MLLMs on both \textit{\textbf{morphological understanding}} and clinical pathology tasks. The benchmark assesses 4 core capabilities through 11 distinct tasks, comprising over 180k VQA pairs derived from 9,850 WSIs spanning 30 cancer types from 8,368 patients. As illustrated in Fig.~\ref{fig:bench}, the core capabilities include morphological analysis, diagnosis, treatment planning, and report generation, comprehensively covering 11 specific clinical tasks in pathology. All open-ended tasks incorporate corresponding explanations, particularly for diagnosis tasks where models must provide both conclusions and supporting morphological descriptions. This design ensures interpretability by making the reasoning process transparent, enabling pathologists to understand and validate the model's decision-making pathway. For morphological analysis, WSI-Bench includes four hierarchical morphology tasks that progress from global to local perspectives, offering a comprehensive evaluation of morphological understanding capabilities. To the best of our knowledge, this represents the first and largest morphology-aware WSI benchmark with comprehensive clinical tasks specifically designed for WSI-level MLLM training and evaluation.

\textbf{WSI-LLaVA}: To enhance \textbf{\textit{explainability}} in pathology tasks, we introduce a novel WSI-level Large Language and Vision Assistant framework (\textbf{WSI-LLaVA}) that leverages detailed {\textit{morphological observations}} to explain diagnostic reasoning. Trained on the WSI-Bench training set, WSI-LLaVA can first provide morphological findings related to the query before presenting the final answer, significantly enhancing model explainability. To mitigate the cross-modal gap between WSI and text, WSI-LLaVA employs a three-stage training approach: WSI-text alignment, feature space alignment, and task-specific instruction tuning. Our framework integrates both patch-level and slide-level encoders to capture local details and global context, providing a comprehensive representation of gigapixel WSIs for advanced pathological analysis and reasoning. 
\textbf{WSI Metrics}: Although natural language understanding (NLU) metrics are commonly used to evaluate medical language tasks, they do not accurately assess performance due to complex and often similar terminology in pathology.\cite{lopez2022natural,garcia2024gpt}. To address these limitations, we introduce two specialized WSI metrics: \textit{WSI-Precision}, which verifies the accuracy of each claim derived from the ground truth against the model's answers, and \textit{WSI-Relevance}, which assesses the alignment of each claim in the model's responses with the ground truth to ensure their relevance. These metrics are specifically designed to evaluate the quality of morphological reasoning and explanatory content in model outputs, providing a more nuanced assessment and explanation than traditional metrics allow.

Our contributions are summarized as follows:
\begin{itemize}
\item We introduce WSI-Bench, the first large-scale morphology-aware benchmark for gigapixel WSI understanding and evaluation, encompassing 180k VQA pairs from 9,850 WSIs across 30 cancer types. This benchmark uniquely evaluates the morphological understanding capability and incorporates morphological descriptions as supporting evidence for each task.

\item To enhance model explainability, we propose WSI-LLaVA, a novel framework for gigapixel WSI analysis and explanation that integrates patch-level and slide-level features through a three-stage training approach, which leverages detailed morphological observations to explain diagnostic reasoning.

\item We develop WSI-specific evaluation metrics that provide a more accurate assessment and explanation of model performance in pathological contexts, which address the limitations of traditional NLU metrics by verifying claim accuracy and response relevance.
%
\item We conduct a comprehensive evaluation of current WSI MLLMs and other MLLMs, shedding light on their strengths and weaknesses in WSI analysis, and prove the effectiveness of our WSI-LLaVA.
\end{itemize}
\section{Related Work}
\label{sec:related_work}

\paragraph{Pathology VQA Benchmarks.}
Prior pathology VQA and report generation benchmarks have advanced the field but exhibit notable limitations. Patch-level benchmarks like PathVQA ~\cite{he2020pathvqa} and Quilt-Instruct~\cite{seyfioglu2024quilt} contribute valuable educational content but remain confined to patch analysis with multiple-choice formats, missing the holistic view required clinically. WSI-VQA ~\cite{chen2025wsi} attempts slide-level analysis but uses simplistic formats with quality issues affecting clinical relevance. For report generation, WSICaption ~\cite{chen2024wsicaption} pioneers WSI-text pair curation from TCGA data, which is later refined by HistGen~\cite{guo2024histgen}. While SlideBench-VQA ~\cite{chen2024slidechat} shows progress, it covers only 10 cancer types and lacks complex dialogues for thorough evaluation. Crucially, these benchmarks have not systematically evaluated morphological understanding capabilities essential for diagnosis. To address this issue, we introduce WSI-Bench, which curates WSI-level VQA pairs focused on morphological analysis, diagnostic reasoning, treatment planning, and report generation—enabling comprehensive assessment.

\paragraph{WSI-Level MLLMs.}
Recent advances in WSI-level multi-modal models demonstrate efforts to tackle the gigapixel nature of whole slide images. Transformer-based architectures like WSI-Caption~\cite{chen2024wsicaption}, HistGen~\cite{guo2024histgen}, WSI-VQA~\cite{chen2025wsi}, and PRISM~\cite{shaikovski2024prism} have made progress in report generation and basic slide understanding, but exhibit limited interaction capabilities beyond their designed tasks. Most existing models are limited to handling multiple-choice questions or producing extremely brief responses of only two or three words. SlideChat~\cite{chen2024slidechat} represents the first attempt to integrate slide-level features with large language models, but lacks sufficient morphological reasoning capabilities and comprehensive output generation. Furthermore, existing models have been predominantly trained and evaluated in limited types of cancer, restricting their clinical applicability to various pathological conditions. WSI-LLaVA framework addresses these limitations through a systematic three-stage training approach that enables comprehensive gigapixel WSI analysis with enhanced explainability. 

\section{WSI-Bench}

In clinical practice, pathologists rely on morphological features for diagnosis, but current WSI MLLMs overlook these crucial details, compromising precision \cite{bradbury2022have}. Thus, we introduce a morphology-aware WSI-Bench benchmark for gigapixel WSI evaluation to include morphological observations for a complete understanding of WSI. It can evaluate the WSI MLLMs across 4 pathological capabilities with 11 tasks for WSI analysis. As shown in Fig.~\ref{fig:bench}, the pathological capabilities include \textit{morphological descriptions}, \textit{diagnosis}, and \textit{treatment planning}, and \textit{report generation}.

\subsection{Pathological Capabilities}
\label{sec:path_cap}
\noindent\textbf{Morphological Analysis.}
To assess the morphological analysis capability of WSI MLLMs, the four hierarchical pathological tasks are introduced for a comprehensive evaluation, ranging from global to local perspectives. These tasks include global morphology description, key diagnostic description, regional structure description, and specific feature description. Each category is designed to capture distinct pathological aspects, enhancing the dataset's comprehensiveness while providing clinically relevant insights.

\noindent\textbf{1) Global Morphology Description} evaluates the ability to provide a comprehensive overview of WSIs, establishing a context for global analysis. We construct the open-ended VQA pairs by a question with fixed content and a global morphological description. The question asks for the overall description of the WSI. The global morphological description is extracted from our WSI-report pairs, which should capture the general tissue architecture and overall tumor presentation, ensuring a broad understanding of the slide’s features.

\noindent\textbf{2) Key Diagnostic Description} focuses on the evaluation of crucial morphological features vital for accurate diagnosis, linking specific observations to their clinical implications. Each open-ended VQA pair includes a question about key morphological features for diagnosis and its answer. We construct the answer by extracting content related to the cancer type's diagnosis from the WSI report.

\noindent\textbf{3) Regional Structure Description} aims to evaluate the ability of broad observations in specific regions, such as \textit{tumor infiltration},\textit{tissue architecture}, and \textit{cellular morphology}. We construct both open-ended and close-ended (multiple-choice questions, MCQs) VQA pairs, which include a question about a specific region and its answer. The answer is obtained by extracting relevant descriptions for the specific region from the WSI report.

\noindent\textbf{4) Specific Feature Description} evaluates detailed morphological examinations of specific aspects (vascular invasion, gland formation patterns, cellular pleomorphism, etc.). We create both open-ended and close-ended (MCQs) VQA pairs about specific features, with answers indicating their presence in WSI reports.

\noindent\textbf{Diagnosis.} To evaluate diagnostic capabilities, we examine four fundamental tasks: \textit{histological typing}, \textit{grading}, \textit{molecular subtyping}, and \textit{staging}. For interpretable and accurate diagnosis, each result includes both the answer and explanatory morphological analysis.

\noindent\textbf{1) Histological Typing} evaluates the ability of histological type classification. It includes both open-ended and close-ended (MCQs) VQA pairs with a question and its answer, where the answer is obtained from the WSI report.

\noindent\textbf{2) Grading} assesses the ability to classify the grade with both open-ended and close-ended (MCQs) VQA pairs, where each VQA pair includes a question about grading and its answer obtained from the WSI report.

\noindent\textbf{3) Molecular Subtyping} evaluates identification of specific subtypes based on histological typing and immunohistochemical data. We create open-ended VQA pairs with questions containing histological typing and immunohistochemical data, with answers from original data labels.

\noindent\textbf{4) Staging} assesses prediction of Tumor-Node-Metastasis (TNM)~\cite{scorsetti2016thymoma} labels based on clinical scenarios. Both open-ended and close-ended (MCQs) VQA pairs present clinical information translated from TNM classifications, requiring inference of TNM classification from the provided context.
\begin{figure*}
  \centering
  \includegraphics[width=\linewidth]{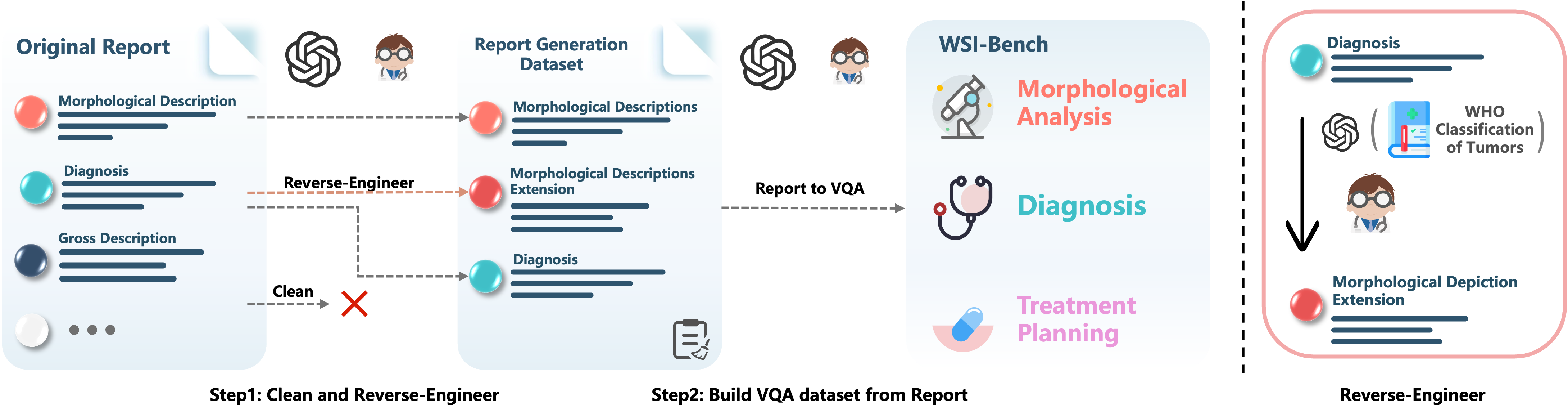}
  \caption{Two-step process for constructing WSI-Bench: Step 1 removes gross descriptions and immunohistochemical results from pathology reports, retaining morphological descriptions and diagnostic results, with enriched descriptions generated via reverse-engineering; Step 2 builds VQA pairs from the refined reports to support pathological tasks.}
  \label{fig_build_data}
\end{figure*}

\noindent\textbf{Treatment Planing.}
In the context of treatment planning, WSI-Bench includes three pathological tasks, including \textit{prognosis}, \textit{treatment recommendations}, and \textit{report generation}. To imitate real-world settings, we include both WSI and its comprehensive information like diagnostic and immunohistochemical results for treatment planning.

\noindent\textbf{1) Treatment recommendations} evaluate treatment suggestion capabilities. We create open-ended VQA pairs with questions including diagnostic results (histological typing, grading, staging) and immunohistochemical data. GPT-4o generates initial recommendations based on WSI reports, refined by clinical experts.

\noindent\textbf{2) Prognosis} evaluates prediction of disease outcomes (good or bad). We create Yes/No VQA pairs with questions containing diagnostic and immunohistochemical data. GPT-4o generates initial prognosis results, refined by clinical experts.


\noindent\textbf{Report Generation.}
 Report generation assesses the ability to create reports from input WSIs. Ground-truth reports are extracted from original pathology reports using GPT-4o. We exclude content not observable in WSIs, retaining only morphological descriptions and diagnostic results (histological typing and grading).

\subsection{Data Construction}
We curate WSI-Bench from The Cancer Genome Atlas (TCGA)\cite{weinstein2013cancer}, including 10,756 WSIs across 30 cancer types. After excluding cases with multiple reports per WSI, the final WSI-Bench contains 9,850 WSIs paired with their corresponding pathology reports. The specific distribution across different cancer types can be found in Appendix Section~\ref{appendix:wsi_bench}. In Fig.~\ref{fig_build_data}, our data preparation involved two main steps: (1) using GPT-4o to extract and clean text from original reports, removing gross descriptions and immunohistochemical results while retaining morphological descriptions and diagnostic results, and (2) implementing a \textbf{reverse-engineer scheme} to enrich morphological descriptions based on diagnostic labels and established standards like the Nottingham grading system~\cite{dilani2019classification}. All generated content was reviewed by clinical experts. Based on these finalized WSI reports, we construct the VQA pairs required for each pathological task, as detailed in Section~\ref{sec:path_cap}. 

\noindent\textbf{Data Statistics and Split.}
In Table~\ref{tab_statistic}, WSI-Bench comprises 179,569 VQA pairs. The dataset includes 175,450 VQA pairs (122,133 open-ended and 53,317 closed-ended) in the training set with 9,642 WSIs, plus 4,119 VQA pairs (2,838 open-ended and 1,281 closed-ended) in the testing set with 208 WSIs. Details on construction process, quality control measures, hallucination prevention methods, and data distribution are in Appendix Section~\ref{appendix:wsi_bench}.



\begin{table}[tp]
    \centering
    \setlength\tabcolsep{0.5pt}
    \scalebox{0.83}{
    \begin{tabular}{llc}
        \toprule
        \textbf{Capabilities} & \textbf{Tasks} & \textbf{VQA Pair Count} \\ 
        \midrule
        \multirow{4}{*}{\makecell{Morphological \\Analysis}} 
            & Global Morphology Description & 9,850 \\ 
            & Key Diagnostic Description & 9,850 \\ 
            & Regional Structure Description & 55,181 \\ 
            & Specific Feature Description & 48,170 \\ 
        \cmidrule{1-3}
        \multirow{4}{*}{Diagnosis} 
            & Histological Typing & 19,697 \\ 
            & Grading & 15,335 \\ 
            & Molecular Subtyping & 977 \\ 
            & Staging & 6,743 \\ 
        \cmidrule{1-3}
        \multirow{2}{*}{\makecell{Treatment\\ Planning}} 
            & Treatment Recommendations & 1,958 \\ 
            & Prognosis & 1,958 \\ 
        \cmidrule{1-3}
        \multicolumn{2}{l}{Report Generation} & 9,850 \\ 
        \midrule 
        \textbf{Total} & & \textbf{179,569 (180k)} \\ 
        \bottomrule
    \end{tabular}}
    \caption{Statistics Analysis on the WSI-Bench.}
    \label{tab_statistic}
\end{table}

\subsection{WSI Metrics}
Although NLU metrics are widely used to evaluate open-ended language tasks, they cannot accurately assess performance in pathology due to its complex and often similar terminology, and the scores themselves lack interpretability~\cite{he2023medeval,santos2022pathologybert}. Thus, we introduce two WSI metrics for open-ended questions: \textbf{\textit{WSI-Precision}} and \textbf{\textit{WSI-Relevance}}.The detailed evaluation workflow for these metrics can be found in Appendix Section
Fig.~\ref{fig:wsi_metrics}

\noindent\textbf{WSI-Precision} measures the accuracy of model responses by comparing them directly to claims derived from the ground-truth (GT) response. It involves three main steps. 1) \textit{Claims Extraction}: We extract distinct claims from the GT response using GPT-4o to break down complex medical information into assessable units. 2) \textit{Individual Score Calculation and Interpretation}: We compare each extracted claim with the generated response using our scoring criterion. We assign Score 1 for complete accuracy with perfect alignment to established facts; Score 0.7 for mostly correct information with minor discrepancies that do not alter clinical significance; Score 0.3 for significant errors in critical content areas affecting the core message; and Score 0 for completely incorrect or unrelated information. Each score is accompanied by a corresponding explanation that justifies the reasoning behind the score. 3) \textit{Final Score Calculation}: We calculate the overall WSI-Precision score by averaging all individual claim scores.

\begin{figure*}
  \centering
  \includegraphics[width=0.8\linewidth]{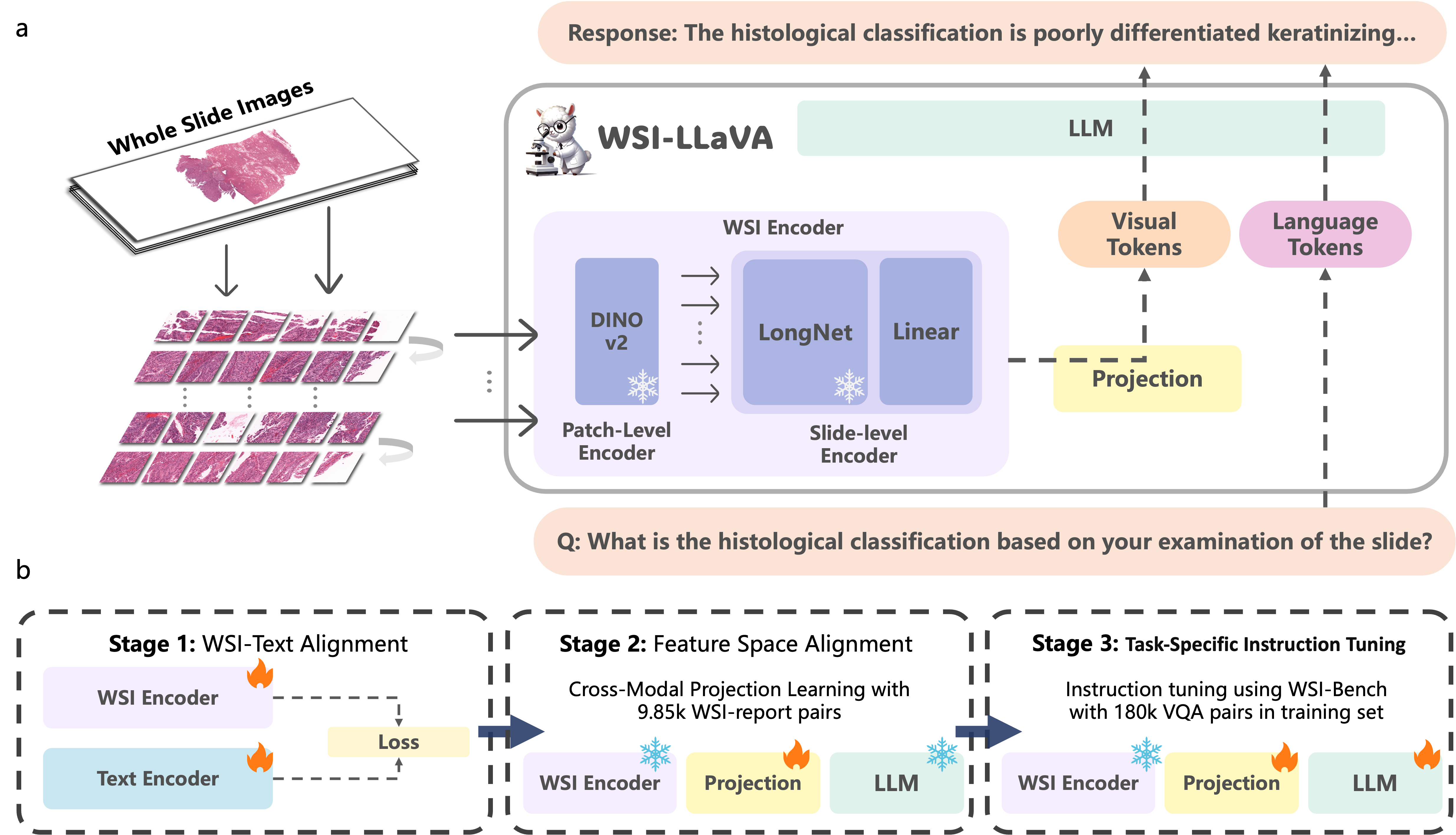}
  \caption{Overview of our gigapixel \textbf{WSI-LLaVA} framework with a three-stage training approach for cross-modal alignment and pathological tasks learning.}
  \label{fig:wsi_llava}
\end{figure*}

    
    

\noindent\textbf{WSI-Relevance} measures the alignment of each claim in the model's responses with the ground truth to ensure their relevance. This metric uses a similar three-step process. The detailed step-by-step procedures, prompt designs, and implementation specifics for both WSI-Precision and WSI-Relevance metrics are provided in Appendix Section~\ref{appendix:wsi_metric}.


 \paragraph{Reliability Validation.} To evaluate the reliability of our proposed WSI metrics, we randomly sample 2\% of open-ended responses for assessment by five medical experts. Results show strong alignment between automated metrics and human evaluation (Pearson correlation: 0.87 for WSI-Precision, 0.82 for WSI-Relevance). Further details are provided in Appendix Section~\ref{appendix:wsi_metrics_validation}.

\section{WSI-LLaVA}
To enhance the explainability, we introduce a \textbf{WSI-LLaVA} framework with a three-stage training approach for gigapixel WSI analysis (Fig.~\ref{fig:wsi_llava}). Trained on the training set of WSI-Bench, it can first provide morphological findings related to the query before presenting the final answer, significantly enhancing model explainability. To mitigate the cross-modal gap, the three-stage training approach includes a WSI-text alignment, feature space alignment, and task-specific instruction tuning. In the first stage, we integrate a patch-level encoder using Prov-GigaPath\cite{xu2024whole}'s tile-level encoder (pre-trained with DINOv2\cite{oquab2023dinov2}) to process $256 \times 256$ pixel patches, and a slide-level encoder based on LongNet\cite{ding2023longnet} to generate global features. These are aligned with textual features encoded by Bio\_ClinicalBERT\cite{alsentzer2019publicly} through contrastive learning on WSI-report pairs. In the second stage, we combine the pre-trained WSI encoder with a projection layer and LLM, freezing the pre-trained models while training only the projection layer on WSI-report pairs. In the last stage, we freeze WSI encoder while training the projection layer and LLM on all tasks in WSI-Bench. Please see Appendix Section~\ref{appendix:implement} for more details. 
\begin{table*}[ht]
  \centering
    \setlength\tabcolsep{3.6pt}
  \scalebox{0.85}{
  \begin{tabular}{l c c c c c c c c c c c}
    \toprule
    \multicolumn{2}{c}{\textbf{}} & \multicolumn{3}{c}{\textbf{Morphological Analysis}} & \multicolumn{3}{c}{\textbf{Diagnosis}} & \multicolumn{3}{c}{\textbf{Treatment Planning}} & \multirow{3}{*}{\textbf{Average}}\\
    \cmidrule(lr){3-5} \cmidrule(lr){6-8} \cmidrule(lr){9-11}
    Model & Input & \multicolumn{2}{c}{Open} & Close & \multicolumn{2}{c}{Open} & Close & \multicolumn{2}{c}{Open} & Close &\\
    & & WSI-P & WSI-R & Acc & WSI-P & WSI-R & Acc & WSI-P & WSI-R & Acc &\\
    \midrule
    Quilt-LLaVA\cite{seyfioglu2024quilt} & Thumbnail & 0.448 & 0.447 & 0.947& 0.586 & 0.604 &0.849& 0.788 & 0.816 & 1.000 & 0.721\\
    GPT-4o\cite{islam2024gpt} & Thumbnail & 0.220 & 0.204 &0.471& 0.472 & 0.457 &0.530& 0.496 & 0.841 & 0.875 & 0.507\\
    WSI-VQA\cite{chen2025wsi} & WSI & 0.395 & 0.462 &0.758& 0.436 & 0.525 & 0.577& 0.791 & 0.595 &0.771 &0.590\\
    \textbf{WSI-LLaVA} & WSI & \textcolor{blue}{0.488} & \textcolor{blue}{0.610} & \textcolor{blue}{0.951} & \textcolor{blue}{0.610} & \textcolor{blue}{0.612} & \textcolor{blue}{0.863} & \textcolor{blue}{0.810} & \textcolor{blue}{0.845} & \textcolor{blue}{1.000} & \textcolor{blue}{0.754}\\
    \bottomrule
  \end{tabular}
  }
    \caption{Quantitative evaluation of WSI and general-purpose MLLMs on our WSI-Bench across three pathological capabilities. WSI-P: WSI-Precision, WSI-R: WSI-Relevance, Acc: accuracy, open: open-ended question, and close: close-ended question.}
  \label{tab:average_performance}
\end{table*}

\begin{table*}[h!]
    \setlength\tabcolsep{3.9pt}
    \centering
    \scalebox{0.85}{
    \begin{tabular}{lcccccccc}
        \toprule
        Models & BLEU-1 & BLEU-2 & BLEU-3 & BLEU-4 & ROUGE-L & METEOR & WSI-Precision & WSI-Relevance \\ 
        \midrule
        GPT-4o\cite{islam2024gpt} &  0.202&0.069 &0.030 &0.016 &0.132 & 0.167& 0.067 &0.138 \\ 
        Quilt-LLaVA\cite{seyfioglu2024quilt} & 0.474 & 0.351 & 0.282 & 0.236 & 0.475 & 0.460 & 0.324 & 0.333 \\ 
        MI-Gen\cite{chen2024wsicaption} & 0.403 & 0.306 & 0.248 & 0.209 & 0.446 & 0.407 & 0.310 & 0.377 \\ 
        Hist-Gen\cite{guo2024histgen} & 0.406 & 0.307 & 0.248& 0.208 & 0.448 & 0.416 & 0.300 & 0.397 \\ 
        \textbf{WSI-LLaVA} & \textcolor{blue}{0.480} & \textcolor{blue}{0.358} & \textcolor{blue}{0.287} & \textcolor{blue}{0.240} & \textcolor{blue}{0.490} & \textcolor{blue}{0.465} &\textcolor{blue}{0.380} & \textcolor{blue}{0.429}\\ 
        \bottomrule
    \end{tabular}
    }
    \caption{Quantitative evaluation of WSI and general-purpose MLLMs in report generation task on WSI-Bench.}
    \label{tab:report_gen}
\end{table*}

\section{Experiment}

\subsection{Quantitative Evaluation}
We evaluate various WSI MLLMs on WSI-Bench dataset, including specialized models for WSI report generation (MI-Gen~\cite{chen2024wsicaption} and Hist-Gen~\cite{guo2024histgen}) and pathological VQA tasks (Quilt-LLaVA~\cite{seyfioglu2024quilt} and WSI-VQA~\cite{chen2025wsi}), and the general-purpose MLLM like GPT-4o. For models with input size constraints, we resize the WSIs to $1024 \times 1024$ pixels. All WSI MLLMs are trained on WSI-Bench's training set and evaluated on its test set. For quantitative evaluation, Table~\ref{tab:average_performance} shows that our WSI-LLaVA achieves the highest average score of 0.754, indicating its superiority over both WSI-based models and general-purpose MLLMs. Moreover, the performance of closed-ended questions for different models is better than that of open-ended questions, suggesting that the open-ended questions are more challenging for current models.

For the report generation task, Table~\ref{tab:report_gen} shows that WSI-LLaVA achieves superior performance across all metrics, leading in both NLU measures (BLEU-1: 0.480, ROUGE-L: 0.490) and our domain-specific metrics (WSI-Precision: 0.380, WSI-Relevance: 0.429), indicating its superiority over general-purpose and specialized models.

\begin{table}[h!]
\centering
\scriptsize
\renewcommand{\arraystretch}{1.5}  
\scalebox{0.8}{
\begin{tabular}{>{\arraybackslash}m{1.3cm} m{6.4cm}} 
\toprule[1.5pt]
\textbf{Image} & \includegraphics[width=2cm]{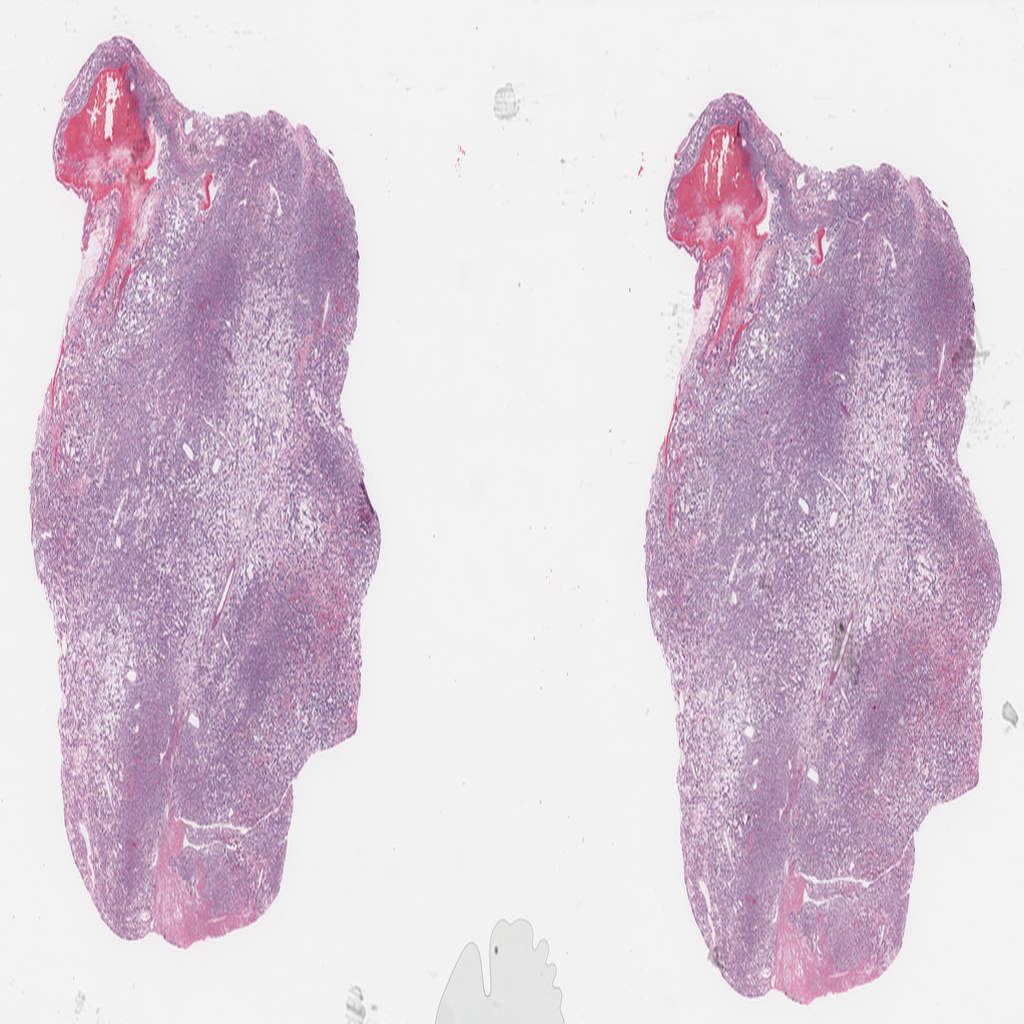} \\
\midrule
\textbf{Question} &Based on the image,what is the histological classification? \\
\midrule
\textbf{Quilt-LLaVA} & The histological classification is \textcolor{red!50!black}{renal cell carcinoma, conventional (clear cell) type}. This classification is supported by \textcolor{green!50!black}{the presence of clear cells} with \textcolor{red!50!black}{distinct borders and abundant cytoplasm, arranged in nests and sheets, along with the noted nuclear features.} \\
\midrule
\textbf{WSI-VQA} & The histological classification based on the observed features is \textcolor{green!50!black}{pheochromocytoma} . This classification is supported by \textcolor{green!50!black}{the presence of polygonal cells} with \textcolor{red!50!black}{clear cytoplasm and distinct cell borders} along with the \textcolor{green!50!black}{presence of sustentacular cells}. \\
\midrule
\textbf{WSI-LLaVA} & The histological classification is \textcolor{green!50!black}{pheochromocytoma}. This is supported by \textcolor{green!50!black}{the presence of large polygonal cells with abundant granular cytoplasm and round to oval nuclei}, along with \textcolor{green!50!black}{the characteristic zellballen pattern} and vascular stroma. \\
\midrule
\textbf{Ground Truth} & The histological classification is pheochromocytoma. This is determined by the presence of the zellballen pattern and typical cytological features such as eosinophilic cytoplasm and finely stippled chromatin within the tumor cells. \\ 
\bottomrule[1.5pt]
\end{tabular}}
\caption{Sample Comparison of VQA. Correct responses are highlighted in green, and incorrect responses are highlighted in red.}
\label{tab:qual_vqa}
\end{table}

\subsection{Qualitative Evaluation}
In Table~\ref{tab:qual_vqa}, all models trained with WSI-Bench gain diagnostic explanation capabilities. Quilt-LLaVA incorrectly diagnoses renal cell carcinoma due to misidentified morphology. WSI-VQA and WSI-LLaVA correctly identify pheochromocytoma, with WSI-LLaVA providing more accurate supporting features (e.g., Zellballen pattern) than WSI-VQA, demonstrating WSI-LLaVA's superior ability to identify key diagnostic features.



\begin{table*}[ht]
\footnotesize
\centering
\scalebox{0.87}{
\begin{tabular}{p{1.1cm} p{12.2cm} p{1cm} p{1cm} p{1.8cm}}
\toprule
\multirow{2}{*}{\textbf{Model}} & \multirow{2}{*}{\textbf{Model Responses}} & \multicolumn{2}{c}{\textbf{WSI Metrics}}  & \multirow{2}{*}{\textbf{NLU Metrics}} \\ 
\cmidrule(lr){3-4}
 &  & \textbf{WSI-P} & \textbf{WSI-R} &  \\
\midrule
\textbf{Ground Truth} & \underline{Microscopic observation of the pathology slide reveals} \textcolor{green!50!black}{poorly differentiated squamous cell carcinoma extensive invasion with minimal keratinization and significant cellular atypia}. The tumor displays \textcolor{green!50!black}{extensive necrosis } \textcolor{orange!80!black}{alongside regions of calcification within the tissue}. Additionally, there is evidence of \textcolor{green!50!black}{tumor infiltration into surrounding structures, with observable lymphovascular invasion}. \underline{Based on these observations, the diagnosis is } \textcolor{green!50!black}{squamous cell carcinoma}, \textcolor{orange!80!black}{Grade 3 (of 3)}.& N/A & N/A & N/A \\
\midrule
\textbf{WSI-LLaVA} &  \underline{Microscopic observation of the pathology slide reveals} \textcolor{green!50!black}{poorly differentiated squamous cell carcinoma} characterized by \ldots\textcolor{green!50!black}{with minimal keratinization.} The tumor cells exhibit \textcolor{green!50!black}{marked nuclear pleomorphism, with enlarged, irregular nuclei and prominent nucleoli.} \ldots \textcolor{green!50!black}{Vascular invasion is evident, with tumor cells infiltrating lymphovascular spaces. Areas of necrosis are present within the tumor mass.} \underline{Based on these observations, the diagnosis is }  \textcolor{green!50!black}{poorly differentiated invasive squamous carcinoma.} & \textcolor{blue}{0.800} & \textcolor{blue}{0.830} & \scriptsize{BLEU-1: 0.521 \newline
BLEU-2: 0.389 \newline
\textcolor{blue}{BLEU-3: 0.326} \newline
\textcolor{blue}{BLEU-4: 0.283} \newline
ROUGE-L: 0.534
\newline
\textcolor{blue}{METEOR: 0.617} }\\
\midrule
\textbf{Hist-Gen} & \underline{Microscopic observation of the pathology slide reveals} \textcolor{green!50!black}{a poorly differentiated squamous cell carcinoma}
characterized by \ldots 
\textcolor{green!50!black}{atypical squamous cells.} The tumor cells exhibit marked nuclear pleomorphism \textcolor{red!50!black}{with prominent nucleoli and prominent nucleoli. \underline{There is no evidence of} vascular invasion or necrosis within the examined tissue.}  \underline{Based on these observations, the diagnosis is }\textcolor{green!50!black}{squamous cell carcinoma poorly differentiated.} & 0.233 & 0.520 & 
\scriptsize{
\textcolor{blue}{BLEU-1: 0.531} \newline
\textcolor{blue}{BLEU-2: 0.402} \newline
BLEU-3: 0.316 \newline
BLEU-4: 0.259 \newline
\textcolor{blue}{ROUGE-L: 0.536} \newline METEOR: 0.560} \\
\bottomrule
\end{tabular}}
\caption{Comparison between WSI metrics and NLU metrics. Green: matched ground-truth content; Red: deviations; Orange: ground truth content missing in model response; Underlined: template language. WSI-P: Precision, WSI-R: Relevance.}
\label{tab:wsi_vs_nlp}
\end{table*}

\begin{table*}[ht]
  \centering
  \setlength\tabcolsep{3pt}
  \scalebox{0.8}{
  \begin{tabular}{l c c c c c c c}
    \toprule
    \multirow{2}{*}{Model} & \multirow{2}{*}{Input} & \multirow{2}{*}{\textbf{WSI-VQA Dataset}} & \multicolumn{4}{c}{\textbf{SlideBench-VQA (BCNB)}} & \multirow{2}{*}{\textbf{CPTAC-NSCLC}} \\
    \cmidrule(lr){4-7}
    & & & \textbf{Tumor Type} & \textbf{Histological Grading} & \textbf{Molecular Subtype} & \textbf{Average} & \\
    \midrule
    Quilt-LLaVA\cite{seyfioglu2024quilt} & Thumbnail & 0.132 & 0.113 & 0.194  & 0.181  & 0.163  & 0.603 \\
    WSI-VQA\cite{chen2025wsi} & WSI & 0.469 & 0.039  & 0.300  & 0.000  & 0.113  & 0.000  \\
    SlideChat\cite{chen2024slidechat} & WSI & - & 0.902  & 0.231  & 0.175  & 0.436  & - \\
    \textbf{WSI-LLaVA} & WSI & \textcolor{blue}{0.546} & \textcolor{blue}{0.905} & \textcolor{blue}{0.463} & \textcolor{blue}{0.292} & \textcolor{blue}{0.553} & \textcolor{blue}{0.721} \\
    \bottomrule
  \end{tabular}
  }
 \caption{Zero-shot external validation on three external datasets, i.e, WSI-VQA dataset, SlideBench-VQA (BCNB), and CPTAC-NSCLC. }
  \label{tab:external_validation}
\end{table*}


\subsection{Ablation Studies}
We conduct ablation studies to verify our WSI encoder and three-stage training scheme (Appendix Table~\ref{tab:combined_ablation}). Without the WSI encoder, performance drops across all pathological capabilities, with overall average decreasing from 68.0\% to 63.2\% (-5\%), demonstrating its effectiveness for WSI feature extraction. For our training approach, removing the WSI-text alignment stage reduces overall performance by 2.2\%, while ablating both this and the feature space alignment stage decreases performance by 3.3\%, confirming each stage's contribution to cross-modal alignment.

\subsection{External Validation}
To assess WSI-LLaVA's generalization capability, we performed zero-shot external validation on three datasets: WSI-VQA dataset, SlideBench(BCNB), and CPTAC-NSCLC. In Table~\ref{tab:external_validation}, WSI-LLaVA consistently outperforms existing methods across all datasets, showing robust generalization to non-TCGA sources. The performance gap widens significantly on tasks requiring detailed morphological interpretation, with WSI-LLaVA achieving 46.3\% vs. second-best 30.0\% for histological grading and 29.2\% vs. second-best 18.1\% for molecular subtyping. These results indicate WSI-LLaVA's strong morphological understanding enables more accurate zero-shot diagnostic assessments.

\subsection{Comparison of NLU Metrics and WSI Metrics}
Table~\ref{tab:wsi_vs_nlp} compares pathology report generation performance using traditional NLU and proposed WSI metrics. While Hist-Gen and WSI-LLaVA showed similar NLU scores (BLEU-1: 0.531 vs. 0.521, ROUGE-L: 0.536 vs. 0.534), their clinical accuracy differed significantly. Hist-Gen incorrectly reported ``no evidence of vascular invasion or necrosis" despite their presence in ground truth. WSI metrics better reflected clinical accuracy, with WSI-LLaVA scoring much higher (WSI-Precision: 0.8) than Hist-Gen (0.233). This reveals NLU metrics' limitations in prioritizing surface similarity and n-gram matching over factual correctness, emphasizing the need for domain-specific metrics that evaluate clinical accuracy rather than linguistic similarity in computational pathology.

\subsection{Evaluation on Morphological Capability}
\begin{figure}
  \centering
  \includegraphics[width=0.9\linewidth]{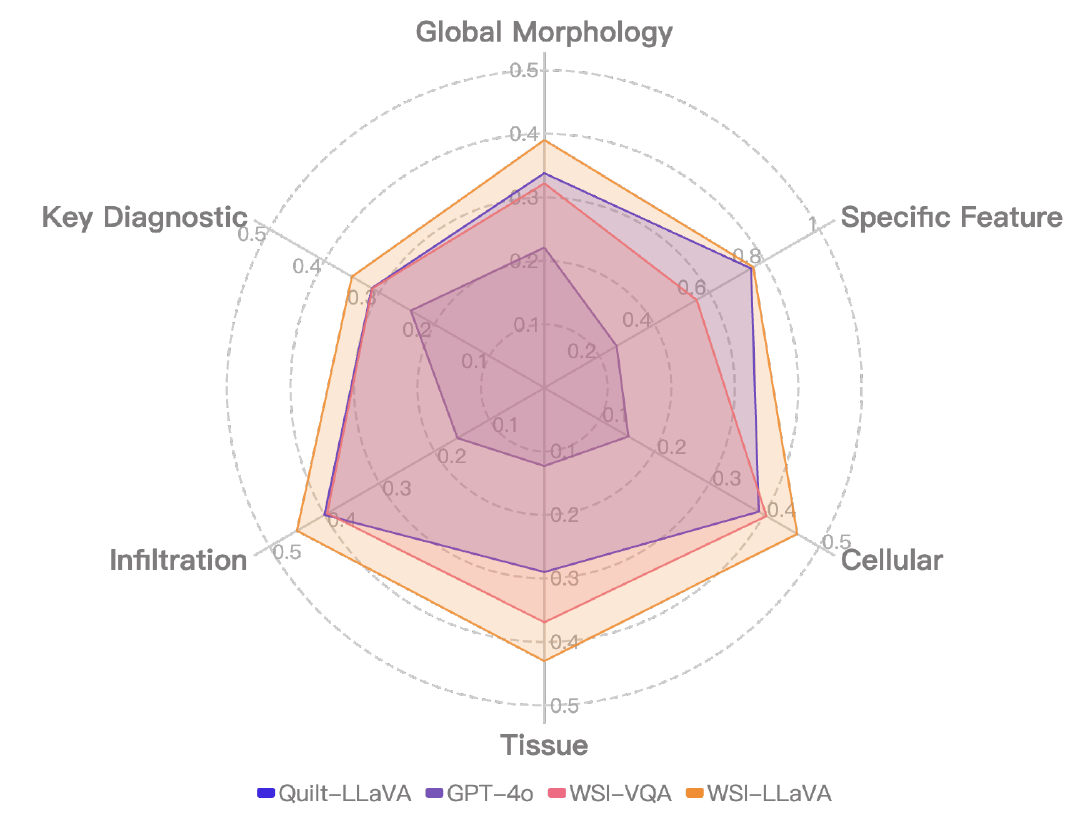}
  \caption{Morphological capability evaluation. The regional structure description includes three description sub-tasks, i.e., infiltration, tissue, and cellular. }
  \label{fig:radar_morph}
\end{figure}

To further analyze WSI MLLMs' morphological capabilities, Fig.~\ref{fig:radar_morph} compares performance across morphological tasks. Current WSI MLLMs and general MLLMs like GPT-4o show limited proficiency in both global and local morphological analysis. WSI-LLaVA consistently outperforms other WSI MLLMs across all six tasks, suggesting its robust morphological understanding enables superior performance at both analysis levels.

\section{Conclusion}
This work introduces WSI-Bench, the first morphology-aware benchmark containing 180k VQA pairs from 9,850 WSIs across 30 cancer types, alongside WSI-LLaVA, an MLLM for WSI analysis. Through three-stage training, WSI-LLaVA bridges the gap between WSIs and text, enabling detailed morphological reasoning and explainability. WSI-LLaVA outperforms existing methods across all pathological tasks on both internal and external datasets, while our WSI metrics provide more clinically relevant assessment than NLU metrics. By improving morphological understanding and explainability, WSI-LLaVA represents significant advancement toward trustworthy AI assistants for computational pathology.


\section*{Acknowledgments} This study was supported by the National Natural Science Foundation of China (Grants 82261138629 and 12326610), the Guangdong Provincial Key Laboratory (Grant 2023B1212060076), the Yongjiang Technology Innovation Project (Grant 2022A-097-G), and the National Natural Science Foundation of China grant (UNNC: B0166).
{
    \small
    \bibliographystyle{ieeenat_fullname}
    \bibliography{main}
}
\clearpage
\setcounter{page}{1}
\maketitlesupplementary
\setcounter{section}{0}
\setcounter{table}{0}

\noindent\textbf{Abstract.}{
In this supplementary material, we provide additional information about the proposed WSI-Bench, WSI-LLaVA, WSI Metrics.
Appendix~\ref{appendix:add_exps} illustrates additional experiments of the proposed WSI-LLaVA on the proposed WSI-Bench and supplementary details that complement the main experiments in the paper. Appendix~\ref{append:discussion} includes further discussion with key insights about WSI evaluation. Appendix~\ref{appendix:wsi_bench} demonstrates the details of WSI-Bench. Appendix~\ref{appendix:wsi_metric} provides the details of the WSI metrics. Appendix~\ref{appendix:implement} provides the implementation details of the proposed method.
}


\section{Additional Experiments} 
\label{appendix:add_exps}
\subsection{Quantitative Evaluation across Each Task}
To quantitatively evaluate each pathological task across three capabilities, we demonstrate the performance of existing MLLMs and our proposed WSI-LLaVA in Table~\ref{tab:comparison_all_tasks}. As shown in the results, WSI-LLaVA achieves superior overall performance with the highest average scores in both WSI-Precision (0.601) and WSI-Relevance (0.608). Specifically, in morphological analysis capability, WSI-LLaVA demonstrates leading WSI-Precision scores across all sub-tasks, with particular strength in specific feature description (0.760). For diagnostic capability, it exhibits robust performance in histological typing (0.410), grading (0.570), and staging (0.830), although Quilt-LLaVA shows slightly better results in molecular subtyping. Regarding treatment planning capability, WSI-LLaVA excels in treatment recommendations with the highest scores in both metrics (0.790 and 0.881), while maintaining competitive performance in prognosis. These comprehensive results validate WSI-LLaVA's balanced and reliable performance across diverse pathological tasks compared to existing MLLMs.

\begin{table*}[ht]
\centering
\caption{Quantitative comparison among MLLMs across three capabilities by using WSI-Precision and WSI-Relevance metrics.}
\renewcommand\arraystretch{1.2}
\setlength{\tabcolsep}{3pt}
\scalebox{0.88}{
\begin{tabular}{cccccccccccc}
\toprule
\multirow{3}{*}{Methods} & \multicolumn{4}{c}{\textbf{Morphological Analysis}} & \multicolumn{4}{c}{\textbf{Diagnosis}} & \multicolumn{2}{c}{\textbf{Treatment Planning}} & \multirow{3}{*}{Average} \\ \cmidrule(lr){2-5} \cmidrule(lr){6-9} \cmidrule(lr){10-11}
 & \makecell{Global\\Morph. Desc.} & \makecell{Key\\Diag. Desc.} & \makecell{Reg.\\Struct. Desc.} & \makecell{Spec.\\Feat. Desc.} & \makecell{Hist.\\Typing} & Grading & \makecell{Mol.\\Subtyping} & Staging & \makecell{Treat.\\Recom.} & Prognosis & \\
\midrule
& \multicolumn{10}{c}{\textbf{WSI-Precision}} \\
\midrule
{Quilt-LLaVA} & 0.338 & 0.314 & 0.389 & 0.752 & 0.339 & 0.505 & \textcolor{blue}{0.675} & 0.824 & 0.764 & 0.812 & 0.571\\
{GPT-4o} & 0.221 & 0.243 & 0.153 & 0.263 & 0.282 & 0.363 & 0.578 & 0.655 & 0.498 & 0.493 & 0.375\\
{WSI-VQA} & 0.322 & 0.313 & 0.389 & 0.554 & 0.377 & 0.430 & 0.388 & 0.550 & 0.708 & \textcolor{blue}{0.874} & 0.490\\
\textbf{WSI-LLaVA} & \textcolor{blue}{0.390} & \textcolor{blue}{0.350} & \textcolor{blue}{0.450} & \textcolor{blue}{0.760} & \textcolor{blue}{0.410} & \textcolor{blue}{0.570} & 0.630 & \textcolor{blue}{0.830} & \textcolor{blue}{0.790} & 0.830 & \textcolor{blue}{0.601}\\
\midrule
& \multicolumn{10}{c}{\textbf{WSI-Relevance}} \\
\midrule
{Quilt-LLaVA} & 0.289 & 0.363 & 0.404 & 0.732 & 0.333 & 0.453 & \textcolor{blue}{0.708} & \textcolor{blue}{0.922} & 0.858 & 0.773  & 0.584 \\
{GPT-4o} & 0.171 & 0.326 & 0.216 & 0.103 & 0.177 & 0.312 & 0.169 & 0.745 & 0.792 & \textcolor{blue}{0.889}  & 0.390\\
{WSI-VQA} & \textcolor{blue}{0.383} & 0.370 & \textcolor{blue}{0.511} & 0.586 & \textcolor{blue}{0.475} & \textcolor{blue}{0.514} & 0.476 & 0.634 & 0.444 & 0.746  &0.514 \\
\textbf{WSI-LLaVA} & 0.333 & \textcolor{blue}{0.382} & 0.465 & \textcolor{blue}{0.766} & 0.380 & 0.511 & 0.675 & 0.882 & \textcolor{blue}{0.881} & 0.808  & \textcolor{blue}{0.608} \\
\bottomrule
\end{tabular}}
\vspace{0.01cm}
\footnotesize Abbreviation: Global Morph. Desc. (Global Morphological Description), Key Diag. Desc. (Key Diagnostic Description), \\ Reg. Struct. Desc. (Regional Structure Description), Spec. Feat. Desc. (Specific Feature Description), Hist. Typing (Histological Typing),\\Grading (Grading), Mol. Subtyping (Molecular Subtyping), Staging (Staging), Treat. Recom. (Treatment Recommendations), Prognosis (Prognosis).
\label{tab:comparison_all_tasks}
\end{table*}

\subsection{Qualitative Evaluation across Each Task}
To qualitatively evaluate each task, we demonstrate the visual results of existing methods and our proposed WSI-LLaVA for 11 pathological tasks, as shown in Table~\ref{tab:sample_comparison_report_generation} to~\ref{tab:sample_comparison_prognosis}. We use color green to highlight content that matches the ground truth and red to indicate content that contradicts it. 
In these Tables, we can observe that WSI-LLaVA's answers are more consistent with the ground truth and exhibit higher accuracy across all tasks. While Quilt-LLaVA’s responses are similar in form to the ground truth, they contain more errors in specific content, especially in morphological description tasks. This may be due to its inability to extract the rich morphological features in WSI images. WSI-VQA’s answers often include repeated invalid information, and some sentences have grammatical errors, indicating the fundamental disadvantages of Transformer-based models compared to LLMs. GPT-4o provides a large amount of irrelevant information, likely because it cannot be trained to read WSI images.

\begin{figure*}[ht]
  \centering
  \includegraphics[width=0.95\linewidth]{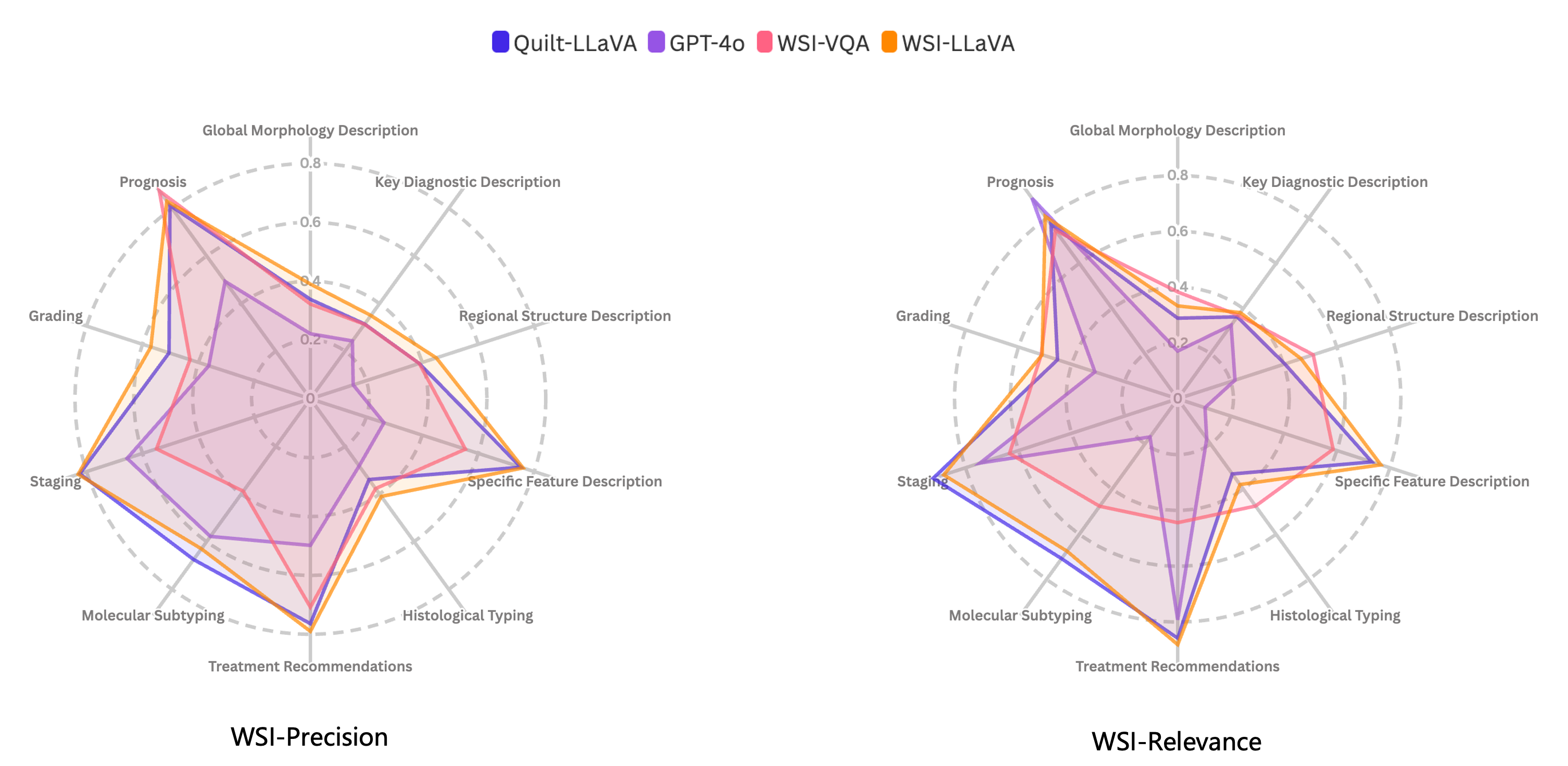} 
\caption{Radar charts of WSI-Precision and WSI-Relevance for MLLMs across different tasks.}
  \label{fig:radar}
\end{figure*}

In the radar chart displaying WSI-Precision metrics (Figure~\ref{fig:radar}), WSI-LLaVA (our model) dominates with broader coverage and higher peaks in most diagnostic categories, particularly excelling in ``Specific Feature Description," ``Staging," and ``Prognosis." This suggests an excellent ability to accurately identify and describe critical pathological features and outcomes. Meanwhile, models like GPT-4o show considerably lower precision, particularly in detailed descriptions, which may limit their utility in nuanced diagnostic scenarios.

The WSI-Relevance radar chart reveals that WSI-LLaVA also leads in relevance (Figure~\ref{fig:radar}), with outstanding performance in ``Staging" and ``Treatment Recommendations," underscoring its capability to deliver clinically pertinent information that aids in treatment planning and prognosis estimation. In contrast, while GPT-4o and WSI-VQA provide valuable insights in specific areas such as ``Prognosis" and ``Regional Structure Description," they exhibit a balanced but generally lower relevance compared to our model, indicating a need for targeted improvements to enhance their practical application in clinical settings.

\subsection{Ablation Studies Details}
\label{appendix:ablation_studies}
\noindent\textbf{Effectiveness of WSI Encoder.} To prove the effectiveness of WSI encoder in extracting WSI features, we ablate the WSI encoder in our proposed WSI-LLaVA framework, as listed in Table~\ref{tab:combined_ablation}. When ablating WSI encoder, across all three pathological capabilities and both report generation tasks, the model with WSI encoder consistently outperforms the variant without it. Notable improvements are observed in the WSI-R metrics for morphological analysis (0.447 to 0.610) and treatment planning (0.816 to 0.845). The overall average performance increases from 0.632 to 0.680, representing a 4.8\% improvement. These results clearly validate the effectiveness of the WSI encoder in extracting meaningful features from whole slide images, contributing to better performance across all evaluation metrics.

\noindent\textbf{Effectiveness of the three-stage training scheme.} To demonstrate the necessity of the three-stage training scheme for the WSI-LLaVA framework, we ablate the first or second stage to compare performance. As shown in Table~\ref{tab:combined_ablation}, when ablating the first stage, the overall performance declines by 2.2\% in average score. When ablating both the first and second stages, the performance decreases significantly by a large margin with an average score of 3.3\%, indicating that each stage contributes to improving performance to different extents. This highlights the effectiveness of three-stage training in mitigating the cross-modal gap.

\begin{table*}[hpt]
  \centering
    \setlength\tabcolsep{2.3pt}
  \scalebox{0.9}{
  \begin{tabular}{c c c c c c c c c c c c c c c}
    \toprule
    \small
    \multirow{3}{*}{\textbf{WSI Encoder}} & \multirow{3}{*}{\textbf{Stage 1}} & \multirow{3}{*}{\textbf{Stage 2}} & \multicolumn{3}{c}{\textbf{Morphological Analysis}} & \multicolumn{3}{c}{\textbf{Diagnosis}} & \multicolumn{3}{c}{\textbf{Treatment Planning}} & \multicolumn{2}{c}{\textbf{Report Generation}} & \multirow{3}{*}{\textbf{Average}}\\
    \cmidrule(lr){4-6} \cmidrule(lr){7-9} \cmidrule(lr){10-12} \cmidrule(lr){13-14}
    & & & \multicolumn{2}{c}{Open} & Close & \multicolumn{2}{c}{Open} & Close & \multicolumn{2}{c}{Open} & Close & \multirow{2}{*}{WSI-P} & \multirow{2}{*}{WSI-R} & \\
    & & & WSI-P & WSI-R & Acc& WSI-P & WSI-R & Acc& WSI-P & WSI-R & Acc& & &\\
    \midrule
    \ding{55} & \ding{51} & \ding{51} & 0.435 & 0.433 & 0.940 & 0.565 & 0.577 & 0.847 & 0.739 & 0.803 & 1.000 & 0.290 & 0.326 & 0.632\\
    \ding{51} & \ding{55} & \ding{55} & 0.425 & 0.483 & 0.945 & 0.561 & 0.541 & 0.837 & 0.761 & 0.798 & 1.000 & 0.356 & 0.367 & 0.643\\
    \ding{51} & \ding{55} & \ding{51} & 0.442 & 0.478 & 0.945 & 0.589 & 0.602 & 0.856 & 0.744 & 0.776 & 1.000& 0.380 & 0.377 & 0.654\\
    \ding{51} & \ding{51} & \ding{51} & \textcolor{blue}{0.488} & \textcolor{blue}{0.487}& \textcolor{blue}{0.951} & \textcolor{blue}{0.610}& \textcolor{blue}{0.612} & \textcolor{blue}{0.863} & \textcolor{blue}{0.810}& \textcolor{blue}{0.845} & \textcolor{blue}{1.000} & \textcolor{blue}{0.380} & \textcolor{blue}{0.429} & \textcolor{blue}{0.680}\\
    \bottomrule
  \end{tabular}}
    \caption{Combined ablation studies on WSI encoder and training stages of WSI-LLaVA. WSI-P: WSI-Precision, WSI-R: WSI-Relevance, Acc: accuracy, Open: open-ended questions, Close: close-ended questions.}
  \label{tab:combined_ablation}
\end{table*}

\subsection{External Validation Details}
\label{appendix:external_validation}

\noindent\textbf{WSI-VQA Dataset.} From the original WSI-VQA dataset~\cite{chen2025wsi}, we removed clinically unreasonable questions. Specifically, we excluded survival time predictions, as these cannot be reliably determined from histological images alone; estimations of 3D tumor size, which exceed the analytical scope of 2D pathology slides; and immunohistochemistry evaluations that depend on specialized staining techniques.

\noindent\textbf{SlideBench-VQA(BCNB).} From the SlideBench dataset~\cite{chen2024slidechat}, we excluded questions that require information on immunohistochemical markers, such as PR, ER, and HER2 status.

\noindent\textbf{CPTAC-NSCLC Dataset.} For ease of evaluation, based on the classification labels of the CPTAC-NSCLC dataset, we constructed the question: "Which carcinoma type is indicated by the tissue slide? Choose the correct option: A) Lung Adenocarcinoma; B) Lung Squamous Cell Carcinoma." The final test set consisted of 1,091 WSIs, with 546 labeled as LUAD (Lung Adenocarcinoma) and 545 as LSCC (Lung Squamous Cell Carcinoma).

\section{Discussion}
\label{append:discussion}

From the above results, three key insights have been deduced as follows:

1) \textbf{WSI-LLaVA demonstrates superior morphological understanding across all dimensions.} As illustrated in Fig.~\ref{fig:radar_morph}, our model consistently outperforms existing approaches in critical morphological analysis tasks. WSI-LLaVA achieves notable advantages in specific feature detection (0.76 vs. 0.752 for Quilt-LLaVA), cellular morphology analysis (0.46 vs. 0.404 for WSI-VQA), and infiltration pattern recognition (0.45 vs. 0.40 for Quilt-LLaVA). The performance gap is particularly pronounced when compared with general-purpose models like GPT-4o, which averages only 0.193 across morphological tasks versus WSI-LLaVA's 0.473. This superiority stems from our model's dual-encoder architecture that effectively captures both local details and global context, combined with a three-stage training approach that ensures proper alignment between visual and textual representations. Such comprehensive morphological understanding directly translates to improved diagnostic accuracy, as demonstrated in our qualitative analysis in Table~\ref{tab:qual_vqa}.

2) \textbf{Morphological understanding enables robust cross-dataset generalization.} External validation results reveal a striking pattern: models with stronger morphological capabilities demonstrate substantially better generalization to unseen datasets. This relationship is most evident in tasks requiring nuanced morphological assessment, where WSI-LLaVA outperforms the next best model by substantial margins on histological grading (46.33\% vs. 23.11\%) and molecular subtyping (29.2\% vs. 18.1\%). Unlike previous approaches that may learn dataset-specific correlations, WSI-LLaVA's focus on fundamental morphological principles provides a more transferable form of knowledge. This finding has significant implications for clinical deployment, suggesting that morphology-aware models can maintain consistent performance across diverse institutional settings and patient populations without requiring extensive retraining or adaptation.

3) \textbf{Traditional NLU metrics fail to capture clinical accuracy in pathology language tasks.} Table~\ref{tab:wsi_vs_nlp} shows that both Hist-Gen and WSI-LLaVA achieve similar scores on standard NLU metrics (e.g., BLEU-1: 0.531 vs. 0.521, ROUGE-L: 0.536 vs. 0.534), despite evident differences in their clinical accuracy. This is because traditional NLU metrics focus on surface similarity, such as lexical overlap, rather than the correctness and relevance of key pathological findings. For example, Hist-Gen's response—"There is no evidence of vascular invasion ..."—is structurally similar to the ground truth and thus receives high NLU metric scores, but its meaning is completely opposite. In contrast, WSI-LLaVA accurately identifies these features, achieving superior WSI metrics (WSI-P: 0.800, WSI-R: 0.830). This demonstrates the limitations of conventional NLU metrics in evaluating pathology language tasks, where capturing clinically relevant information is essential. Our proposed WSI-Precision and WSI-Relevance metrics address these limitations by focusing on factual correctness and clinical relevance rather than linguistic similarity. These metrics provide a more reliable and clinically meaningful assessment of model performance in pathological contexts.
WSI-LLaVA demonstrates substantially higher clinical precision and relevance (WSI-Precision: 0.800 vs. 0.233, WSI-Relevance: 0.830 vs. 0.520), highlighting the inability of traditional metrics to reflect clinical utility.

This discrepancy stems from three key limitations of conventional metrics:

1) \textbf{Surface Similarity}: High word overlap between generated text and references inflates scores without considering clinical accuracy. For example, phrases like "Microscopic observation reveals poorly differentiated squamous cell carcinoma" appear in both Hist-Gen and WSI-LLaVA, leading to similar scores despite differences in factual correctness.

2) \textbf{n-gram Matching}: Shared terms such as "nuclear pleomorphism" and "prominent nucleoli" contribute to high scores, even when these terms are used in incorrect or contradictory contexts.

3) \textbf{Semantic Insensitivity}: Conventional metrics fail to penalize clinically significant errors, such as Hist-Gen's incorrect statement that "there is no evidence of vascular invasion or necrosis," which directly contradicts the reference text.

Our proposed WSI-Precision and WSI-Relevance metrics address these limitations by focusing on factual correctness and clinical relevance rather than linguistic similarity. These metrics provide a more reliable and clinically meaningful assessment of model performance in pathological contexts.

\section{WSI-Bench}
\label{appendix:wsi_bench}
To establish a comprehensive and large-scale WSI-Bench, we include an automated data construction process using GPT-4o and a subsequent validation process by clinical experts. 

\subsection{Data Distribution} 
\begin{figure*}[ht]
  \centering
  \includegraphics[width=1\linewidth]{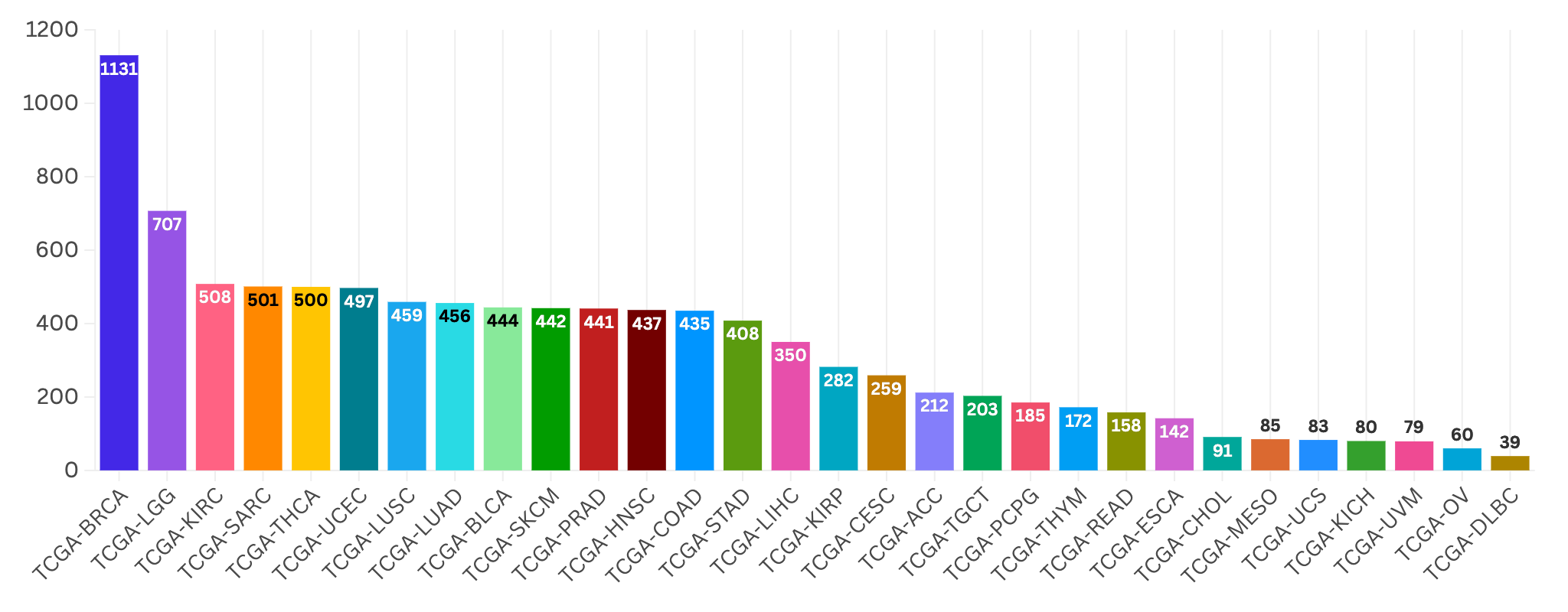} 
  \caption{Data distribution of WSI samples across various TCGA cancer types.}
  \label{fig:type-count}
\end{figure*}

In Fig. \ref{fig:type-count}, we illustrate the distribution of WSI across various cancer types as categorized in the Cancer Genome Atlas (TCGA) database.
Cancer types are commonly identified by standardized abbreviations. Specifically, TCGA-ACC refers to Adrenocortical Carcinoma; TCGA-BLCA is Bladder Urothelial Carcinoma; TCGA-BRCA denotes Breast Invasive Carcinoma; TCGA-CESC stands for Cervical Squamous Cell Carcinoma and Endocervical Adenocarcinoma; TCGA-CHOL represents Cholangiocarcinoma; TCGA-COAD is Colon Adenocarcinoma; TCGA-DLBC refers to Lymphoid Neoplasm Diffuse Large B-cell Lymphoma; TCGA-ESCA denotes Esophageal Carcinoma; TCGA-HNSC stands for Head and Neck Squamous Cell Carcinoma; TCGA-KICH represents Kidney Chromophobe; TCGA-KIRC is Kidney Renal Clear Cell Carcinoma; TCGA-KIRP denotes Kidney Renal Papillary Cell Carcinoma; TCGA-LGG refers to Brain Lower Grade Glioma; TCGA-LIHC stands for Liver Hepatocellular Carcinoma; TCGA-LUAD represents Lung Adenocarcinoma; TCGA-LUSC is Lung Squamous Cell Carcinoma; TCGA-MESO denotes Mesothelioma; TCGA-OV refers to Ovarian Serous Cystadenocarcinoma; TCGA-PCPG stands for Pheochromocytoma and Paraganglioma; TCGA-PRAD represents Prostate Adenocarcinoma; TCGA-READ is Rectum Adenocarcinoma; TCGA-SARC denotes Sarcoma; TCGA-SKCM refers to Skin Cutaneous Melanoma; TCGA-STAD stands for Stomach Adenocarcinoma; TCGA-TGCT represents Testicular Germ Cell Tumors; TCGA-THCA is Thyroid Carcinoma; TCGA-THYM denotes Thymoma; TCGA-UCEC stands for Uterine Corpus Endometrial Carcinoma; TCGA-UCS represents Uterine Carcinosarcoma; and TCGA-UVM refers to Uveal Melanoma.

\subsection{Training and Testing Distribution}
\begin{figure*}[h]
\centering
\includegraphics[width=\linewidth]{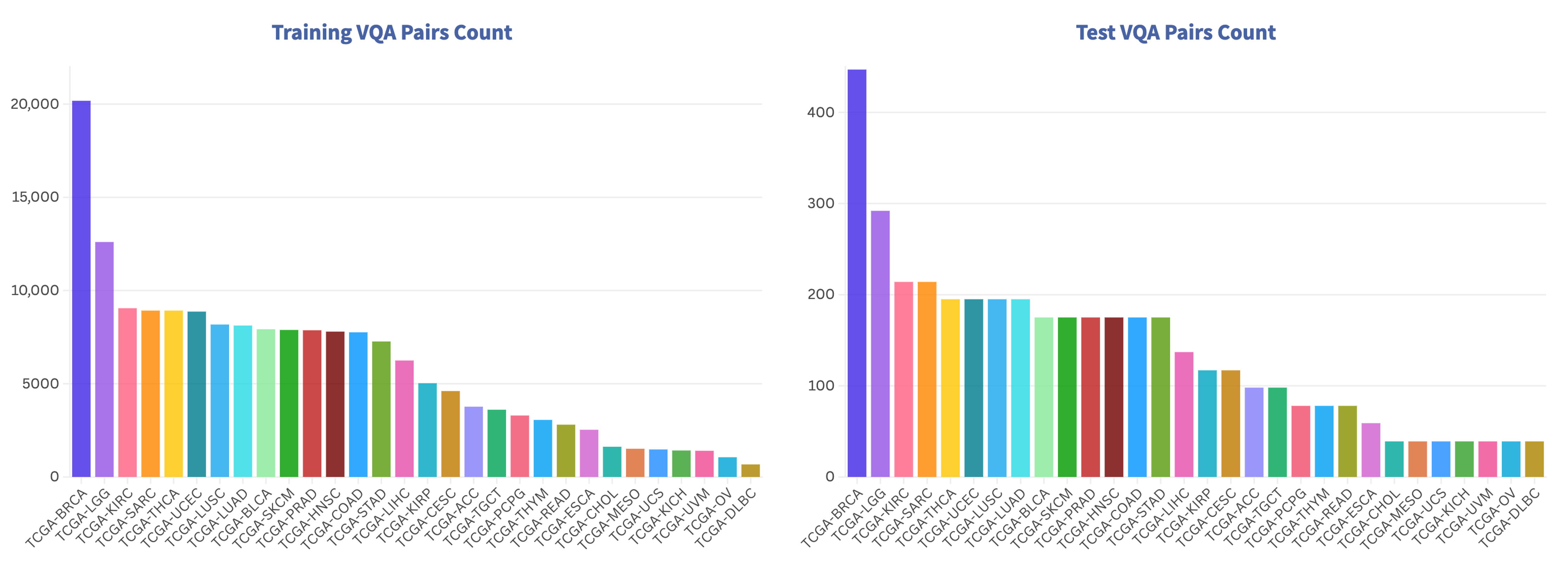}
\caption{Distribution of VQA pairs across training and testing sets by cancer type.}
\label{fig:train_test_distribution}
\end{figure*}

We strictly separated training and testing sets by cancer type to evaluate model generalization. For each cancer type, we allocated 2\% of WSIs for testing (at least 2 WSIs for smaller cancer types). The final distribution includes 9,642 WSIs with 175,450 VQA pairs for training and 208 WSIs with 4,119 VQA pairs for testing.Figure \ref{fig:train_test_distribution} shows the distribution of VQA pairs across training and testing sets for each cancer type. 

\subsection{Reverse-Engineer Scheme}
 To enrich morphological descriptions, we introduce a reverse-engineer scheme to provide a morphological description extension. It uses diagnostic labels to identify the corresponding morphological characteristics according to tumor classification standards like the Nottingham grading system.  This extension is included in the WSI reports that are then refined by clinical experts. 
This reverse-engineering process is supported by the breadth of medical literature included in GPT-4o’s training data, ensuring a high level of precision that mirrors the expertise of professional pathologists.

Specifically, we take the instance of breast cancer where reports utilize the Nottingham grading system~\cite{dilani2019classification}, a method frequently used in TCGA datasets. The grading system assesses three critical aspects: gland formation, nuclear pleomorphism, and mitotic count, assigning each a score from 1 to 3. Consider a report that assigns a score of 3 for both gland formation and nuclear pleomorphism. This high score indicates minimal gland formation and substantial variation in nuclear size and shape, suggestive of aggressive tumor characteristics. Furthermore, if the pathology report confirms a diagnosis of invasive ductal carcinoma, it characterizes the cancer cells as having breached duct walls and forming invasive clusters. Such diagnostic details are crucial as they provide clear visual markers that can be directly observed in WSIs. We incorporate these specific findings into our dataset to enrich the morphological descriptions, ensuring that they reflect real-world pathological assessments and are tightly aligned with established diagnostic criteria.

Table~\ref{tab:nottingham_detailed} presents an example of the Nottingham grading system applied to breast cancer, a method derived from the WHO Classification of Tumors~\cite{dilani2019classification}. This system is recognized for its rigor and authority in classifying tumor severity. It systematically scores three critical histological features: gland formation, nuclear pleomorphism, and mitotic count. Each feature is evaluated based on specific morphological characteristics observable in tissue samples. The scores correlate directly with these features, ensuring that the derived descriptions are not only standardized but also clinically meaningful. By adhering to this methodology, the descriptions generated are both accurate reflections of the histopathological findings and relevant for clinical assessment and treatment planning.

To address \textbf{hallucinations}, we generate three versions of each report and use another LLM to verify if the content aligns with the original reports. The model performs five iterations of selection, and if one version is selected three or more times, it is deemed accurate. If none achieve this threshold, a pathologist reviews and selects the appropriate content. Furthermore, to avoid the pitfalls of incomplete filtering, especially concerning immunohistochemical data and macroscopic descriptions, we use regular expressions to identify and exclude 30 keywords, with any flagged text revised by a pathologist.

\subsection{Data Construction}

\begin{figure*}[ht]
  \centering
  \begin{minipage}{\textwidth}
    \centering
    \includegraphics[width=1\linewidth]{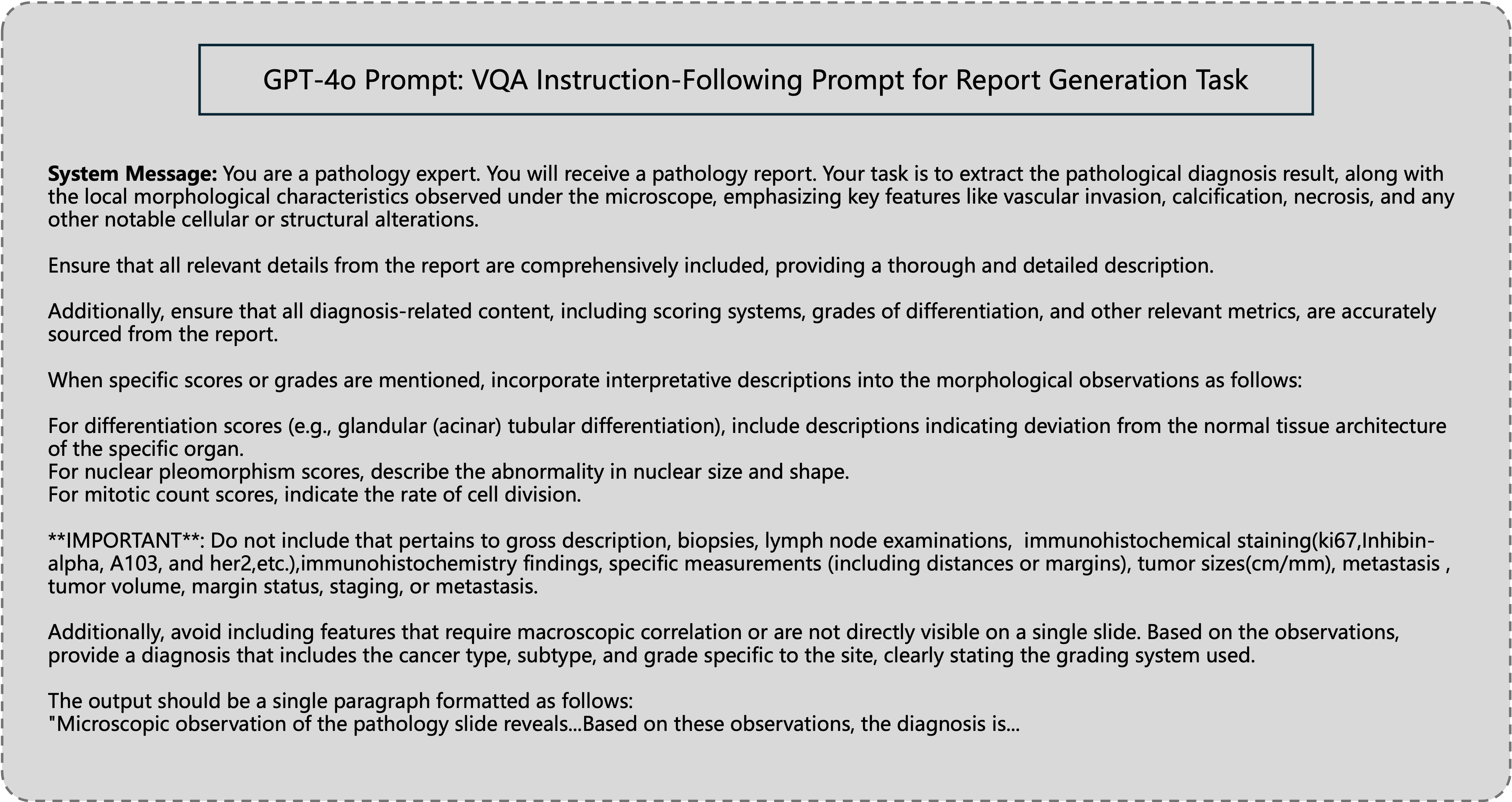}
    \caption{The prompt used to generate VQA pairs for report generation task.}
    \label{fig:prompt_report_generation}
  \end{minipage}

  \vspace{1em} 

  \begin{minipage}{\textwidth}
    \scriptsize
    \centering
    \captionof{table}{VQA example for report generation task.}
    \begin{tabular}{>{\raggedright\arraybackslash}m{0.2\linewidth} >{\raggedright\arraybackslash}m{0.7\linewidth}}
    \toprule[1.5pt]  
    \multicolumn{2}{l}{\textbf{Example of Report Generation}} \\
    \midrule
    \textbf{WSI} & \includegraphics[width=0.3\linewidth]{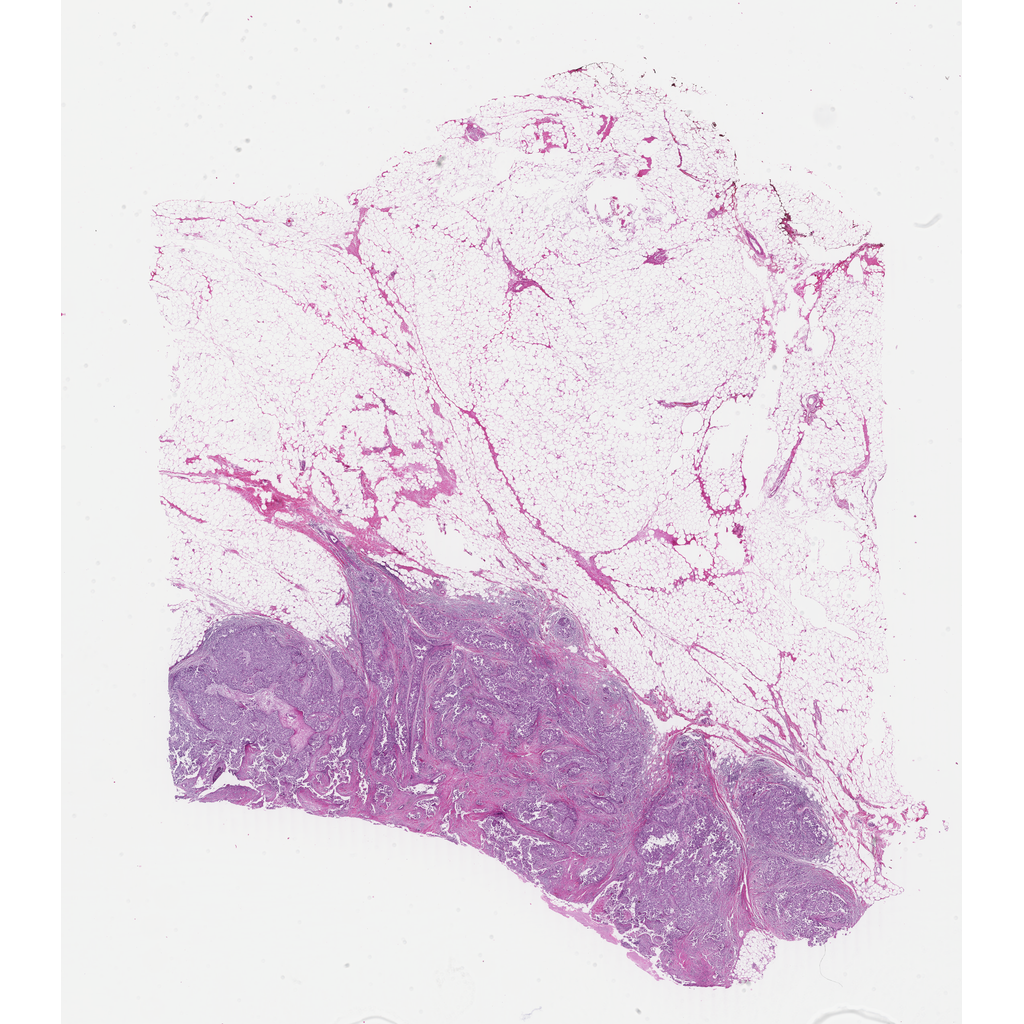} \\ 
    \textbf{Question} & What are the main observations in the whole slide histopathology image, and what is the likely diagnosis? \\
    \midrule
    \textbf{Answer} &  Microscopic observation of the pathology slide reveals tumor cells infiltrating the surrounding stroma with minimal formation of glandular structures, significantly deviating from normal tissue architecture, and showing a poorly differentiated state with a tubule formation score of 3. The nuclei exhibit considerable size and shape variability, corresponding to a nuclear pleomorphism score of 3. There is a moderate rate of cell division observed, reflected by a mitotic count score of 2. Vascular invasion is not unequivocally present, and no necrosis or microcalcifications are observed. The intraductal component is noted to be less than 25\% with prominent cancerization. Based on these observations, the diagnosis is infiltrating ductal carcinoma, no special type, Nottingham Grade III, poorly differentiated, with a total Nottingham score of 8/9. \\ 
    \bottomrule[1.5pt]
    \end{tabular}
    \label{tab:report_generation_example}
  \end{minipage}
\end{figure*}

\noindent\textbf{Report Generation Task.} Fig.~\ref{fig:prompt_report_generation} illustrate how prompts are utilized to guide GPT-4o in generating textual reports that encompass critical histopathological assessments. Table~\ref{tab:report_generation_example} shows the example of the WSI and the corresponding report. These examples demonstrate the process of crafting prompts that effectively elicit comprehensive and clinically relevant reports from the model, ensuring essential diagnostic information is accurately captured.

\begin{figure*}[ht]
  \centering
  \begin{minipage}{\textwidth}
    \centering
    \includegraphics[width=1\linewidth]{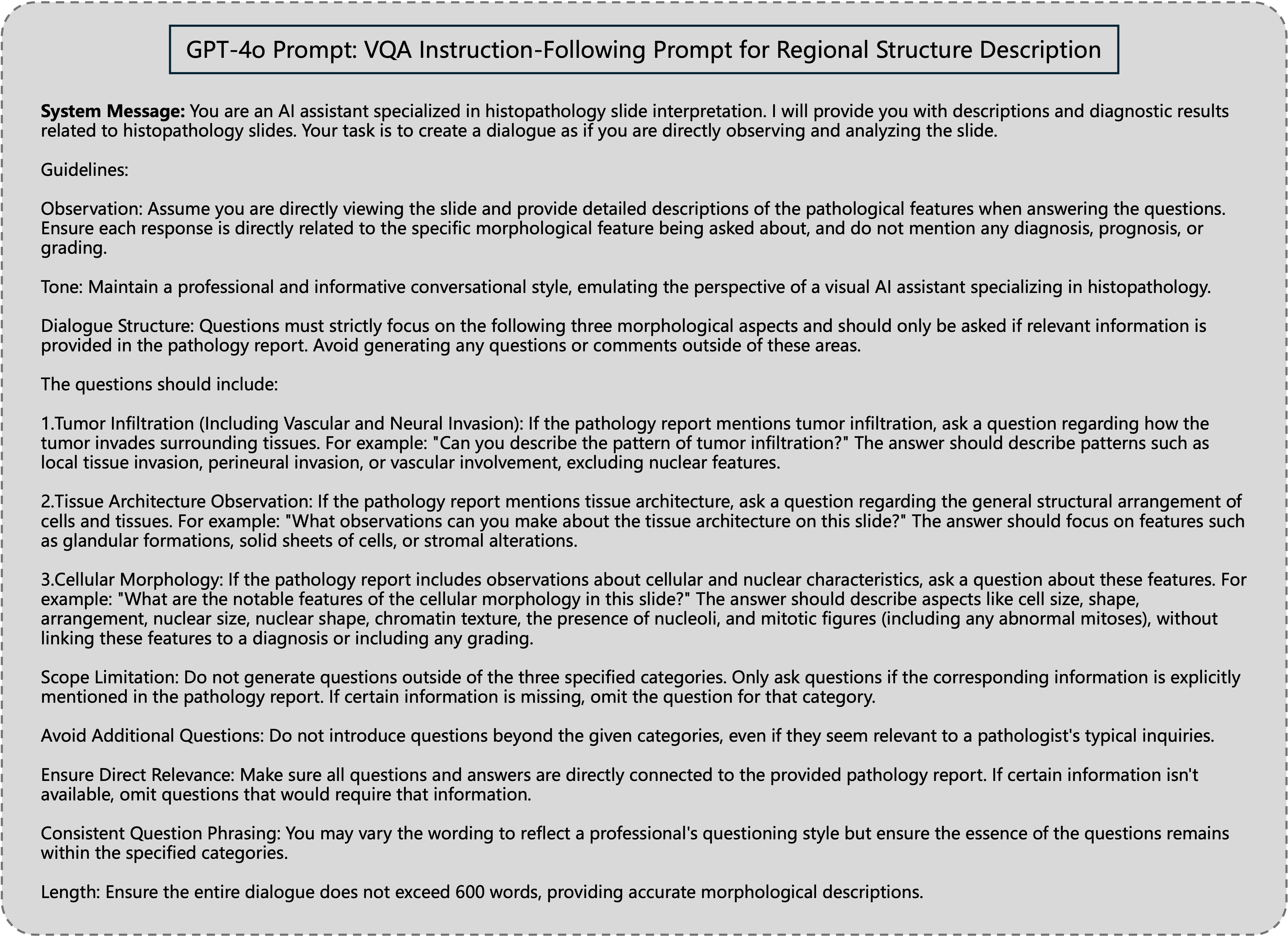}
    \caption{The prompt used to generate VQA pairs for regional structure description}
    \label{fig:prompt_regional_structure_description}
  \end{minipage}
  
  \vspace{1em} 
  
  \begin{minipage}{\textwidth}
    \scriptsize
    \centering
    \captionof{table}{VQA example for regional structure description task.}
    \begin{tabular}{>{\raggedright\arraybackslash}m{0.2\linewidth} >{\raggedright\arraybackslash}m{0.7\linewidth}}
    \toprule[1.5pt]  
    \multicolumn{2}{l}{\textbf{Example of Regional Structure Description}} \\
    \midrule
    \textbf{WSI} & \includegraphics[width=0.3\linewidth]{sample/TCGA-A2-A0CX-01Z-00-DX1.F07C75AB-E568-45CB-B497-37C712490393.svs_original.png} \\ \\
    \textbf{Question} & What observations can you make about the tissue architecture on this slide? \\
    \midrule
    \textbf{Answer} & The tissue architecture is significantly disrupted, marked by minimal formation of glandular structures. The poorly differentiated state results in a loss of the typical organized glandular architecture, and instead, cells are haphazardly arranged. There is less than 25\% of the slide showing an intraductal component, and this area exhibits prominent cancerization of lobules. \\ 
    \bottomrule[1.5pt]
    \end{tabular}
    \label{tab:regional_structure_description_example}
  \end{minipage}
\end{figure*}

\begin{figure*}[ht]
  \centering
  \begin{minipage}{\textwidth}
    \centering
    \includegraphics[width=1\linewidth]{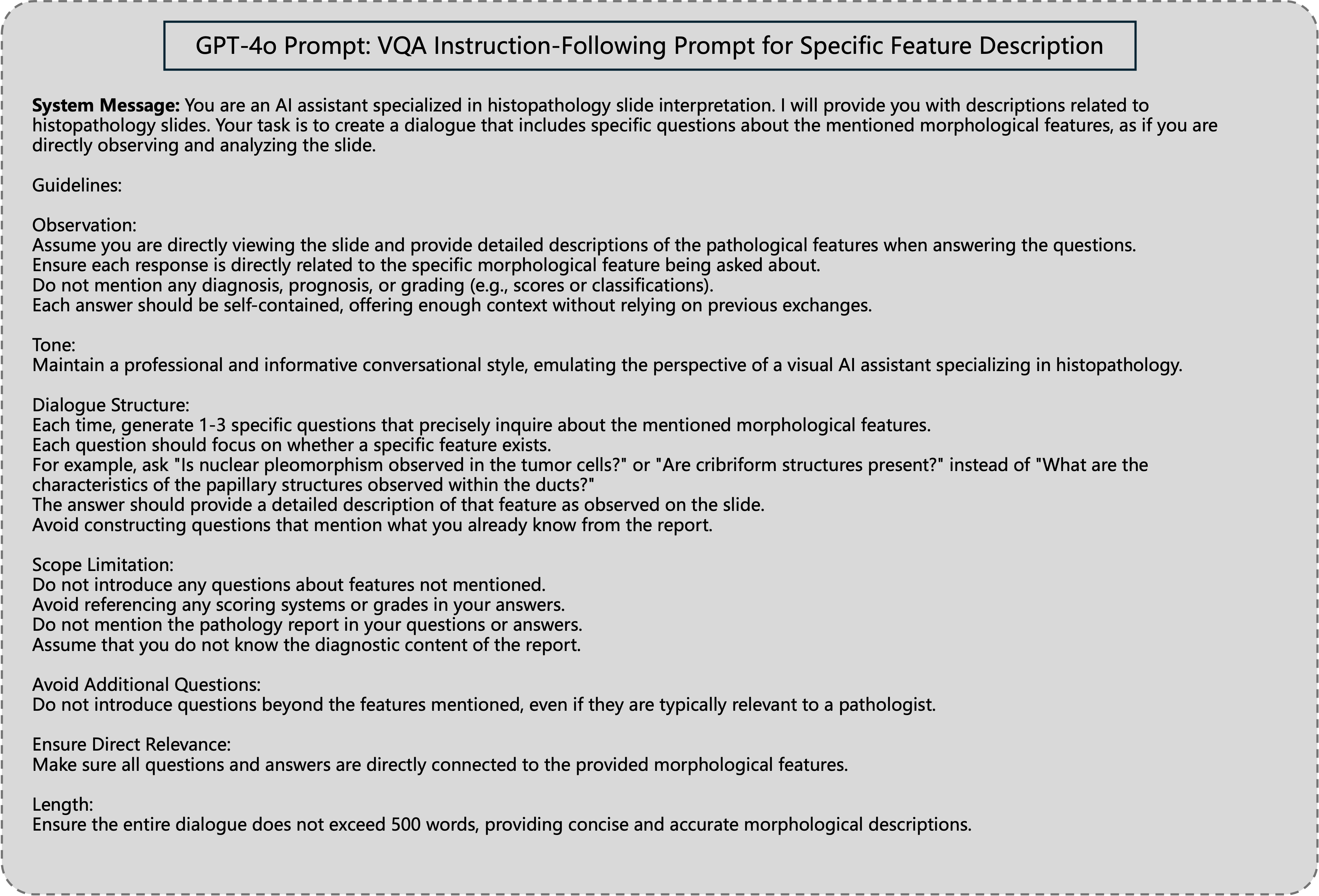}
    \caption{The prompt used to generate VQA pair for specific feature description}
    \label{fig:prompt_specific_feature_description}
  \end{minipage}
  
  \vspace{1em} 
  
  \begin{minipage}{\textwidth}
    \scriptsize
    \centering
    \captionof{table}{VQA example for specific feature description task.}
    \begin{tabular}{>{\raggedright\arraybackslash}m{0.2\linewidth} >{\raggedright\arraybackslash}m{0.7\linewidth}}
    \toprule[1.5pt]  
    \multicolumn{2}{l}{\textbf{Example of Specific Feature Description}} \\
    \midrule
    \textbf{WSI} & \includegraphics[width=0.3\linewidth]{sample/TCGA-A2-A0CX-01Z-00-DX1.F07C75AB-E568-45CB-B497-37C712490393.svs_original.png} \\ \\
    \textbf{Question} & Do the nuclei of the tumor cells show nuclear pleomorphism? \\
    \midrule
    \textbf{Answer} & Indeed, the nuclei of the tumor cells exhibit considerable nuclear pleomorphism, featuring significant variability in size and shape. They present irregular contours, are substantially larger than normal, and possess dense chromatin, highlighting marked atypical nuclear features. \\ 
    \bottomrule[1.5pt]
    \end{tabular}
    \label{tab:specific_feature_description_example}
  \end{minipage}
\end{figure*}

\begin{figure*}[ht]
  \centering
  \includegraphics[width=1\linewidth]{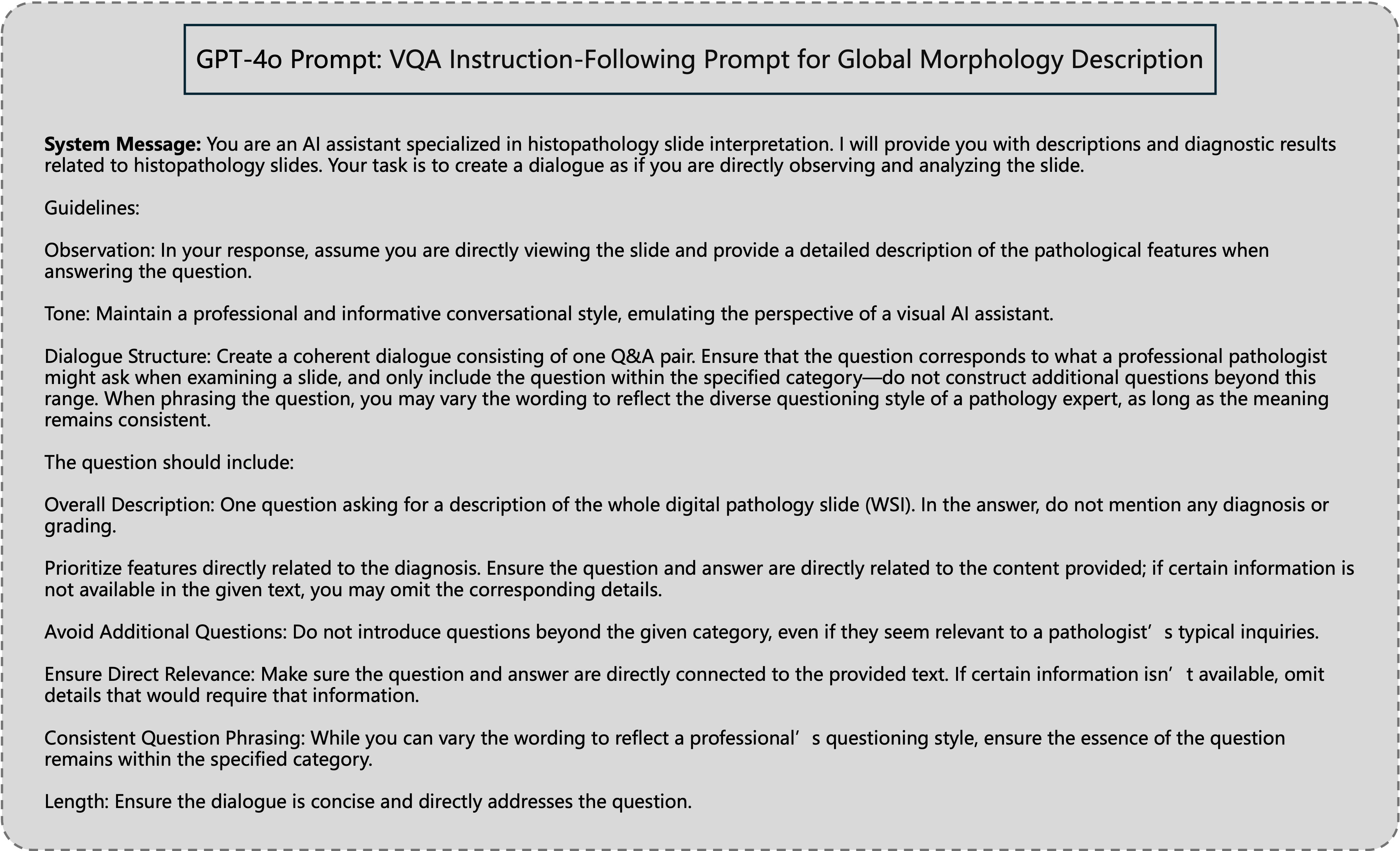} 
  \caption{The prompt used to generate VQA pair for global morphology description task.}
  \label{fig:prompt_global_morphology_description}
\end{figure*}

\begin{figure*}[ht]
  \centering
  \includegraphics[width=1\linewidth]{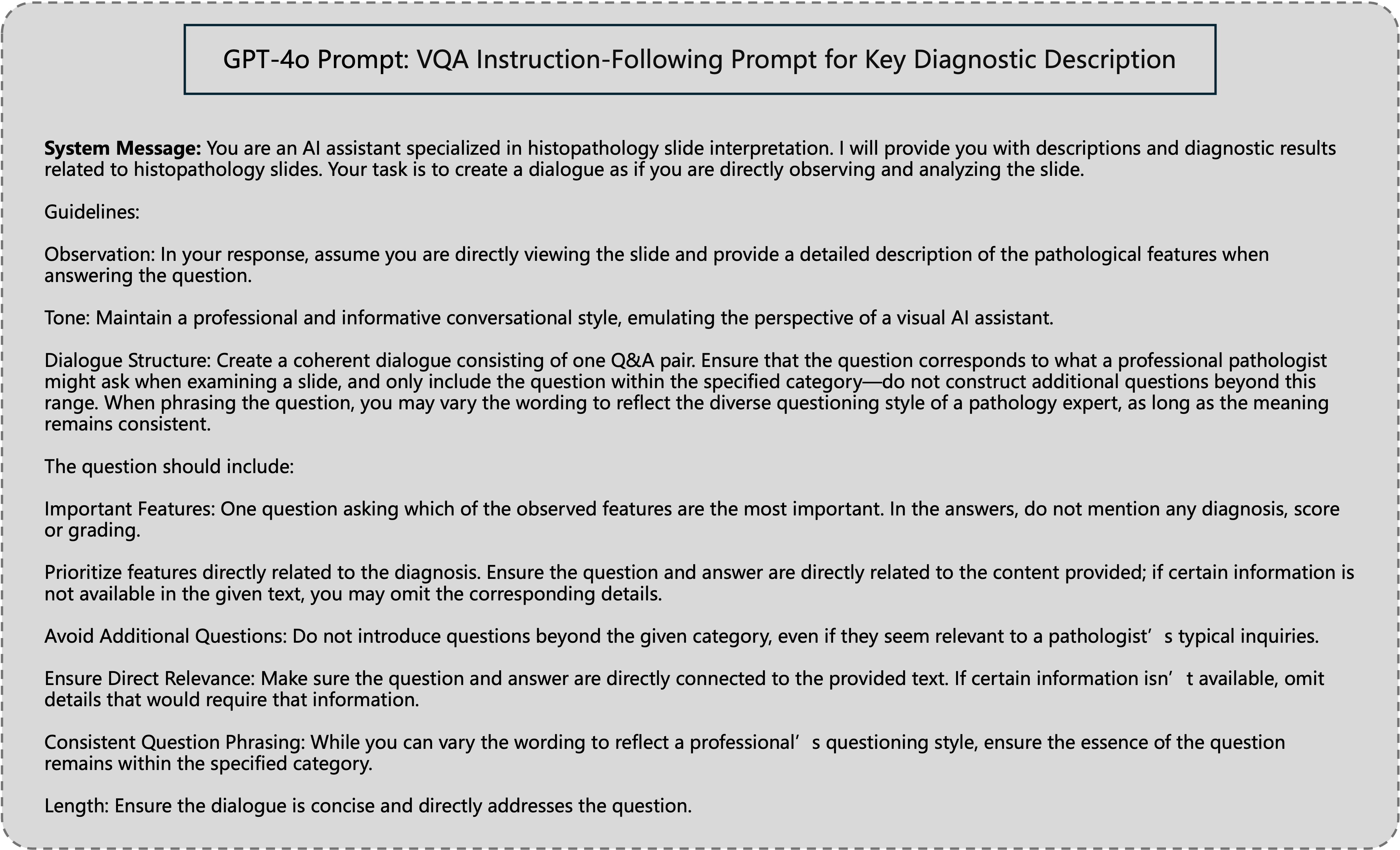} 
  \caption{The prompt used to generate VQA pair for key diagnostic description task.}
  \label{fig:prompt_key_diagnostic_description}
\end{figure*}

\begin{figure*}[ht]
  \centering
  \includegraphics[width=1\linewidth]{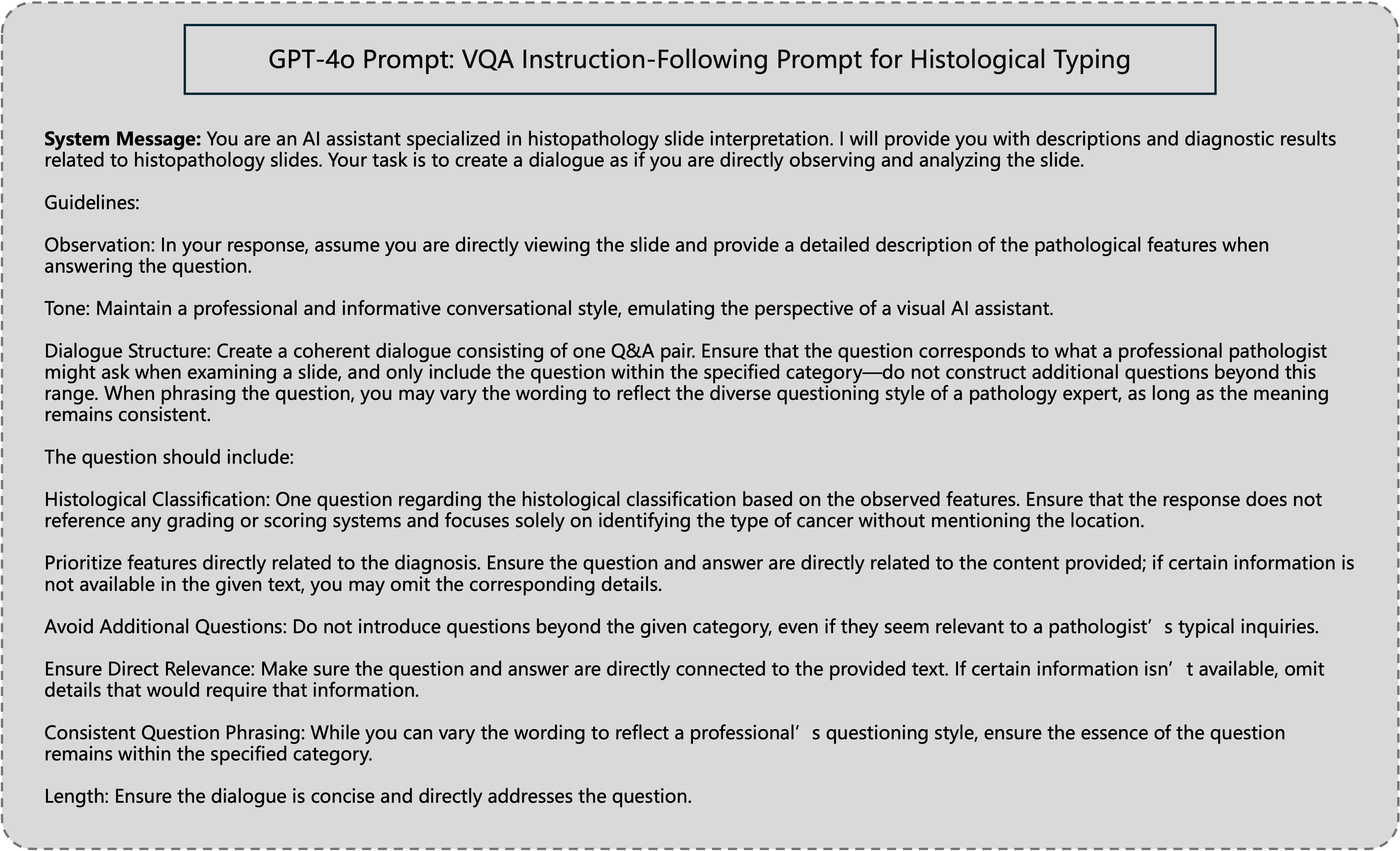} 
  \caption{The prompt used to generate VQA pair for histological typing task.}
  \label{fig:prompt_histological_typing}
\end{figure*}

\begin{figure*}[ht]
  \centering
  \includegraphics[width=1\linewidth]{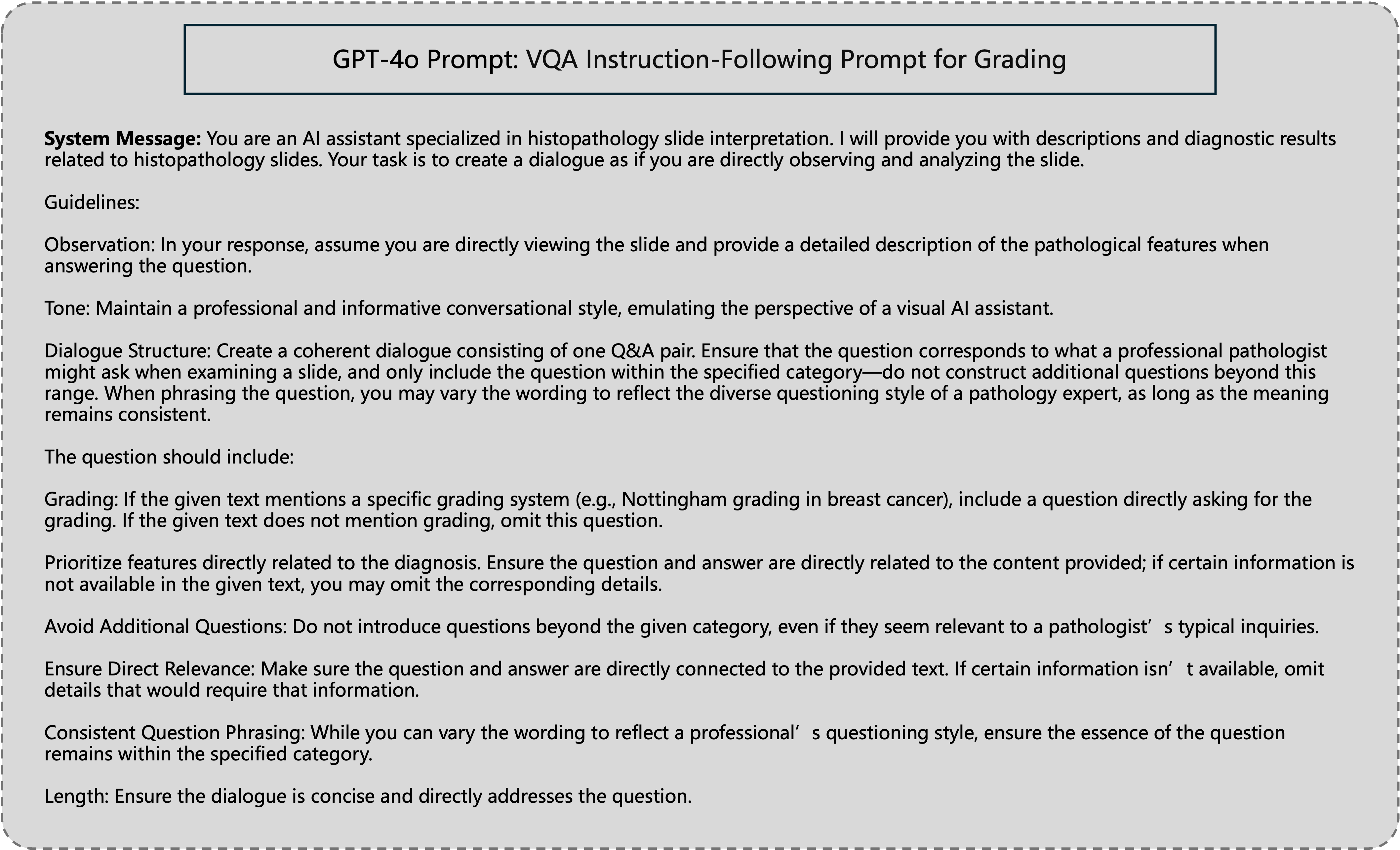} 
  \caption{The prompt used to generate VQA pair for grading task.}
  \label{fig:prompt_grading}
\end{figure*}

\begin{figure*}[ht]
  \centering
  \begin{minipage}{\textwidth}
    \centering
    \includegraphics[width=1\linewidth]{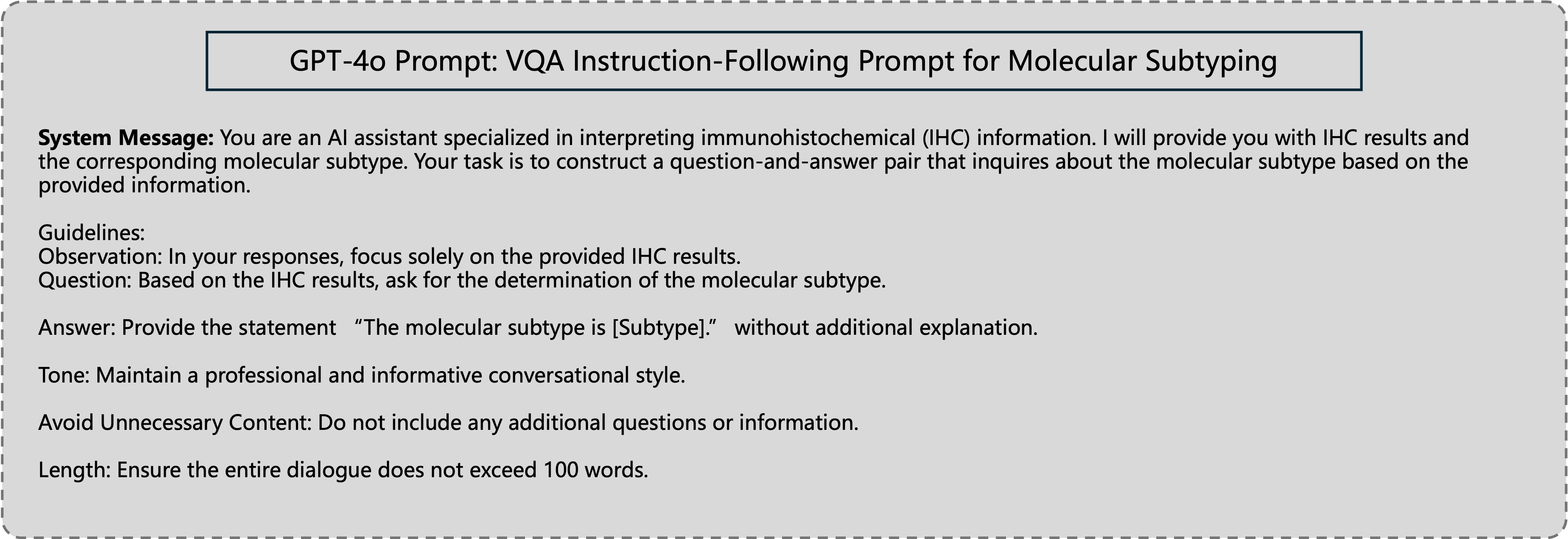}
    \caption{The prompt used to generate VQA pair for molecular subtyping}
    \label{fig:prompt_molecular_subtyping}
  \end{minipage}
  
  \vspace{1em} 
  
  \begin{minipage}{\textwidth}
    \scriptsize
    \centering
    \captionof{table}{VQA example for molecular subtyping task.}
    \begin{tabular}{>{\raggedright\arraybackslash}m{0.2\linewidth} >{\raggedright\arraybackslash}m{0.7\linewidth}}
    \toprule[1.5pt]  
    \multicolumn{2}{l}{\textbf{Example of Molecular Subtyping}} \\
    \midrule
    \textbf{WSI} & \includegraphics[width=0.3\linewidth]{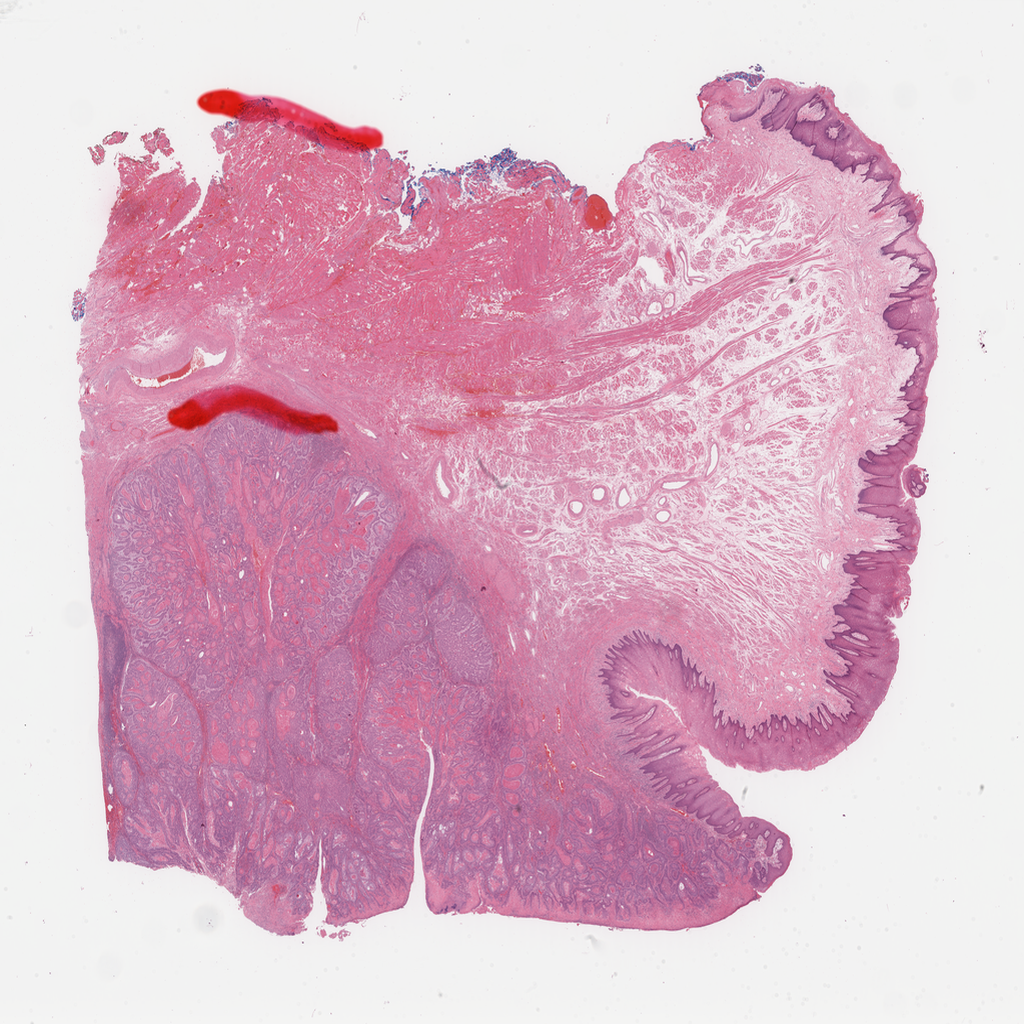} \\ \\
    \textbf{Question} & Based on the IHC results showing P16 negativity and HR, HPV ISH status negativity, what is the molecular subtype in this head and neck squamous cell carcinoma? \\
    \midrule
    \textbf{Answer} & The molecular subtype is HNSC HPV- \\ 
    \bottomrule[1.5pt]
    \end{tabular}
    \label{tab:molecular_subtyping_task_example}
  \end{minipage}
\end{figure*}

\begin{figure*}[ht]
  \centering
  \begin{minipage}{\textwidth}
    \centering
    \includegraphics[width=1\linewidth]{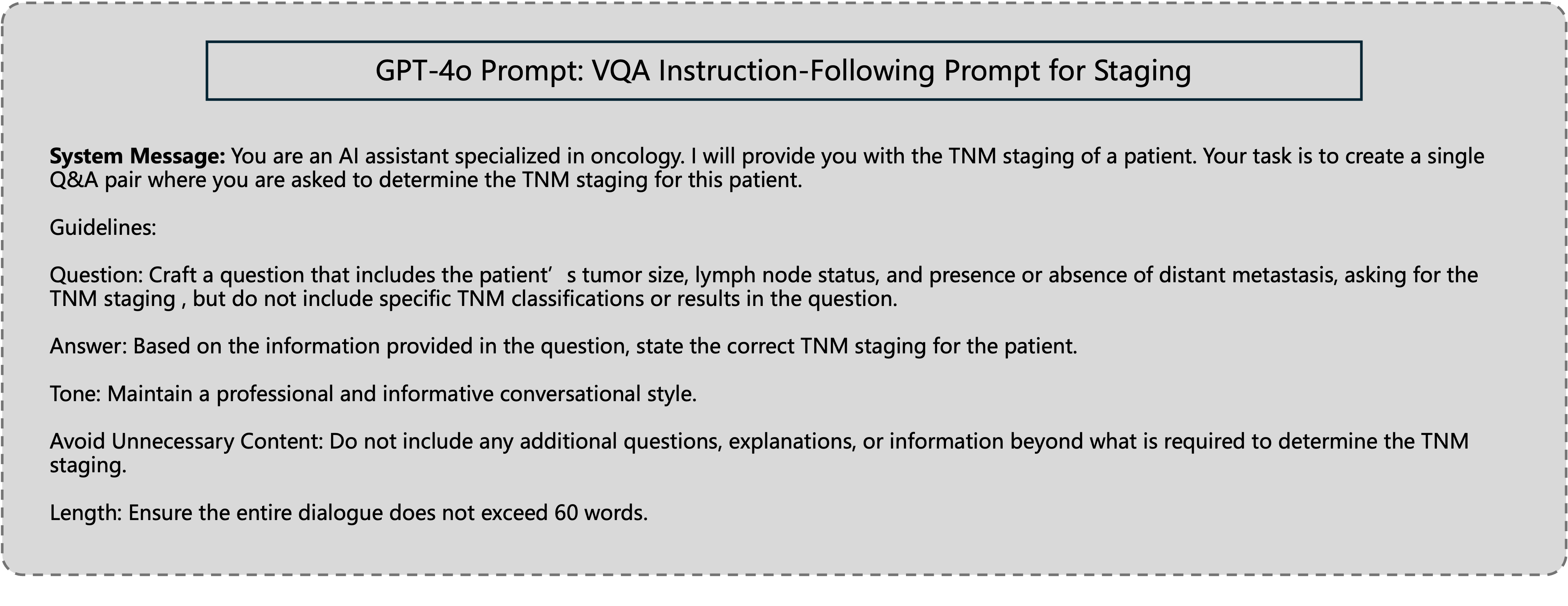}
    \caption{The prompt used to generate VQA pair for Staging}
    \label{fig:prompt_staging}
  \end{minipage}
  
  \vspace{1em} 
  
  \begin{minipage}{\textwidth}
    \scriptsize
    \centering
    \captionof{table}{VQA example for staging task.}
    \begin{tabular}{>{\raggedright\arraybackslash}m{0.2\linewidth} >{\raggedright\arraybackslash}m{0.7\linewidth}}
    \toprule[1.5pt]  
    \multicolumn{2}{l}{\textbf{Example of Staging}} \\
    \midrule
    \textbf{WSI} & \includegraphics[width=0.3\linewidth]{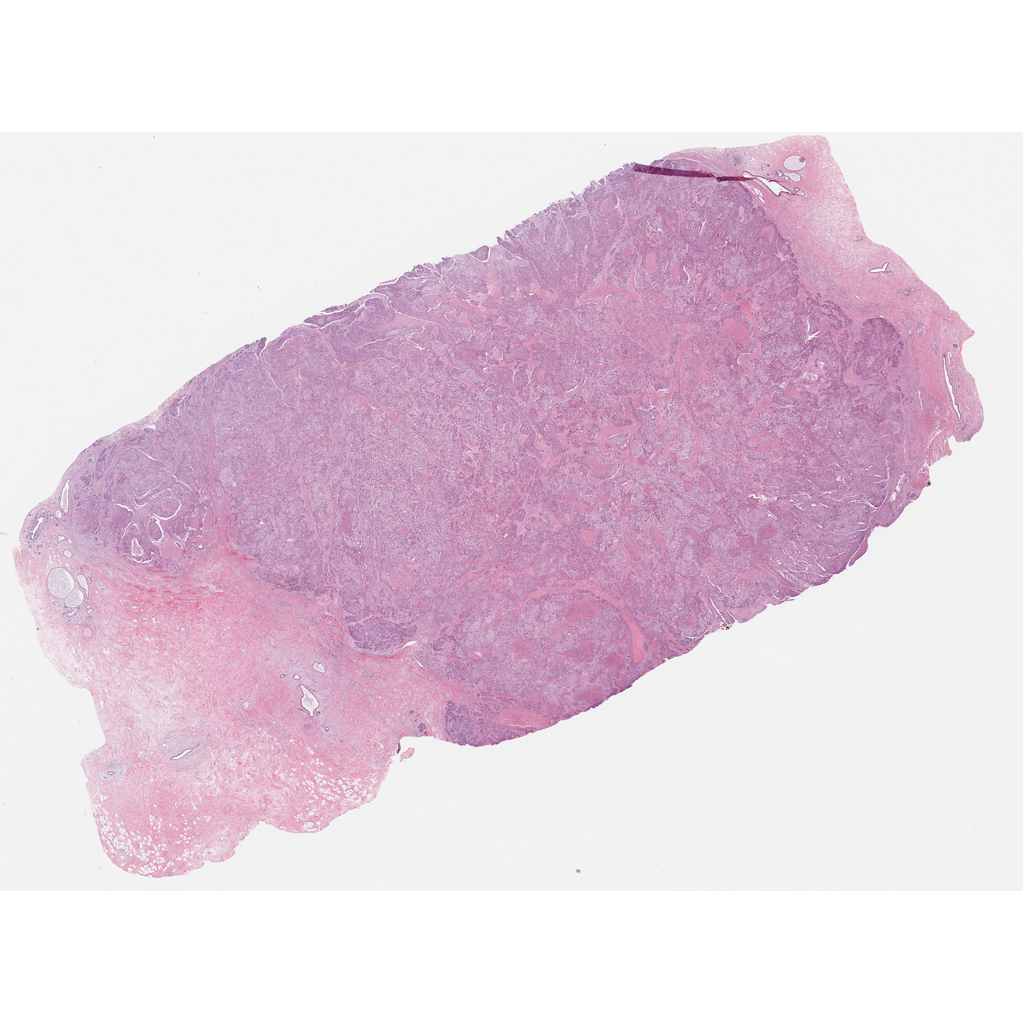} \\ \\
    \textbf{Question} & A patient has a 3 cm tumor, no regional lymph node involvement, and no distant metastasis. What is the TNM staging for this patient? \\
    \midrule
    \textbf{Answer} & The TNM staging for this patient is T2 N0 M0. \\ 
    \bottomrule[1.5pt]
    \end{tabular}
    \label{tab:staging_task_example}
  \end{minipage}
\end{figure*}

\begin{figure*}[ht]
  \centering
  \begin{minipage}{\textwidth}
    \centering
    \includegraphics[width=1\linewidth]{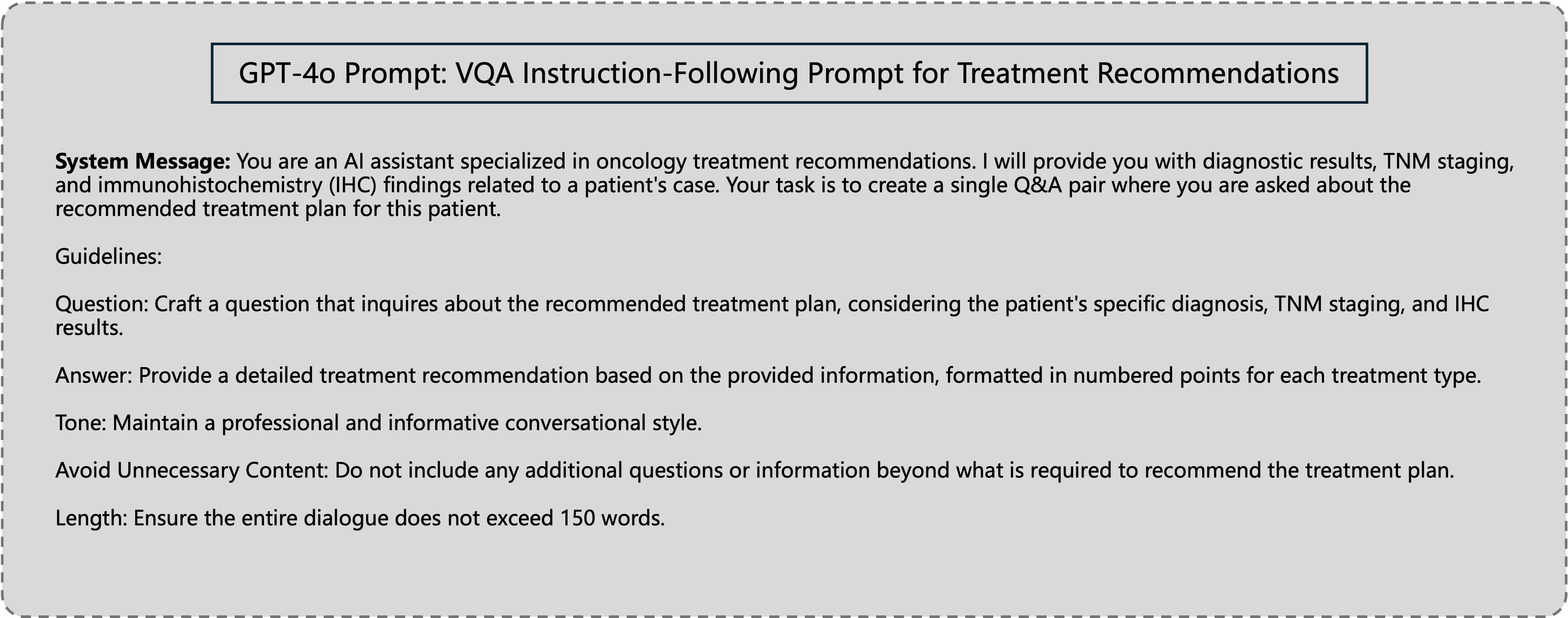}
    \caption{The prompt used to generate VQA pair for treatment recommendations.}
    \label{fig:prompt_treatment_recommendations}
  \end{minipage}
  
  \vspace{1em} 
  
  \begin{minipage}{\textwidth}
    \scriptsize
    \centering
    \captionof{table}{VQA example for treatment recommendations task.}
    \begin{tabular}{>{\raggedright\arraybackslash}m{0.2\linewidth} >{\raggedright\arraybackslash}m{0.7\linewidth}}
    \toprule[1.5pt]  
    \multicolumn{2}{l}{\textbf{VQA example of treatment recommendations task}} \\
    \midrule
    \textbf{WSI} & \includegraphics[width=0.3\linewidth]{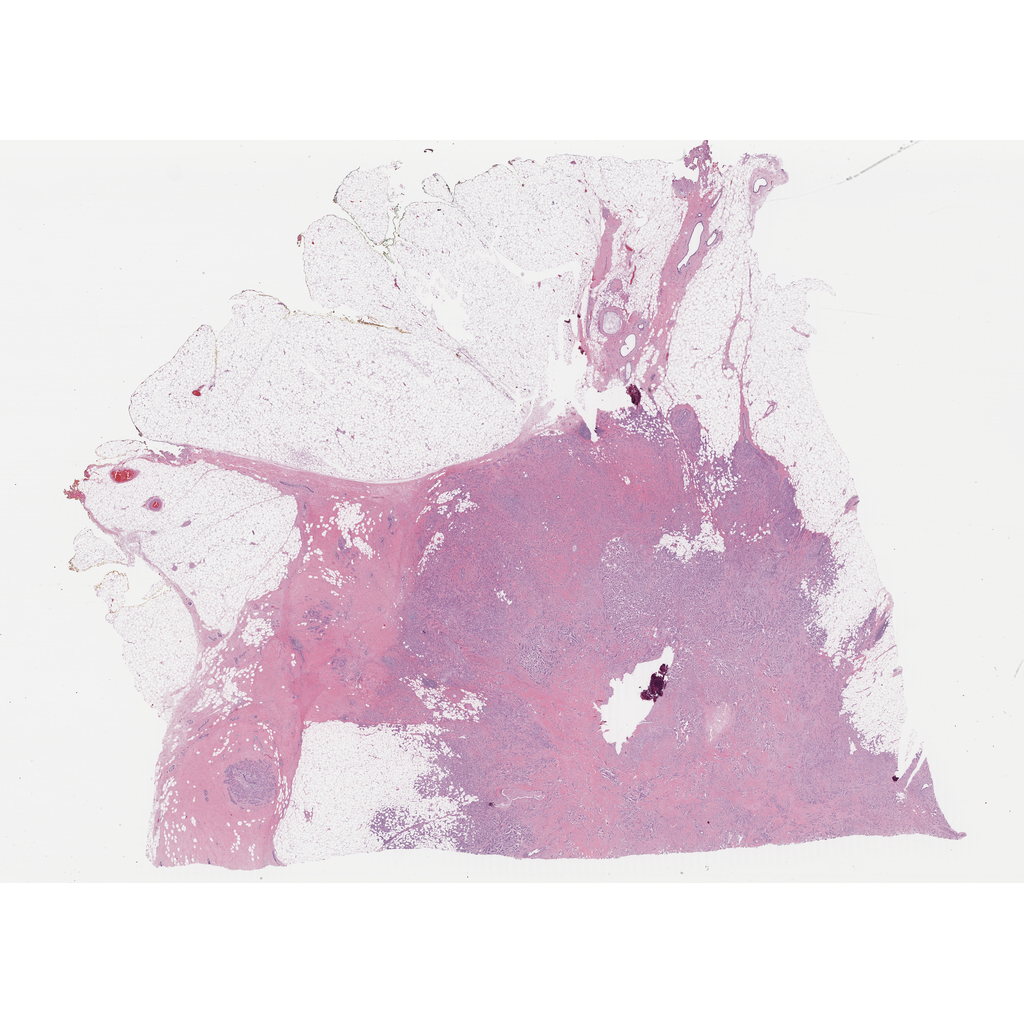} \\ \\
    \textbf{Question} & What is the recommended treatment plan for a patient with invasive ductal carcinoma, Nottingham grade 2, TNM staging T2N1M0, ER-positive, PR-positive, and HER2-negative? \\
    \midrule
    \textbf{Answer} & 
    1. Surgery: Perform breast-conserving surgery or mastectomy, depending on tumor size and patient preference, followed by sentinel lymph node biopsy or axillary lymph node dissection. 
    
    2. Chemotherapy: Consider adjuvant chemotherapy due to nodal involvement and to reduce recurrence risk. 
    
    3. Endocrine Therapy: Initiate tamoxifen or an aromatase inhibitor for at least 5 years, given the ER/PR positivity. 
    
    4. Radiation Therapy: Advise adjuvant radiation therapy, especially if breast-conserving surgery was performed, to minimize local recurrence risk. \\ 
    \bottomrule[1.5pt]
    \end{tabular}
    \label{tab:treatment_recommendations_task_example}
  \end{minipage}
\end{figure*}

\begin{figure*}[htbp]
  \centering
  \begin{minipage}{\textwidth}
    \centering
    \includegraphics[width=1\linewidth]{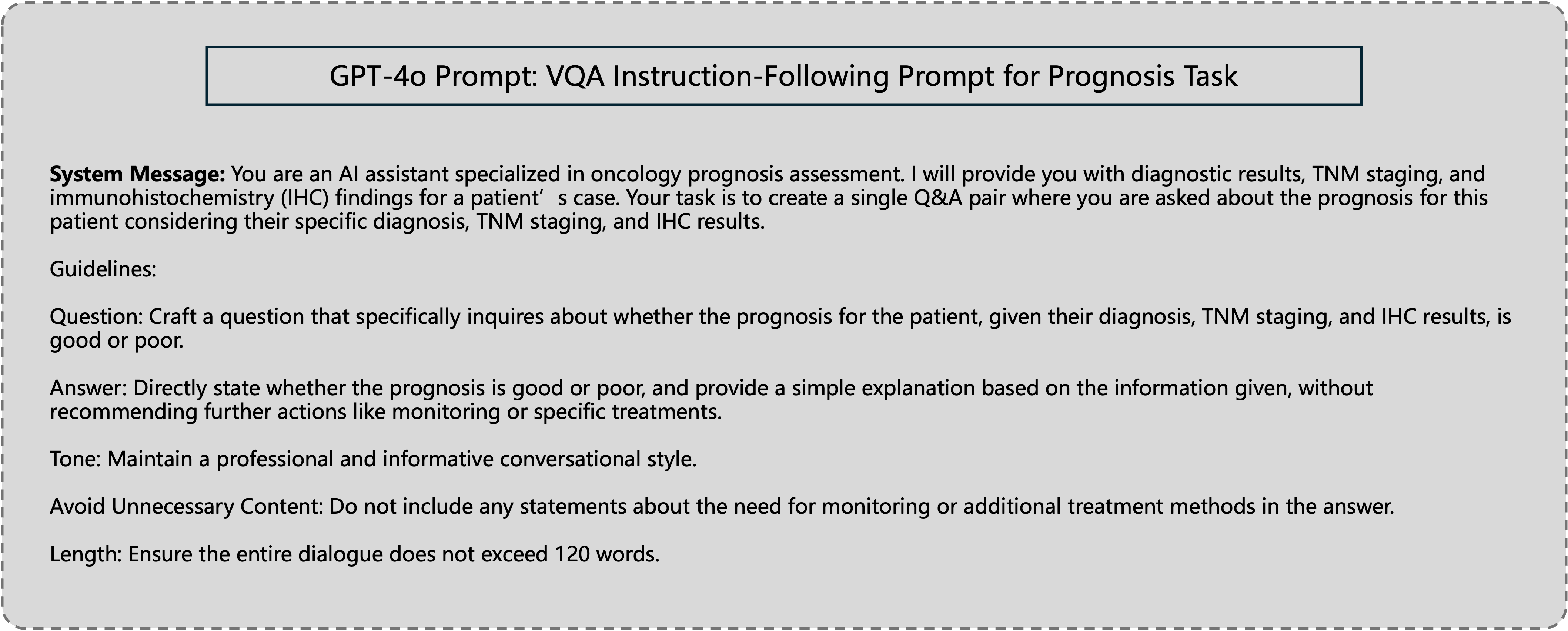}
    \caption{The prompt used to generate VQA pair for prognosis task.}
    \label{fig:prompt_prognosis}
  \end{minipage}
  
  \vspace{1em} 
  
  \begin{minipage}{\textwidth}
    \scriptsize
    \centering
     \captionof{table}{VQA example for prognosis task.}
    \begin{tabular}{>{\raggedright\arraybackslash}m{0.2\linewidth} >{\raggedright\arraybackslash}m{0.7\linewidth}}
    \toprule[1.5pt]  
    \multicolumn{2}{l}{\textbf{Example of Prognosis}} \\
    \midrule
    \textbf{WSI} & \includegraphics[width=0.3\linewidth]{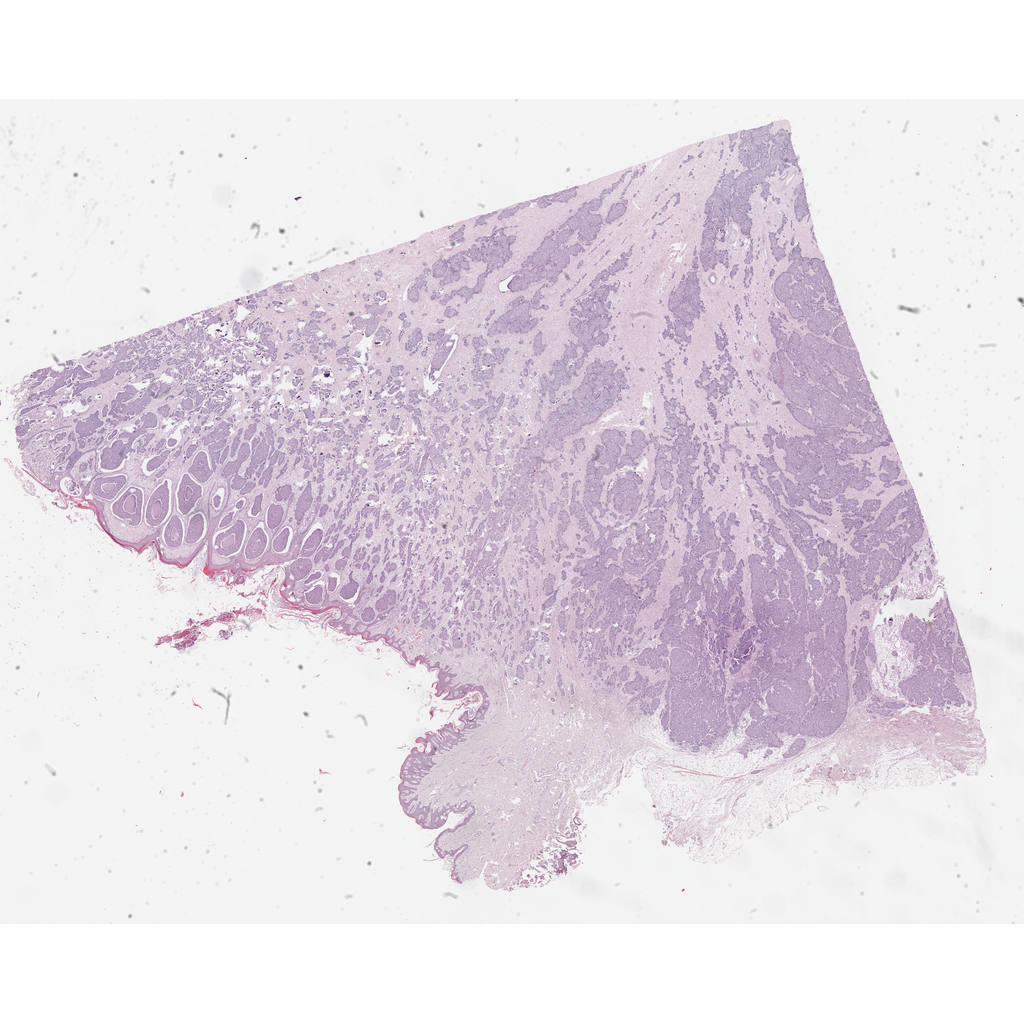} \\ \\
    \textbf{Question} & Given the diagnosis of invasive ductal carcinoma (Nottingham Grade 3), TNM staging of T4BN1M0, and IHC results showing 10-75\% positivity for hormone receptors with negative HER2 protein, is the prognosis good or poor? \\
    \midrule
    \textbf{Answer} & The prognosis is poor. The high histologic grade, advanced tumor size, and lymph node involvement indicate aggressive disease, despite partial hormone receptor positivity. Negative HER2 limits targeted therapy options. \\ 
    \bottomrule[1.5pt]
    \end{tabular}
    \label{tab:prognosis_task_example}
  \end{minipage}
\end{figure*}

\noindent\textbf{Other Pathological Tasks.} We present detailed examples for other pathological tasks showcasing how each prompt is crafted to elicit precise responses from the model, aligning with clinical expectations and diagnostic standards. The prompts, referenced in Fig.~\ref{fig:prompt_regional_structure_description} to \ref{fig:prompt_prognosis}, are pivotal in defining the scope and focus of each pathological task. Tables \ref{tab:regional_structure_description_example} to \ref{tab:prognosis_task_example} visualize the examples for each task.






\subsection{Expert Review and Quality Control}

To ensure medical accuracy, clinical applicability, and educational effectiveness of the generated VQA pairs, we employed a structured expert review process with a proportional sampling strategy due to the extensive dataset size. The expert panel comprised two senior pathologists and five junior physicians with diverse clinical experience.

The detailed quality control workflow was systematically organized as follows:

\begin{itemize}
    \item \textbf{Sampling Strategy:} For open-ended VQA pairs, we randomly selected 10\% of the generated pairs for thorough expert review. This sampling aimed to statistically ensure at least 90\% accuracy across all pairs. For close-ended VQA pairs, we randomly reviewed 5\% to verify their quality.

    \item \textbf{Medical Accuracy Check:} Senior pathologists rigorously validated the sampled pairs by cross-checking the diagnostic accuracy and alignment between answers and visual evidence presented in Whole Slide Images (WSIs), strictly adhering to established clinical guidelines and diagnostic criteria.

    \item \textbf{Clinical Relevance Evaluation:} Both senior and junior reviewers assessed the clinical practicality of sampled VQA pairs, ensuring that they reflected realistic clinical scenarios frequently encountered in pathological diagnosis.

    \item \textbf{Educational Clarity Review:} Junior physicians specifically evaluated whether the VQA pairs were clearly phrased, comprehensible, and effective as educational resources suitable for learners at varying stages of medical training.
\end{itemize}

If significant inaccuracies or issues were identified during the review, prompts were adjusted accordingly, and affected VQA pairs were regenerated. For close-ended pairs, we randomly reviewed 5\% using the same rigorous criteria. This structured sampling and iterative review ensured high-quality, clinically accurate, and educationally valuable VQA pairs across the entire dataset.

\section{WSI Metrics}
\label{appendix:wsi_metric}
We introduce two key metrics developed to evaluate WSI MLLMs: \textbf{\textit{WSI-Precision}} and \textbf{\textit{WSI-Relevance}}. These metrics are essential for assessing the accuracy and relevance of model-generated responses in a clinical context.

\begin{figure*}
  \centering
  \includegraphics[width=1\linewidth]{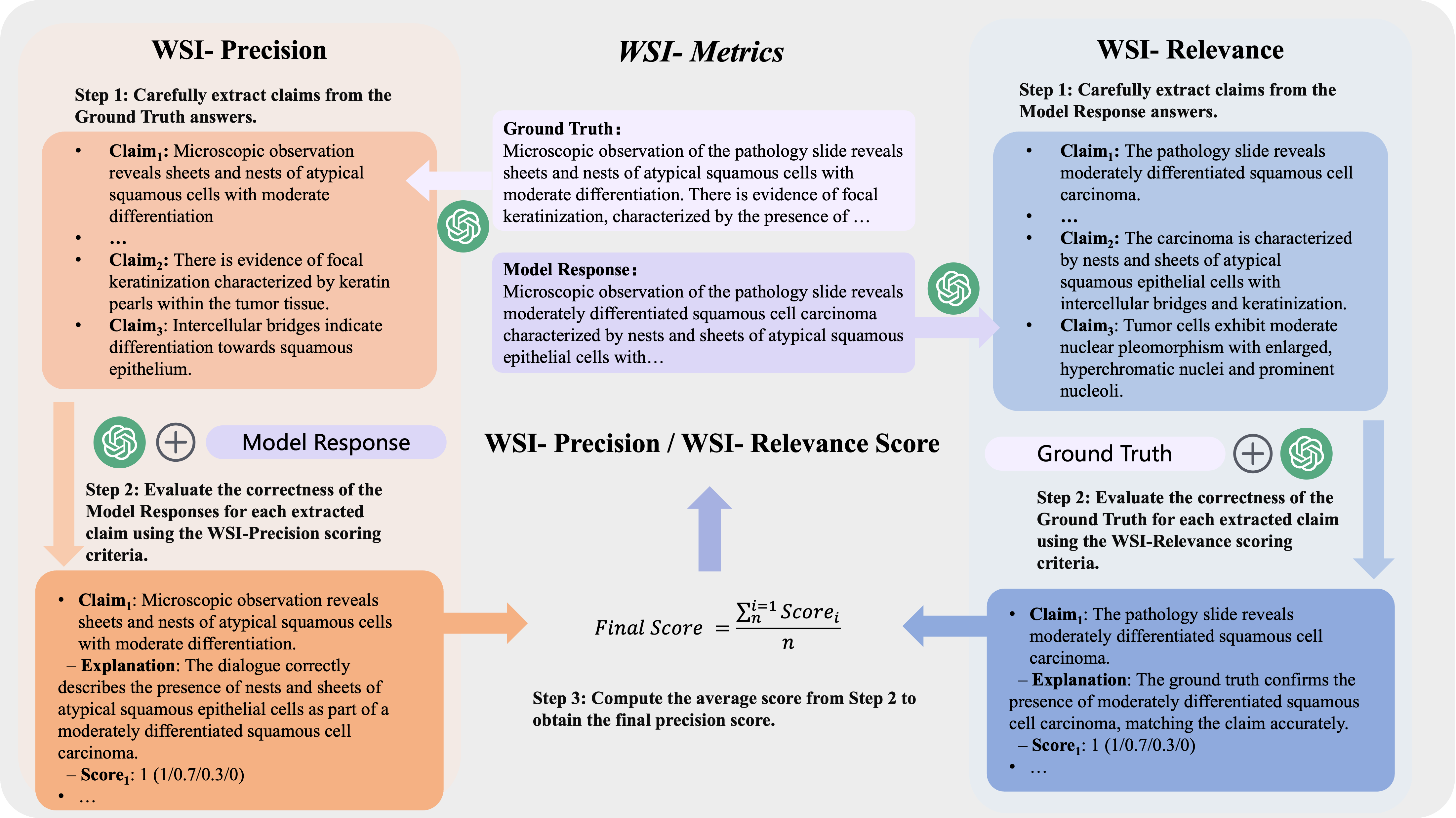}
  \caption{Evaluation workflow for WSI-Precision and WSI-Relevance metrics}
  \label{fig:wsi_metrics}
\end{figure*}

\subsection{WSI-Precision}
WSI-Precision measures the accuracy of model responses by comparing them directly to claims derived from the ground-truth response. As shown in Table~\ref{tab:stepwise_process_precision}, it includes three main steps to compute the evaluation result for the model response and ground-truth response, including the clams extraction, individual score calculation, and final score calculation.

\noindent\textbf{Claims Extraction.}
To break down complex medical information into assessable units, we extract distinct claims from the ground-truth response by using GPT-4o. As shown in Fig.~\ref{fig:claims-extraction-prompt}, we design the prompt to guide GPT-4o for claims extraction. The second row in Table~\ref{tab:stepwise_process_precision} shows the extracted claims for the ground-truth response.

\begin{figure*}[ht]
  \centering
  \includegraphics[width=1\linewidth]{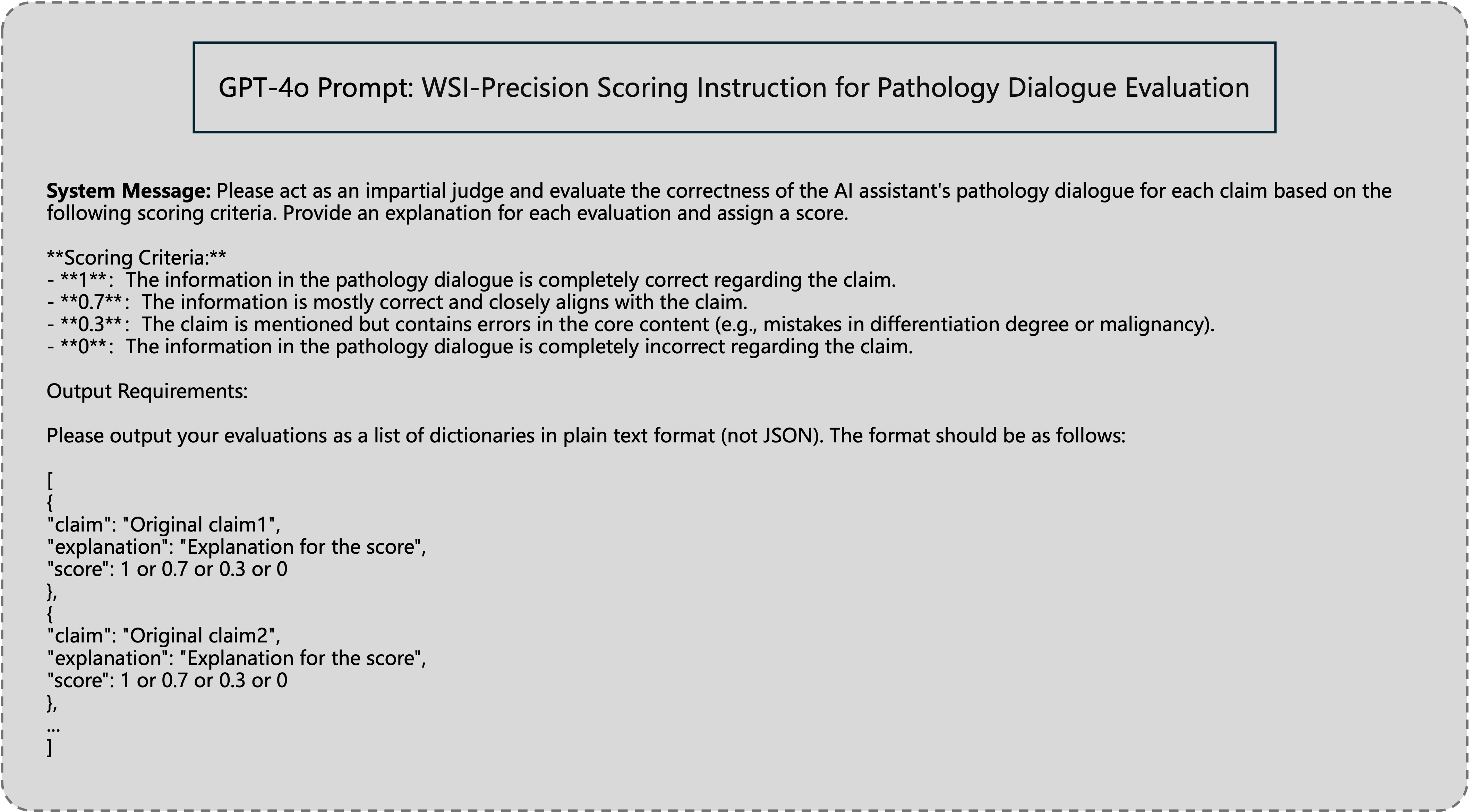} 
  \caption{The prompt for the WSI-Precision metric, outlining scoring criteria for assessing the correctness of model responses against extracted claims.}
  \label{fig:wsi-precision-prompt}
\end{figure*}

\noindent\textbf{Individual Score Calculation.}
With the extracted claims, we compare them with the generated response using our scoring criterion, which is defined as follows:
\begin{itemize}
\item \textbf{Score 1:} The information in the model’s response is completely accurate with respect to the claim. The response perfectly aligns with established facts, correctly addressing all elements of the claim.
\item \textbf{Score 0.7:} The information is mostly correct and closely aligns with the claim, with minor discrepancies that do not fundamentally alter the truth or clinical significance of the response.
\item \textbf{Score 0.3:} The claim is acknowledged, but the response contains significant errors in critical content areas, such as the degree of differentiation or malignancy status, affecting the core message.
\item \textbf{Score 0:} The information in the model’s response is completely incorrect or unrelated to the claim, indicating a fundamental misunderstanding or misrepresentation of the clinical facts.
\end{itemize}
Following this scoring criterion, we design the prompt to guide GPT-4o to compute the individual score for each claim, as demonstrated in Fig.~\ref{fig:wsi-precision-prompt}. In Table~\ref{tab:stepwise_process_precision}, the third row shows the score of each claim.

\noindent\textbf{Final Score Calculation.}
With the individual score for each claim, we calculate the overall WSI-Precision score by averaging all the individual claim scores.

\subsection{WSI-Relevance}
WSI-Relevance assesses the relevance of the claims extracted from model responses in relation to the ground-truth response. Similarly, it also includes three main steps to evaluate the model response and ground-truth response, including the clams extraction, individual score calculation, and final score calculation, as shown in Table~\ref{tab:stepwise_process_relevance}.

\begin{figure*}[ht]
  \centering
  \includegraphics[width=1\linewidth]{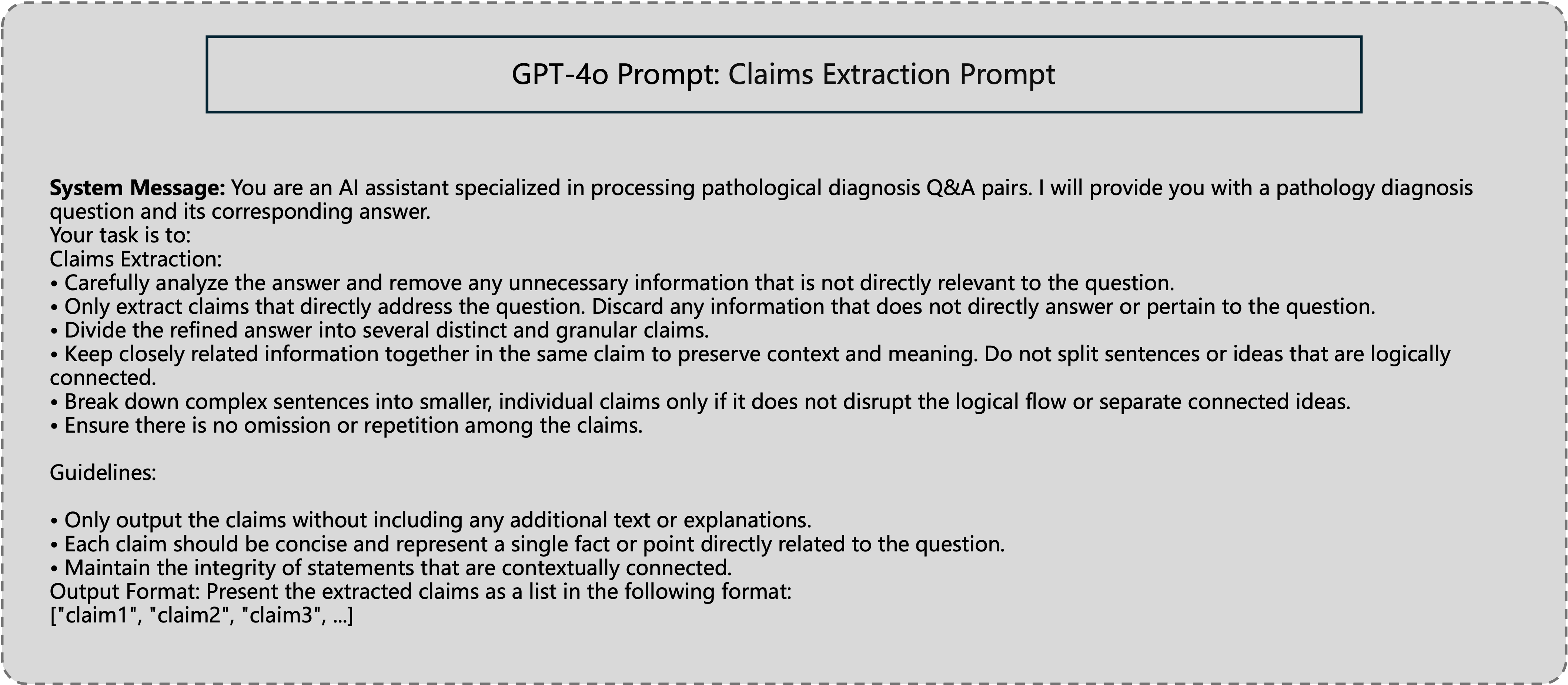} 
  \caption{The prompt for the claims extraction. It provides detailed guidelines for GPT-4o, to identify and refine relevant information from context into distinct claims. These claims serve as the basis for subsequent evaluations.}
  \label{fig:claims-extraction-prompt}
\end{figure*}

\noindent\textbf{Claims Extraction.}
To extract the claims from the model response, we design the prompt to guide GPT-4o for claims extraction, as illustrated in Fig.~\ref{fig:claims-extraction-prompt}. Table~\ref{tab:stepwise_process_relevance} visualizes the extracted claims of model response in second row.

\begin{figure*}[ht]
  \centering
  \includegraphics[width=1\linewidth]{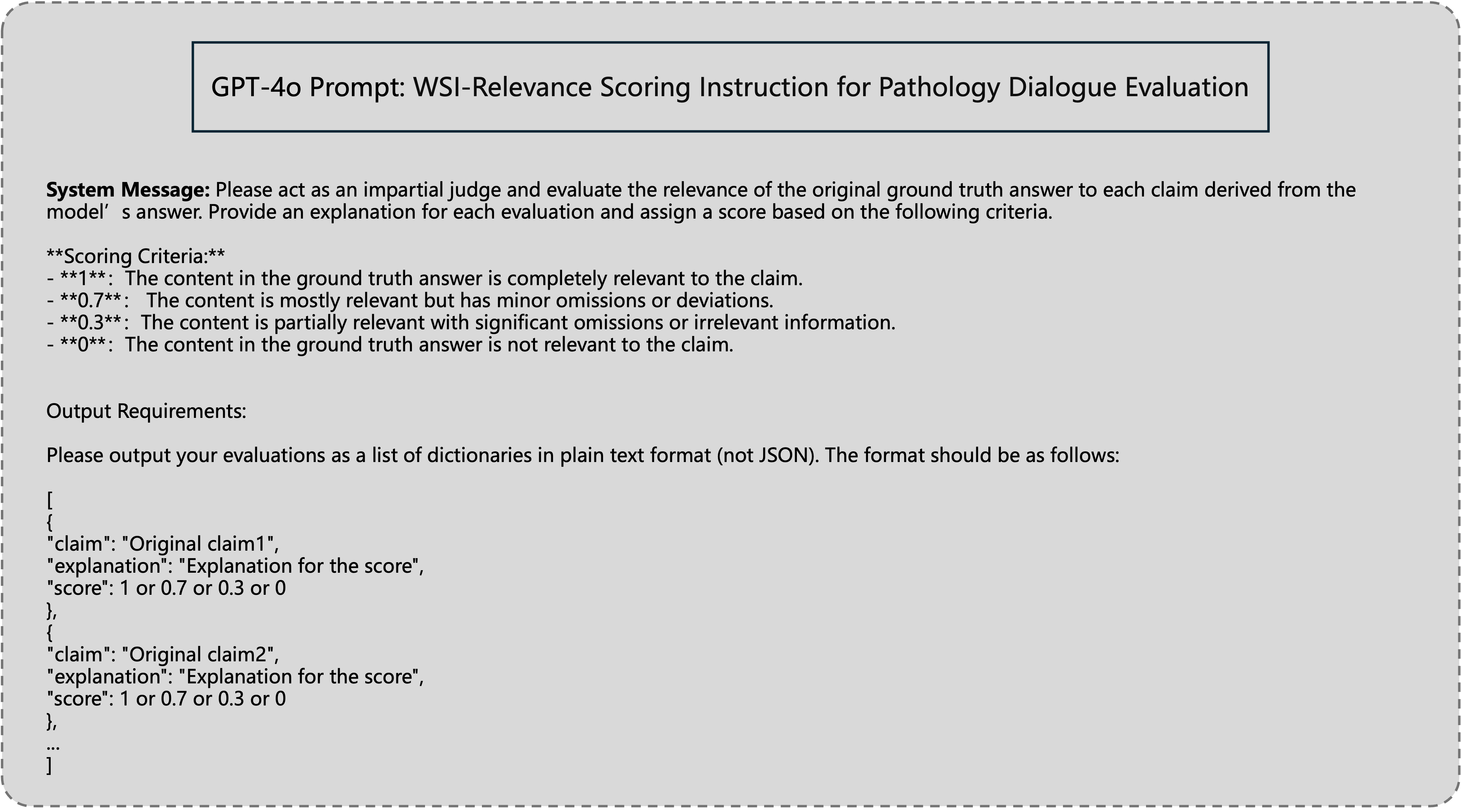} 
  \caption{The prompt for the WSI-Relevance metric, detailing scoring criteria for evaluating the relevance of ground-truth response to claims derived from model responses.}
  \label{fig:wsi-relevance-prompt}
\end{figure*}

\noindent\textbf{Individual Score Calculation.}
With the extracted claims, we compare them with the ground-truth response using our scoring criterion which is the same as that for WSI-Precision. Based on the scoring criterion, we devise the prompt to guide GPT-4o to compute the individual score for each claim, as shown in Fig.~\ref{fig:wsi-relevance-prompt}.

\noindent\textbf{Final Score Calculation.}
With the individual scores for each claim, we average them and obtain the final WSI-Relevance score.

By utilizing both WSI-Precision and WSI-Relevance, we aim to provide a comprehensive assessment of the diagnostic accuracy and clinical relevance of model outputs. This structured evaluation ensures that models are not only producing factually correct responses but are also providing information that is pertinent and useful in real-world medical settings.

\subsection{Evidence for WSI Metrics Validation}
\label{appendix:wsi_metrics_validation}
To rigorously validate our proposed WSI-specific metrics (WSI-Precision and WSI-Relevance), we conducted a systematic sampling-based evaluation comparing automated metric scores against human expert judgments. Due to the large volume of data in our test set (124,011 open-ended responses), we randomly sampled 2\% of the test set from each pathological capability (morphological analysis, diagnosis, treatment planning, and report generation), resulting in 2488 representative WSI-LLaVA model responses for evaluation. For automated evaluation, we tested multiple large language models including GPT-4o, GPT-4o-mini, Claude-3.5-Sonnet, and Gemini-1.5-flash. Considering cost, response speed, and consistency, we ultimately selected GPT-4o. While Claude-3.5-Sonnet demonstrated slightly better performance, it was more expensive and slower than GPT-4o. GPT-4o-mini and Gemini-1.5-flash were faster and more cost-effective but provided lower quality evaluations. A panel of five medical doctors with extensive clinical experience independently assessed each response for precision  and relevance using the same four-point scoring system employed by our automated metrics.

The comparative analysis demonstrated strong alignment between automated WSI metrics and human expert evaluation, with a Pearson correlation of 0.87 for WSI-Precision and 0.82 for WSI-Relevance. Table~\ref{tab:wsi_metrics_validation} shows the average performance of different models compared to human expert evaluation across all evaluated samples. As shown in the table, expert scores were consistently slightly higher than automated metrics, suggesting that automated metrics are marginally more conservative.

\begin{table}[h]
\centering
\begin{tabular}{lcccc}
\hline
\textbf{Model} & \textbf{WSI-Precision} & \textbf{WSI-Relevance} \\
\hline
GPT-4o & 0.572 & 0.624 \\
GPT-4o-mini & 0.539 & 0.602 \\
Claude-3.5-Sonnet & 0.580 & 0.631 \\
Gemini-1.5-flash & 0.535 & 0.597 \\
\hline
\textbf{Human} & 0.636 & 0.689 \\
\hline
\end{tabular}
\caption{Comparison of LLM-based evaluation with human expert evaluation.}
\label{tab:wsi_metrics_validation}
\end{table}

The inter-rater reliability between pathologists showed a Cohen's kappa of 0.78, indicating substantial agreement. Discrepancies between automated and human evaluations were most pronounced in cases involving complex morphological descriptions requiring subtle interpretations, instances where clinical context significantly influenced relevance judgment, and cases where semantic variations in terminology had different clinical implications.

Our sampling-based validation study also revealed several key advantages of WSI metrics over traditional NLU metrics: (1) focus on clinical significance rather than linguistic similarity, (2) interpretability through the claims-based approach, (3) superior contextual understanding within specialized pathology domains, and (4) robustness to stylistic variations. These findings strongly support the validity of our proposed WSI metrics as clinically relevant evaluation measures for assessing model performance in whole slide image analysis tasks.

\section{WSI-LLaVA} 
\label{appendix:implement}
\subsection{Network Architecture}
Our model's network architecture is divided into three components: a WSI encoder, a text encoder, a projection layer, and a large language model (LLM). This design enables efficient processing of gigapixel pathology WSI data.

\noindent\textbf{WSI Encoder.} WSI Encoder, is responsible for extracting meaningful features from gigapixel pathology images, consisting of a patch-level encoder and a slide-level encoder. The patch-level encoder employs Prov-GigaPath\cite{xu2024whole}'s tile-level encoder, which is specifically designed for pathology data. The tile-level encoder is pre-trained using the DINOv2\cite{oquab2023dinov2} framework on 1.3 billion $256 \times 256$ pathology image tiles derived from over 170,000 WSIs spanning 31 major tissue types. Its task is to encode each $256\times 256$ image tile, extracting efficient feature representations at the patch level. The slide-level encoder is based on the LongNet~\cite{ding2023longnet} architecture. LongNet is an improved Transformer that utilizes dilated self-attention mechanisms to reduce the computational complexity of traditional Transformers from O(n\textsuperscript{2}) to a more manageable range. This enables the model to efficiently aggregate features from tens of thousands or even hundreds of thousands of tiles, capturing global spatial contexts while minimizing computational costs.

\noindent\textbf{Text Encoder.} We use Bio\_ClinicalBERT\cite{alsentzer2019publicly} as our text encoder which is a BERT\cite{devlin2018bert} model pre-trained on biomedical and clinical text to optimize its performance in these domains.

\noindent\textbf{Projection Layer.} To achieve cross-modal alignment, a projection layer is designed between the WSI encoder and the LLM. The projection layer consists of a two-layer multi-layer perceptron (MLP) module, which maps visual features into a feature space compatible with the language model.

\noindent\textbf{Large Language Model.} We use Vicuna-7b-v1.5~\cite{zheng2023judging} as the pre-trained model for the large language model, which is a highly efficient and powerful open-source language model tailored for dialogue generation tasks. Through fine-tuning, Vicuna demonstrates exceptional adaptability to various medical scenarios, generating accurate and natural diagnostic reports.

\subsection{Training Details}
\label{appendix:training_details}
 The training process is divided into three stages, i.e. WSI-text alignment, feature space alignment, and task-specific instruction tuning.

\noindent\textbf{Stage 1: WSI-Text Alignment.} 
The learning rate is set to 0.001, with a batch size of 64. During this stage, only a two-layer MLP projection layer is trained to align WSI and text features, while both the WSI encoder and text encoder remain frozen. The training runs for 50 epochs, with a temperature parameter set to 0.02.

\noindent\textbf{Stage 2: Feature Space Alignment.}  
In this stage, both the WSI encoder and the LLM are frozen, and only the projection layer is trained. The learning rate is set to 1e-3, with 1 training epoch and a global batch size of 256. The maximum sequence length is set to 2048, weight decay is disabled (set to 0), and the warmup ratio is configured as 0.03 to ensure a smooth start to training.

\noindent\textbf{Stage 3: Task-Specific Instruction Tuning.}
In this stage, we freeze the WSI encoder and train the projection layer along with the LLM. The learning rate is set to 2e-5, with 1 training epoch and a global batch size of 128. The maximum sequence length is set to 2,048, weight decay is set to 0, and the warmup ratio is maintained at 0.03. We utilize LoRA (Low-Rank Adaptation) with a rank of 128 and an alpha value of 256 for parameter-efficient fine-tuning, enabling adaptation to the target dataset with minimal additional parameters. To enhance computational efficiency, we use DeepSpeed with ZeRO-3 optimization for distributed training. Additionally, BF16 precision is adopted, and TensorFloat32 is enabled to accelerate matrix operations while maintaining numerical stability.

\begin{table*}[ht]
\renewcommand\arraystretch{1.1}
\centering
\caption{Evaluation process for WSI-Precision. Step 1 involves extracting claims from the ground-truth response, step 2 evaluates the correctness of the model responses against these claims using a defined scoring system, and step 3 calculates the average score from Step 2 as the final WSI-Precision score.}
\scalebox{0.97}{
\begin{tabular}{m{0.9cm} m{6.1cm} m{9cm}}

\toprule
 &{\small \textbf{Model Response}}  &  {\small \textbf{Ground-Truth Response}} \\\midrule
&  {\small Microscopic observation of the pathology slide reveals moderately differentiated squamous cell carcinoma characterized by nests and sheets of atypical squamous epithelial cells with intercellular bridges and keratinization. The tumor cells exhibit moderate nuclear pleomorphism with enlarged, hyperchromatic nuclei and prominent nucleoli. Mitotic figures are present, indicating active cell division. There is no evidence of vascular invasion, necrosis, or calcification within the examined tissue. Based on these observations, the diagnosis is moderately differentiated keratinizing squamous carcinoma.} & {\small Microscopic observation of the pathology slide reveals sheets and nests of atypical squamous cells with moderate differentiation. There is evidence of focal keratinization, characterized by the presence of keratin pearls within the tumor tissue. Intercellular bridges are noted among the squamous cells, indicating differentiation towards squamous epithelium. The cellular morphology shows moderate pleomorphism, with variability in nuclear size and shape, and occasional prominent nucleoli. Mitotic figures are present but not excessively numerous. Vascular invasion is not specified in the observations, and necrosis is not prominently noted. Based on these observations, the diagnosis is squamous cell carcinoma, moderately differentiated.}  \\ \hline
{\small \textbf{Steps}} & {\small \textbf{Descriptions}} & {\small \textbf{Outputs}} \\
\midrule
{\small Step 1} & {\small Carefully extract claims from the Ground Truth answers. This involves identifying distinct pieces of information that are directly relevant to the context, preserving logical connections within related ideas. The claims serve as the foundation for evaluating the model’s responses in subsequent steps.} & 
{\small
\begin{itemize}
    \item Claim$_1$: Microscopic observation reveals sheets and nests of atypical squamous cells with moderate differentiation.
    \item Claim$_2$: There is evidence of focal keratinization characterized by keratin pearls within the tumor tissue.
    \item Claim$_3$: Intercellular bridges indicate differentiation towards squamous epithelium.
    \item Claim$_4$: The cellular morphology shows moderate pleomorphism with variability in nuclear size and shape and occasional prominent nucleoli.
    \item Claim$_5$: Mitotic figures are present but not excessively numerous.
    \item Claim$_6$: Vascular invasion is not specified.
    \item Claim$_7$: Necrosis is not prominently noted.
    \item Claim$_8$: The diagnosis is squamous cell carcinoma, moderately differentiated.
    \vspace{-0.5cm}
\end{itemize}} \\
\midrule
{\small Step 2} & {\small Evaluate the correctness of the Model Responses for each extracted claim using the WSI-Precision scoring criteria. Assign a score to each claim (1, 0.7, 0.3, or 0) based on how accurately the model’s response aligns with the clinical information in the Ground Truth. Provide an explanation for each score and present the evaluations as a list of dictionaries.} &

{\small \begin{itemize}
    \item Claim$_1$: \textbf{Microscopic observation reveals sheets and nests of atypical squamous cells with moderate differentiation.}
        \begin{itemize}
            \item Explanation: The dialogue correctly describes the presence of nests and sheets of atypical squamous epithelial cells as part of a moderately differentiated squamous cell carcinoma.
            \item Score$_1$: \textbf{1}
        \end{itemize}
    \item Claim$_2$: \textbf{There is evidence of focal keratinization characterized by keratin pearls within the tumor tissue.}
        \begin{itemize}
            \item Explanation: The dialogue mentions keratinization but does not specifically mention keratin pearls. However, keratinization typically involves keratin pearls, aligning closely with the claim.
            \item Score$_2$: \textbf{0.7}
        \end{itemize}
    \item \textit{...}
    \item Claim$_6$: \textbf{Vascular invasion is not specified.}
        \begin{itemize}
            \item Explanation: The dialogue explicitly states there is no evidence of vascular invasion, which contradicts the claim that vascular invasion is not specified.
            \item Score$_6$: \textbf{0}
        \end{itemize}
    \item \textit{... (Additional evaluations omitted for brevity) ...}
    \vspace{-0.5cm}
\end{itemize}}\\\midrule
{\small Step 3} & {\small Compute the average score from Step 2 to obtain the final precision score. This score represents the overall correctness of the model’s responses in relation to the extracted claims.} & {\small 
\[
\text{Final Score} = \frac{\sum_{i=1}^n \text{Score}_i}{n}
\]
Where:
\begin{itemize}
    \item \(\text{Score}_i\): The score assigned to the \(i\)-th claim.
    \item \(n\): The total number of evaluated claims.
    \vspace{-0.5cm}
\end{itemize}} \\\bottomrule
\end{tabular}}
\label{tab:stepwise_process_precision}
\end{table*}

\begin{table*}[ht]
\renewcommand\arraystretch{1.1}
\centering
\caption{Evaluation process for WSI-Relevance. Step 1 involves extracting claims from the model response, step 2 assesses the relevance of these claims to the Ground-Truth Response, and step 3 calculates the average relevance score from Step 2 as the final WSI-Relevance score.}
\scalebox{0.97}{
\begin{tabular}{m{0.9cm} m{6.1cm} m{9cm}}
\toprule
 &{\small \textbf{Model Response}}  &  {\small \textbf{Ground-Truth Response}} \\\midrule
&  {\small Microscopic observation of the pathology slide reveals moderately differentiated squamous cell carcinoma characterized by nests and sheets of atypical squamous epithelial cells with intercellular bridges and keratinization. The tumor cells exhibit moderate nuclear pleomorphism with enlarged, hyperchromatic nuclei and prominent nucleoli. Mitotic figures are present, indicating active cell division. There is no evidence of vascular invasion, necrosis, or calcification within the examined tissue. Based on these observations, the diagnosis is moderately differentiated keratinizing squamous carcinoma.} & {\small Microscopic observation of the pathology slide reveals sheets and nests of atypical squamous cells with moderate differentiation. There is evidence of focal keratinization, characterized by the presence of keratin pearls within the tumor tissue. Intercellular bridges are noted among the squamous cells, indicating differentiation towards squamous epithelium. The cellular morphology shows moderate pleomorphism, with variability in nuclear size and shape, and occasional prominent nucleoli. Mitotic figures are present but not excessively numerous. Vascular invasion is not specified in the observations, and necrosis is not prominently noted. Based on these observations, the diagnosis is squamous cell carcinoma, moderately differentiated.}  \\ \hline
{\small \textbf{Steps}} & {\small \textbf{Descriptions}} & {\small \textbf{Outputs}} \\
\midrule
{\small Step 1} & {\small Extract claims from the Model Response, focusing on identifying distinct and relevant pieces of information that form the basis of the pathological assessment. These claims must preserve the logical and clinical coherence of the model's response, serving as the foundation for relevance evaluation in the following steps.} & 
{\small
\begin{itemize}
    \item Claim$_1$: The pathology slide reveals moderately differentiated squamous cell carcinoma.
    \item Claim$_2$: The carcinoma is characterized by nests and sheets of atypical squamous epithelial cells with intercellular bridges and keratinization.
    \item Claim$_3$: Tumor cells exhibit moderate nuclear pleomorphism with enlarged, hyperchromatic nuclei and prominent nucleoli.
    \item Claim$_4$: Mitotic figures are present, indicating active cell division.
    \item Claim$_5$: There is no evidence of vascular invasion, necrosis, or calcification within the examined tissue.
    \item Claim$_6$: The diagnosis is moderately differentiated keratinizing squamous carcinoma.
    \vspace{-0.5cm}
\end{itemize}} \\
\midrule
{\small Step 2} & {\small Assess the relevance of each extracted claim from the Model Response to the Ground-Truth Response. Assign a relevance score (1, 0.7, 0.3, or 0) based on how well each claim reflects the clinical details and context provided in the Ground Truth. Provide a justification for each relevance score, ensuring that the evaluation clearly outlines how each claim aligns or deviates from the ground-truth observations.} &

{\small \begin{itemize}
    \item Claim$_1$: \textbf{The pathology slide reveals moderately differentiated squamous cell carcinoma.}
        \begin{itemize}
            \item Explanation: The ground truth confirms the presence of moderately differentiated squamous cell carcinoma, matching the claim accurately.
            \item Score$_1$: \textbf{1}
        \end{itemize}
    \item Claim$_2$: \textbf{The carcinoma is characterized by nests and sheets of atypical squamous epithelial cells with intercellular bridges and keratinization.}
        \begin{itemize}
            \item Explanation: The ground truth describes similar cellular structures, but the focus on keratinization is less pronounced, leading to a slightly lower score.
            \item Score$_2$: \textbf{0.7}
        \end{itemize}
    \item \textit{...}
    \item Claim$_4$: \textbf{Mitotic figures are present, indicating active cell division.}
        \begin{itemize}
            \item Explanation: The presence of mitotic figures is confirmed, aligning with the claim of active cell division, and is fully supported by the ground truth.
            \item Score$_4$: \textbf{1}
        \end{itemize}
    \item \textit{... (Additional evaluations omitted for brevity) ...}
    \vspace{-0.5cm}
\end{itemize}}\\\midrule
{\small Step 3} & {\small Compute the average score from Step 2 to obtain the final relevance score.This score quantifies how closely the model’s responses align with the key contextual elements of the ground truth.} & {\small 
\[
\text{Final Score} = \frac{\sum_{i=1}^n \text{Score}_i}{n}
\]
Where:
\begin{itemize}
    \item \(\text{Score}_i\): The score assigned to the \(i\)-th claim.
    \item \(n\): The total number of evaluated claims.
    \vspace{-0.5cm}
\end{itemize}} \\\bottomrule
\end{tabular}}
\label{tab:stepwise_process_relevance}
\end{table*}

\begin{table*}[h!]
\centering
\caption{Detailed diagnostic steps of the Nottingham grading system for breast cancer, illustrating the scoring and corresponding morphological descriptions based on gland formation, nuclear pleomorphism, and mitotic activity.}
\begin{tabular}{|c|c|c|}
\hline
\textbf{Diagnostic Step} & \textbf{Score} & \textbf{Morphological Description} \\
\hline
\textbf{Gland Formation} & 
\begin{tabular}[c]{@{}c@{}}1 - Well-formed \\ 2 - Moderately formed \\ 3 - Poorly formed\end{tabular} & 
\begin{tabular}[c]{@{}c@{}}Glands are well-differentiated \\ Glands show moderate differentiation \\ Minimal to no gland formation\end{tabular} \\
\hline
\textbf{Nuclear Pleomorphism} & 
\begin{tabular}[c]{@{}c@{}}1 - Small, uniform \\ 2 - Moderate variability \\ 3 - Marked variability\end{tabular} & 
\begin{tabular}[c]{@{}c@{}}Nuclei are small and uniform \\ Nuclei show moderate variability in size and shape \\ Nuclei show significant variation in size and shape\end{tabular} \\
\hline
\textbf{Mitotic Count} & 
\begin{tabular}[c]{@{}c@{}}Low - Few mitoses \\ Medium - Moderate mitoses \\ High - Numerous mitoses\end{tabular} & 
\begin{tabular}[c]{@{}c@{}}Few mitotic figures are present \\ Moderate number of mitotic figures \\ High number of mitotic figures\end{tabular} \\
\hline
\end{tabular}
\label{tab:nottingham_detailed}
\end{table*}

\clearpage

\begin{table*}[h]
\scriptsize
\centering
\caption{The prompt used to generate VQA pairs for global morphology description task.}
\begin{tabular}{>{\raggedright\arraybackslash}m{0.2\linewidth} >{\raggedright\arraybackslash}m{0.7\linewidth}}
\toprule[1.5pt]  
\multicolumn{2}{l}{\textbf{VQA example of global morphology description}} \\
\midrule
\textbf{WSI} & \includegraphics[width=0.3\linewidth]{sample/TCGA-A2-A0CX-01Z-00-DX1.F07C75AB-E568-45CB-B497-37C712490393.svs_original.png} \\ \\
\textbf{Question} & Can you provide an overall description of the whole slide image (WSI)? \\
\midrule
\textbf{Answer} & The whole slide image reveals malignant epithelial cells infiltrating the surrounding stroma with minimal glandular structures. The normal tissue architecture is significantly disrupted. The tumor cells are poorly differentiated, displaying large variability in nuclear size and shape, indicative of substantial nuclear pleomorphism. A moderate rate of cell division is observed among the tumor cells. There is no evident vascular invasion, necrosis, or microcalcifications. The intraductal component is less than 25\% with notable cancerization present. \\ 
\bottomrule[1.5pt]
\end{tabular}
\label{tab:global_morphology_description_example}
\end{table*}

\begin{table*}[h!]
\scriptsize
\centering
\caption{The prompt used to generate VQA pairs for key diagnostic description task.}
\begin{tabular}{>{\raggedright\arraybackslash}m{0.2\linewidth} >{\raggedright\arraybackslash}m{0.7\linewidth}}
\toprule[1.5pt]  
\multicolumn{2}{l}{\textbf{VQA example for key diagnostic description task}} \\
\midrule
\textbf{WSI} & \includegraphics[width=0.3\linewidth]{sample/TCGA-A2-A0CX-01Z-00-DX1.F07C75AB-E568-45CB-B497-37C712490393.svs_original.png} \\ \\
\textbf{Question} & Among the features observed, which are particularly critical for making a diagnosis? \\
\midrule
\textbf{Answer} & Among the features observed, which are particularly critical for making a diagnosis?", "text": "Key features essential for diagnosis include the extensive nuclear pleomorphism and minimal glandular structure formation, both of which point to poor differentiation. The moderate mitotic rate is also significant in assessing cellular proliferation. Another important factor is the presence of a prominent cancerization of ducts, albeit the intraductal component is less than 25\%. These aspects collectively depict a highly invasive and aggressive nature of the tumor. \\ 
\bottomrule[1.5pt]
\end{tabular}
\label{tab:key_diagnostic_description_example}
\end{table*}

\begin{table*}[h!]
\scriptsize
\centering
\caption{VQA example for histological typing task}
\begin{tabular}{>{\raggedright\arraybackslash}m{0.2\linewidth} >{\raggedright\arraybackslash}m{0.7\linewidth}}
\toprule[1.5pt]  
\multicolumn{2}{l}{\textbf{Example of Histological Typing}} \\
\midrule
\textbf{WSI} & \includegraphics[width=0.3\linewidth]{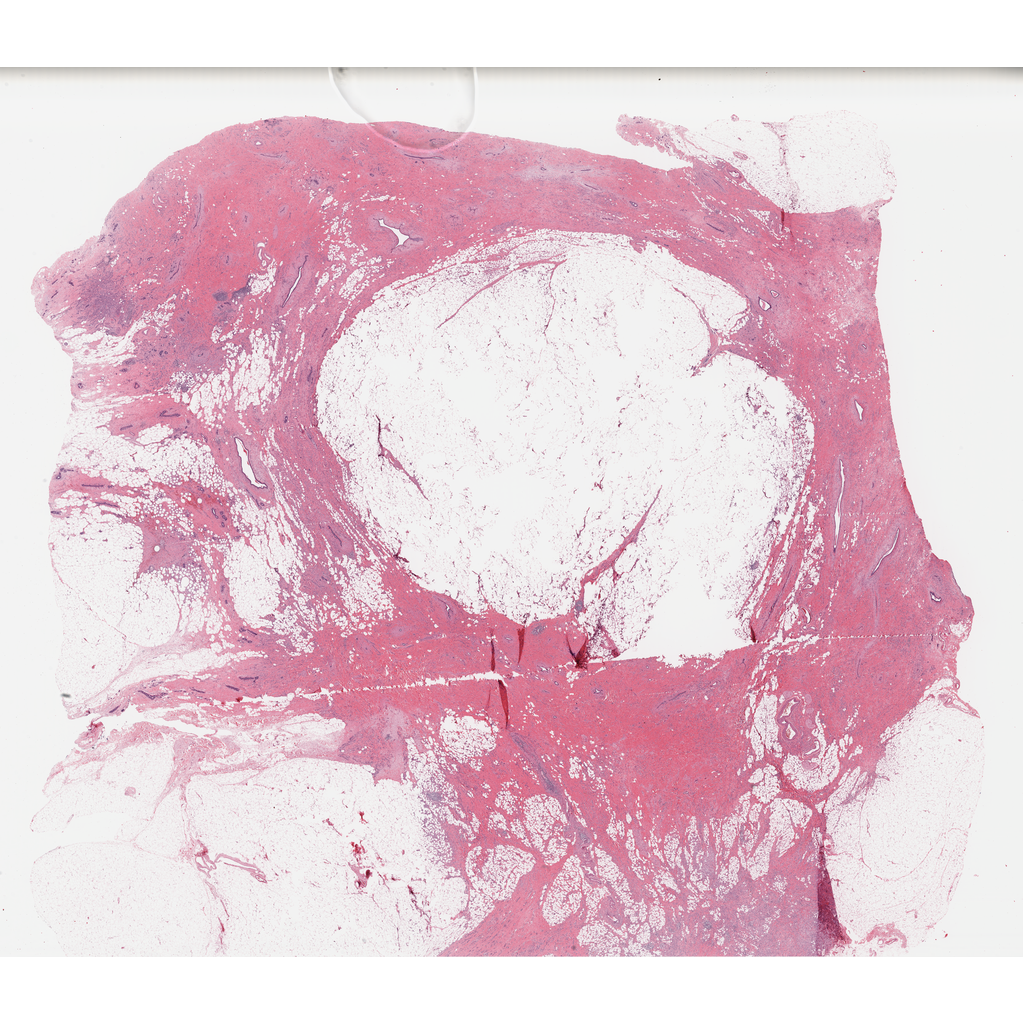} \\ \\
\textbf{Question} & How would you classify the histological type of cancer observed on the slide? \\
\midrule
\textbf{Answer} & The observed slide is indicative of invasive lobular carcinoma. This classification is due to the presence of epithelial cells with minimal glandular differentiation, infiltrating the stroma in a characteristic pattern. The moderate nuclear pleomorphism and the absence of vascular or lymphatic invasion are consistent with this histological type. \\ 
\bottomrule[1.5pt]
\end{tabular}
\label{tab:histological_typing_example}
\end{table*}

\begin{table*}[h!]
\scriptsize
\centering
\caption{VQA example for grading task.}
\begin{tabular}{>{\raggedright\arraybackslash}m{0.2\linewidth} >{\raggedright\arraybackslash}m{0.7\linewidth}}
\toprule[1.5pt]  
\multicolumn{2}{l}{\textbf{Example of Grading}} \\
\midrule
\textbf{WSI} & \includegraphics[width=0.3\linewidth]{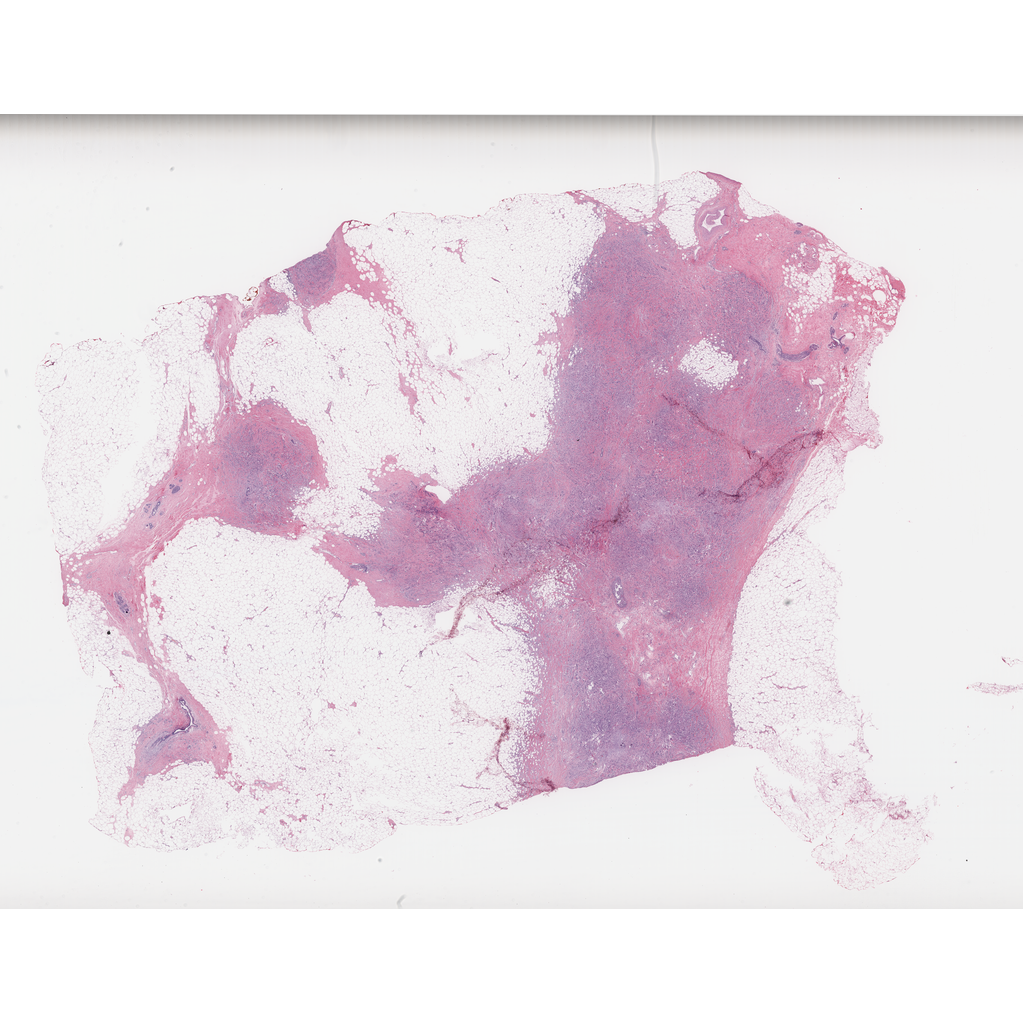} \\ \\
\textbf{Question} & What is the Nottingham Histologic Grade of this tumor based on the observed features? \\
\midrule
\textbf{Answer} & The tumor is classified as Nottingham Histologic Grade 3, with a total score of 8 out of 9. This grading is based on the poor tubule formation (score of 3), marked nuclear pleomorphism (score of 3), and moderate mitotic activity (score of 2). This indicates a high-grade tumor with poor differentiation and suggests a potentially more aggressive clinical course. \\ 
\bottomrule[1.5pt]
\end{tabular}
\label{tab:grading_example}
\end{table*} 

\begin{table*}[h!]
\scriptsize
\centering
\caption{A comparative example of Report Generation outputs from different models}
\begin{tabular}{>{\raggedright\arraybackslash}m{0.2\linewidth} >{\raggedright\arraybackslash}m{0.7\linewidth}}
\toprule[1.5pt]  
\multicolumn{2}{l}{\textbf{Sample Comparison of Report Generation}} \\
\midrule
\textbf{WSI} & \includegraphics[width=0.3\linewidth]{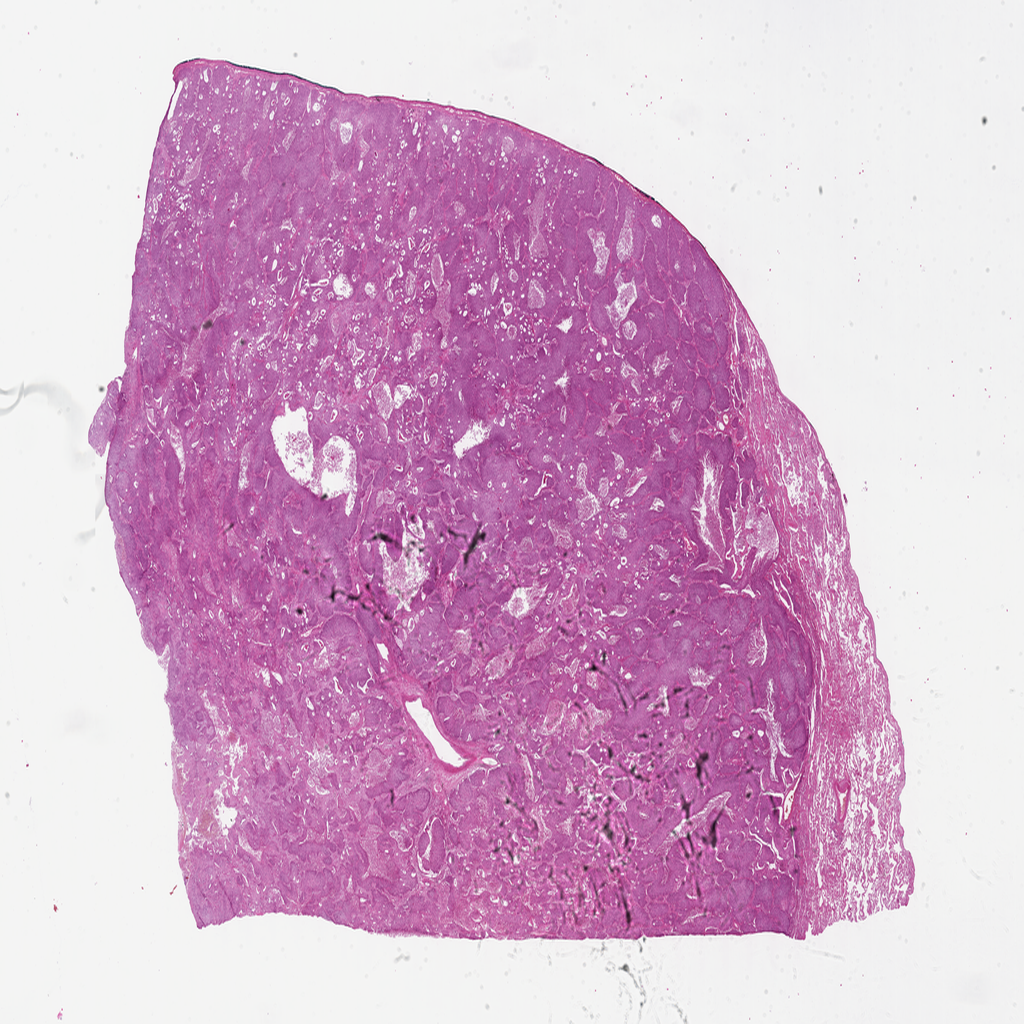} \\ 
\textbf{Question} & What are the main observations in the whole slide histopathology image, and what is the likely diagnosis? \\
\midrule
\textbf{WSI-LLaVA} & Microscopic observation of the pathology slide reveals \textcolor{green!50!black}{moderately differentiated squamous cell carcinoma characterized by nests and sheets of atypical squamous epithelial cells with intercellular bridges and keratinization}. \textcolor{green!50!black}{The tumor cells exhibit moderate nuclear pleomorphism with enlarged, hyperchromatic nuclei and prominent nucleoli. Mitotic figures are present}, \textcolor{red!80!black}{indicating active cell division.} \textcolor{green!50!black}{There is no evidence of vascular invasion, necrosis, or calcification within the examined tissue. Based on these observations, the diagnosis is moderately differentiated keratinizing squamous carcinoma}. \\
\midrule
\textbf{MI-Gen} & microscopic observation of the pathology slide reveals a \textcolor{red!80!black}{poorly} \textcolor{green!50!black}{differentiated squamous cell carcinoma characterized} by a lack of squamous cell carcinoma . the tumor exhibits a lack of squamous differentiation with a lack of squamous differentiation . \textcolor{green!50!black}{the tumor cells exhibit moderate nuclear pleomorphism with variability in nuclear size and shape and there is a moderate rate of cell division .vascular invasion is not identified . necrosis is not observed within the tumor . based on these observations the diagnosis is squamous cell carcinoma} . \\
\midrule
\textbf{Hist-Gen} & microscopic observation of the pathology slide reveals a \textcolor{red!80!black}{poorly } \textcolor{green!50!black}{differentiated squamous cell carcinoma characterized by irregular nests and sheets of atypical squamous cells with intercellular bridges and keratinization .} \textcolor{red!80!black}{the tumor cells exhibit a high degree of nuclear pleomorphism .} \textcolor{green!50!black}{there is no evidence of vascular invasion or necrosis within the tumor tissue .} the surrounding lung parenchyma shows areas of chronic inflammation and chronic inflammation . based on these observations the diagnosis is \textcolor{green!50!black}{squamous cell carcinoma} \textcolor{red!80!black}{poorly differentiated } . \\
\midrule
\textbf{Quilt-LLaVA} & Microscopic observation of the pathology slide reveals\textcolor{green!50!black}{ sheets and nests of atypical squamous cells with moderate differentiation. There is evidence of focal keratinization, characterized by the presence of keratin pearls within the tumor tissue. Intercellular bridges are noted among the squamous cells, indicating differentiation towards squamous epithelium. The cellular morphology shows moderate pleomorphism, with variability in nuclear size and shape, and occasional prominent nucleoli. Mitotic figures are present but not excessively numerous. Vascular invasion is not specified in the observations, and necrosis is not prominently noted. Based on these observations, the diagnosis is squamous cell carcinoma, moderately differentiated.} \\
\midrule
\textbf{GPT-4o} & \textcolor{red!80!black}{This histopathology image shows a well-circumscribed, nodular lesion with a dense, cellular stroma. There are areas of necrosis and hemorrhage, and the tissue architecture appears disrupted. The lesion has a heterogeneous appearance with varying cellular densities and some fibrotic regions.**Diagnosis**: The features are consistent with a uterine fibroid (leiomyoma), which is characterized by a smooth muscle cell growth forming a circumscribed mass with areas of degeneration. }\\
\midrule
\textbf{Ground Truth} & Microscopic observation of the pathology slide reveals sheets and nests of atypical squamous cells with moderate differentiation. There is evidence of focal keratinization, characterized by the presence of keratin pearls within the tumor tissue. Intercellular bridges are noted among the squamous cells, indicating differentiation towards squamous epithelium. The cellular morphology shows moderate pleomorphism, with variability in nuclear size and shape, and occasional prominent nucleoli. Mitotic figures are present but not excessively numerous. Vascular invasion is not specified in the observations, and necrosis is not prominently noted. Based on these observations, the diagnosis is squamous cell carcinoma, moderately differentiated. \\ 

\bottomrule[1.5pt]
\end{tabular}
\label{tab:sample_comparison_report_generation}
\end{table*} 

\begin{table}[h!]
{
\scriptsize
\centering
\caption{A comparative example of Global Morphology Description outputs from different models}
\renewcommand{\arraystretch}{1.5} 
\begin{tabular}{>{\arraybackslash}m{1.3cm} m{6.4cm}}
\toprule[1.5pt]
\multicolumn{2}{l}{\textbf{Sample Comparison of Global Morphology Description}} \\
\midrule
\textbf{WSI} & \includegraphics[width=4cm]{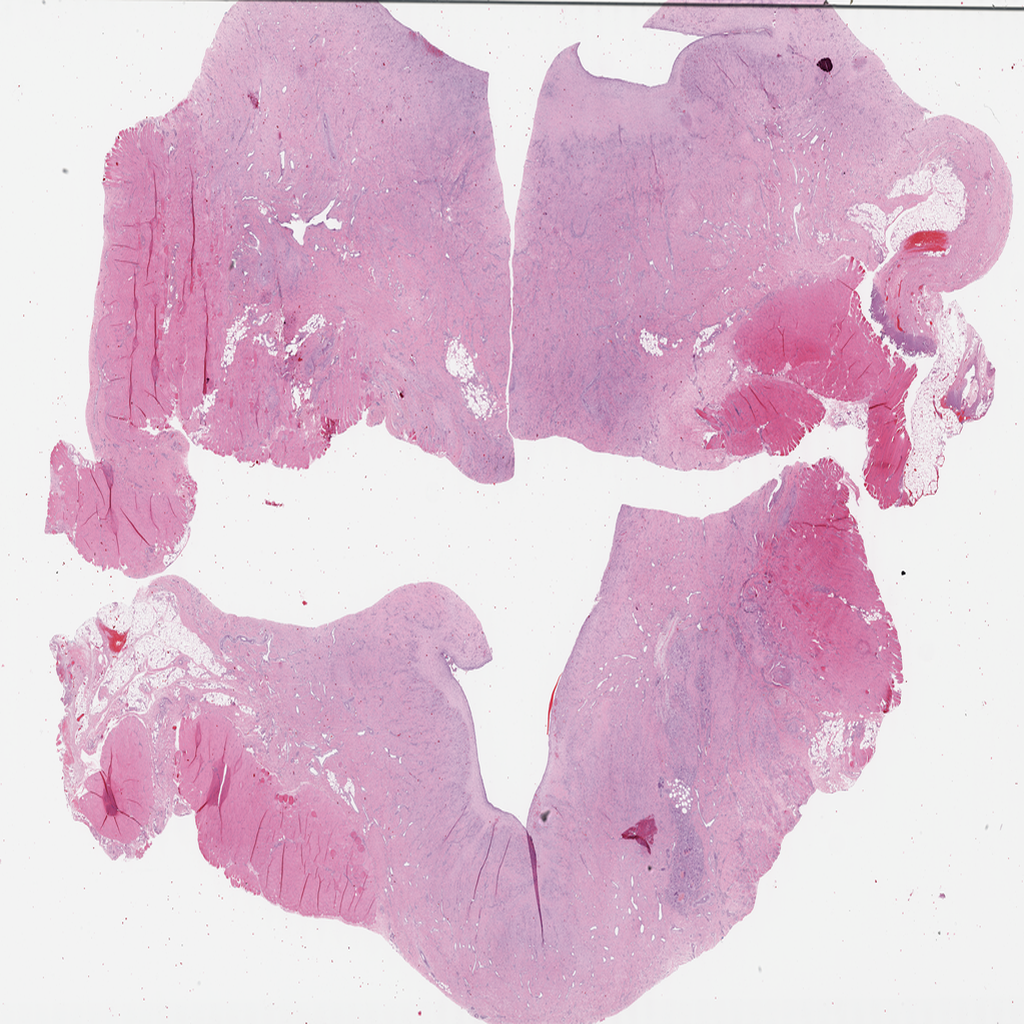} \\ 
\textbf{Question} & Can you provide an overall description of the whole slide image (WSI)? \\
\midrule
\textbf{WSI-LLaVA} &\textcolor{green!50!black}{ The whole slide image reveals exuberant granulation tissue characterized by notable fibroblastic proliferation and myxoid changes. The cellular component is primarily composed of spindle-shaped cells within a myxoid matrix. There is an absence of normal tissue architecture as the granulation tissue dominates the observed section. Importantly, no residual sarcoma tissue is identifiable within this examined section. }\\
\midrule
\textbf{Quilt-LLaVA} & \textcolor{red!80!black}{The whole slide image reveals a tumor with a diffuse architectural pattern, characterized by a mixture of alveolar formations and glandular cell differentiation. There is a notable presence of large cells with prominent nucleoli and moderate nuclear atypia. The mitotic activity is moderate, with 8 mitoses per 10 high power fields. The tumor shows significant vascular invasion, with tumor cells infiltrating the renal vein. Additionally, there is focal necrosis within the tumor tissue.} \\
\midrule
\textbf{WSI-VQA} & \textcolor{red!80!black}{the whole slide image reveals a biphasic tumor composed of spindle-shaped cells arranged in a storiform pattern .} these spindle cells are arranged in a storiform pattern . the spindle cells are arranged in a storiform pattern . the spindle cells are arranged in a storiform pattern . there is a notable absence of necrosis or calcification within the tumor. \\
\midrule
\textbf{GPT-4o} & The whole slide image appears to be a tissue section stained with hematoxylin and eosin (H\&E), commonly used in histopathology. The sample is divided into several sections, likely representing different orientations or parts of the tissue.

- **Coloration**: The pink and purple hues suggest typical H\&E staining, where eosin stains cytoplasmic components pink and hematoxylin stains nuclear material blue or purple.

- **Tissue Architecture**: The sections show varying thickness and include both smooth and irregular contours, indicating possible different tissue types or regions.

- **Other Observations**: There are areas that seem to contain more densely packed cells and some regions with more open or less cellular areas, suggesting different tissue densities or types.

A detailed examination at higher magnification would be necessary for further analysis, such as identifying specific pathological features or structures. \\
\midrule
\textbf{Ground Truth} & The whole slide image demonstrates exuberant granulation tissue characterized by notable fibroblastic proliferation and myxoid changes. The cellular component is primarily composed of spindle-shaped cells within a myxoid matrix. There is an absence of normal tissue architecture as the granulation tissue dominates the observed section. Importantly, no residual sarcoma tissue is identifiable within this examined section. \\ 
\bottomrule[1.5pt]
\end{tabular}}
\label{tab:sample_comparison_global_morphology_description}
\end{table}

\begin{table}[h!]
{
\scriptsize
\centering
\caption{A comparative example of Key Diagnostic Description outputs from different models}
\renewcommand{\arraystretch}{1.5}  
\begin{tabular}{>{\arraybackslash}m{1.3cm} m{6.4cm}}
\toprule[1.5pt]
\multicolumn{2}{l}{\textbf{Sample Comparison of Key Diagnostic Description}} \\
\midrule
\textbf{WSI} & \includegraphics[width=4.5cm]{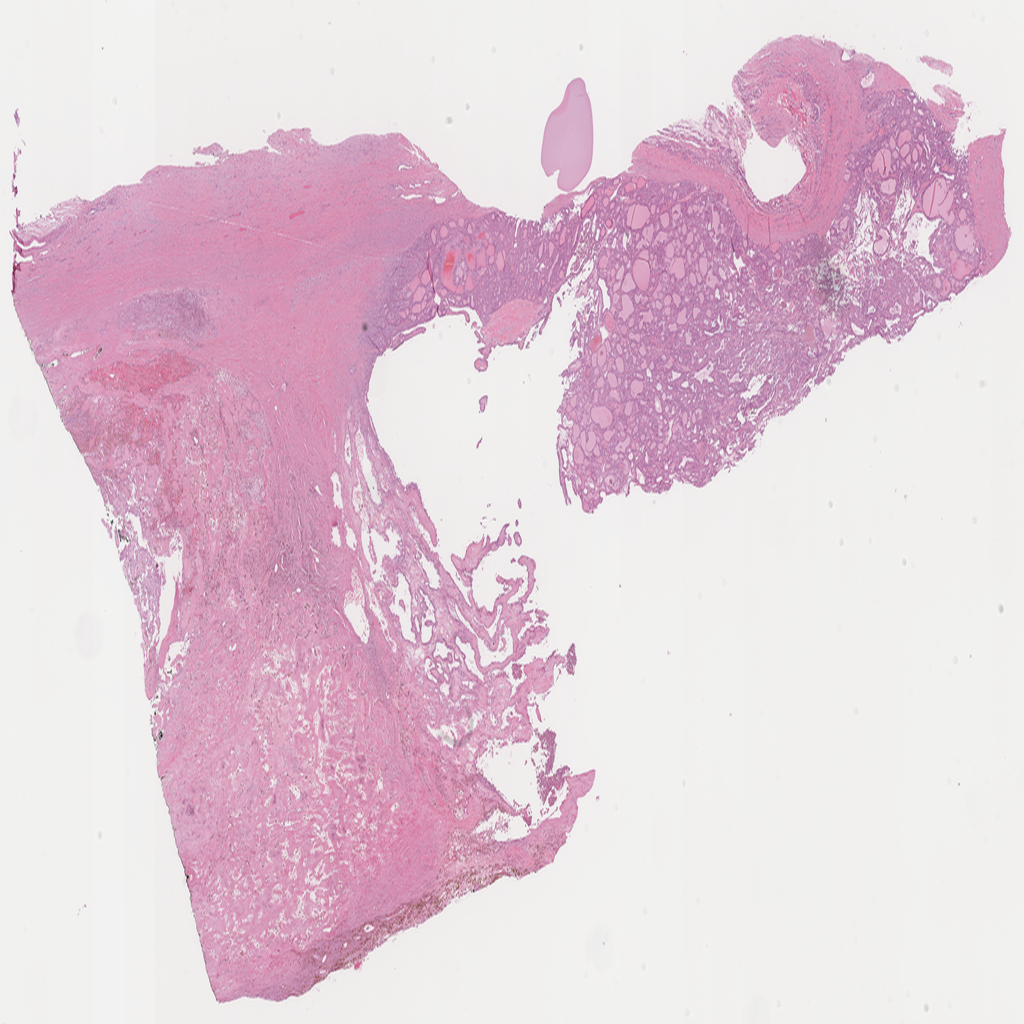} \\ 
\textbf{Question} & Which observed features are most significant for making a diagnosis? \\
\midrule
\textbf{WSI-LLaVA} & \textcolor{green!50!black}{The most significant features include the papillary architecture with fibrovascular cores, the presence of psammoma bodies, and the nuclear characteristics such as overlapping, grooves, and intranuclear cytoplasmic inclusions. The absence of vascular invasion, necrosis, and calcification also plays a crucial role in the diagnostic process.} \\
\midrule
\textbf{Quilt-LLaVA} & \textcolor{green!50!black}{The most significant features include the irregular clusters of epithelial cells} with poor glandular differentiation, \textcolor{green!50!black}{marked nuclear pleomorphism}, and the presence of perineural invasion. The high mitotic activity and evidence of \textcolor{green!50!black}{lymphovascular invasion }are also critical, as they indicate aggressive tumor behavior. \\
\midrule
\textbf{WSI-VQA} & \textcolor{green!50!black}{the most significant features include the papillary architecture with fibrovascular cores} and the presence of fibrovascular cores . \textcolor{green!50!black}{the absence of vascular invasion and necrosis further supports the diagnosis . }the absence of necrosis and calcification further supports the diagnosis . \\
\midrule
\textbf{GPT-4o} & To make a diagnosis from this histopathological image, the following observed features are significant:

1. **Cellular Architecture and Arrangement**: Observe the overall structure and any abnormal formations or patterns within the tissue.

2. **Presence of Lesions or Masses**: Look for areas indicating hyperplasia, neoplasia, or other abnormal growths.

3. **Inflammation Indicators**: Note any signs of infiltration by immune cells, such as lymphocytes or macrophages.

4. **Vascular Changes**: Identify any alterations in blood vessels, such as congestion, hemorrhage, or new vessel formation.

5. **Stromal Changes**: Assess changes in the supporting tissue, like fibrosis or stromal desmoplasia.

6. **Glandular Structures or Cyst Formation**: If applicable, evaluate the formation of glands or cystic spaces.

A detailed examination considering these features, along with clinical correlation, aids in reaching an accurate diagnosis. \\
\midrule
\textbf{Ground Truth} & The key features of high diagnostic importance include the branching papillary formations and the unique nuclear features of the neoplastic epithelial cells. The presence of crowding, overlapping nuclei with prominent grooves, and occasional intranuclear inclusions are particularly crucial. The absence of vascular invasion, necrosis, or calcification also provides essential context in the diagnostic evaluation. \\ 
\bottomrule[1.5pt]  
\end{tabular}}
\label{tab:sample_comparison_key_diagnostic_description}
\end{table}

\begin{table}[h!]
{
\scriptsize
\centering
\caption{A comparative example of Regional Structure Description outputs from different models}
\renewcommand{\arraystretch}{1.5}  
\begin{tabular}{>{\arraybackslash}m{1.3cm} m{6.4cm}}
\toprule[1.5pt]
\multicolumn{2}{l}{\textbf{Sample Comparison of Regional Structure Description }} \\
\midrule
\textbf{WSI} & \includegraphics[width=4cm]{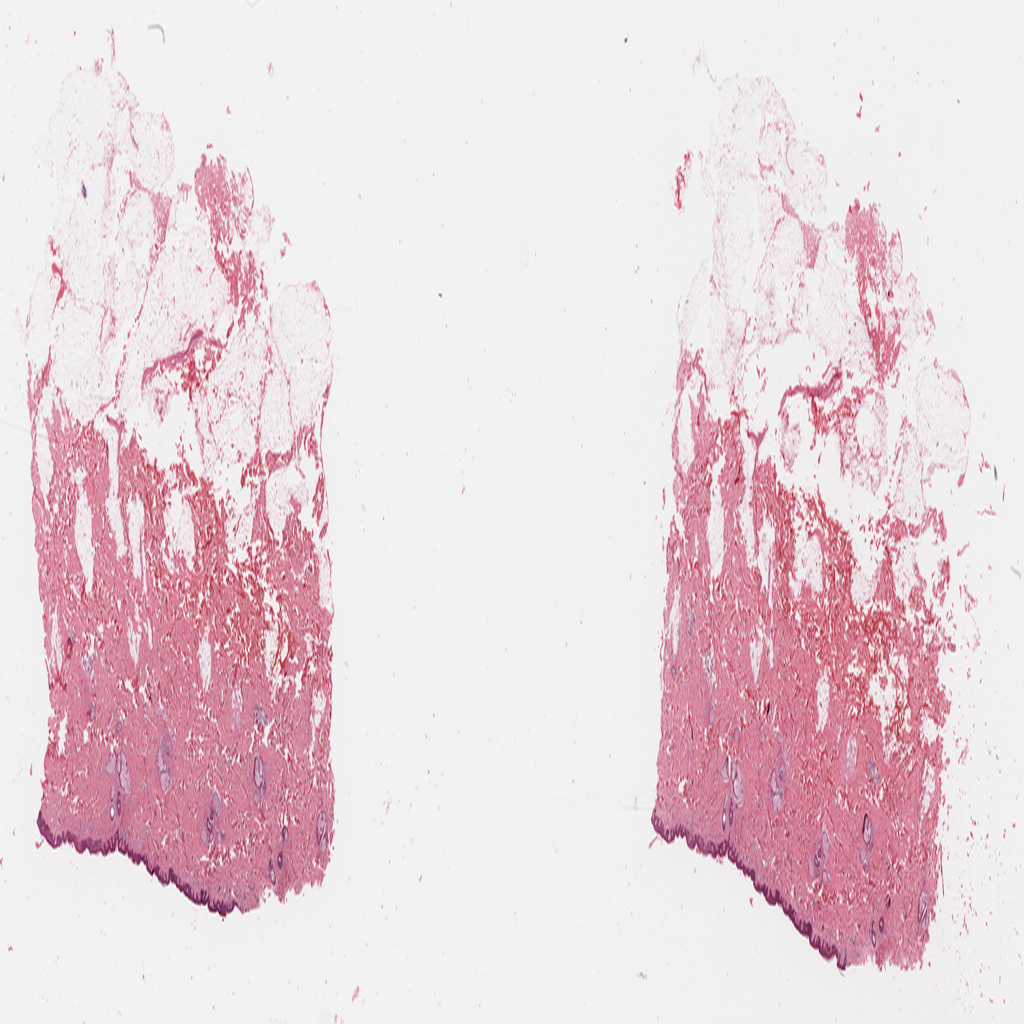} \\ 
\textbf{Question} & What observations can you make about the tissue architecture on this slide? \\
\midrule
\textbf{WSI-LLaVA} &\textcolor{green!50!black}{ The tissue architecture is characterized by a sheet-like growth pattern, where the tumor cells are densely packed, exhibiting minimal stromal separation. This creates an overall cohesive mass with sparse connective tissue presence.} \\
\midrule
\textbf{Quilt-LLaVA} & \textcolor{red!80!black}{The tissue architecture is characterized by nests and sheets of atypical melanocytes. These cells are arranged in a disorganized pattern, with a lack of the normal architectural structure typically seen in healthy tissue. The presence of necrosis within the tumor further disrupts the tissue architecture.} \\
\midrule
\textbf{WSI-VQA} & \textcolor{red!80!black}{the tissue architecture is characterized by papillary structures lined by neoplastic epithelial cells . these papillae are lined by layers of cuboidal to columnar epithelial cells . the papillae are lined by a single layer of cuboidal to columnar epithelial cells} \\
\midrule
\textbf{GPT-4o} & The tissue slide shows several features:

1. **Layered Structure**: The tissue exhibits a layered architecture indicative of an organ with distinct layers, such as the gastrointestinal tract.

2. **Heterogeneity**: There are regions with varied staining intensity, suggesting variability in cellular composition or density.

3. **Crypt Structures**: There are glandular or crypt-like structures present, typical of mucosal tissues like the colon.

4. **Villi or Projections**: Some areas show protrusions or folds, which could be part of the mucosal surface architecture.

5. **Potential Inflammation**: If there is any increased cellularity in certain areas, it could indicate inflammation or infiltration.

\textcolor{green!50!black}{6. **Connective Tissue**: Lighter stained areas may suggest the presence of connective or submucosal tissue.}

These observations can help identify the type of tissue and any pathological changes if present. \\
\midrule
\textbf{Ground Truth} & The tissue architecture is characterized by a sheet-like growth pattern, where the tumor cells are densely packed, exhibiting minimal stromal separation. This creates an overall cohesive mass with sparse connective tissue presence. \\ 
\bottomrule[1.5pt]  
\end{tabular}}
\label{tab:sample_comparison_regional_structure_description}
\end{table}

\begin{table}[h!]
{
\scriptsize
\centering
\caption{A comparative example of Specific Feature Description outputs from different models}
\renewcommand{\arraystretch}{1.5}  
\begin{tabular}{>{\arraybackslash}m{1.3cm} m{6.4cm}}
\toprule[1.5pt]
\multicolumn{2}{l}{\textbf{Sample Comparison of Specific Feature Description }} \\
\midrule
\textbf{WSI} & \includegraphics[width=4.7cm]{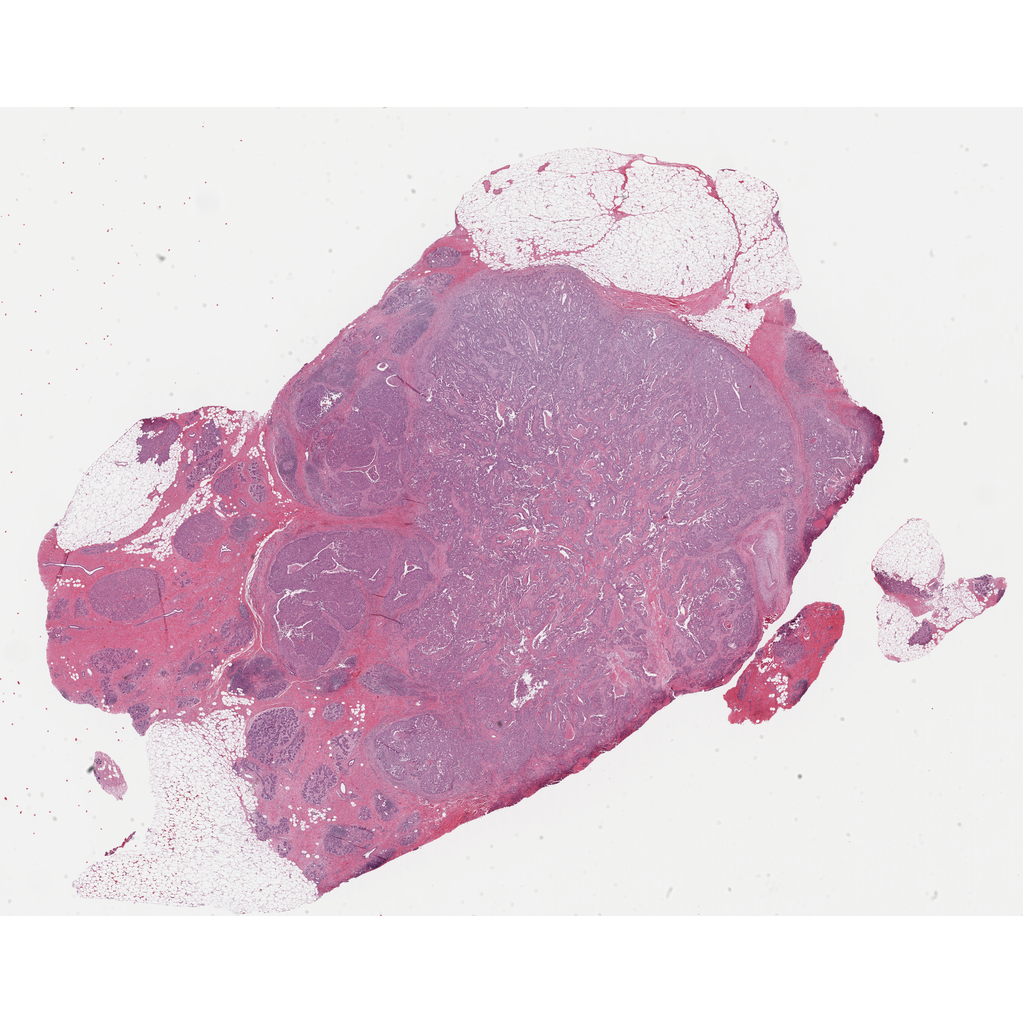} \\ 
\textbf{Question} & Do the epithelial cells infiltrate the surrounding stromal tissue or exhibit invasiveness? \\
\midrule
\textbf{WSI-LLaVA} &\textcolor{green!50!black}{ Yes, the epithelial cells are infiltrating the surrounding stromal tissue, demonstrating invasive characteristics.} They have breached their normal boundaries and are penetrating into the stroma, disrupting the typical tissue architecture. This invasion is accompanied by a desmoplastic stromal response, indicating the body's reaction to the invading tumor cells. \\
\midrule
\textbf{Quilt-LLaVA} & \textcolor{green!50!black}{Yes, the epithelial cells are infiltrating the surrounding stromal tissue, demonstrating invasive characteristics. }They have breached their normal boundaries and are penetrating into the stroma, disrupting the typical tissue architecture. This invasion is accompanied by minimal glandular formation, indicating a significant deviation from normal tissue organization. \\
\midrule
\textbf{WSI-VQA} & \textcolor{green!50!black}{yes the epithelial cells are infiltrating the surrounding stromal tissue demonstrating invasive characteristics .} they have breached their usual boundaries and are penetrating into the stroma disrupting the normal tissue architecture . this infiltration is a key feature of the epithelial cells . \\
\midrule
\textbf{GPT-4o} &\textcolor{red!80!black}{ The image shows a histological section. However, at this magnification, it is difficult to determine if epithelial cells are infiltrating the surrounding stromal tissue or exhibiting invasiveness. }To make an accurate assessment, a higher magnification would be necessary to examine the interactions between epithelial cells and stroma more closely. If invasiveness is suspected, it is best evaluated through additional histological analysis with higher resolution images. \\
\midrule
\textbf{Ground Truth} & Yes, the epithelial cells are infiltrating the surrounding stromal tissue. They extend beyond their usual confines, permeating into the stroma and demonstrating distinct invasive characteristics. This infiltration disrupts the normal architecture of the surrounding tissue. \\ 
\bottomrule[1.5pt]  
\end{tabular}}
\label{tab:sample_comparison_specific_feature_description}
\end{table}

\begin{table}[h!]
{
\scriptsize
\centering
\caption{A comparative example of Histological Typing outputs from different models}
\renewcommand{\arraystretch}{1.5} 
\begin{tabular}{>{\arraybackslash}m{1.3cm} m{6.4cm}}
\toprule[1.5pt]
\multicolumn{2}{l}{\textbf{Sample Comparison of Histological Typing }} \\
\midrule
\textbf{WSI} & \includegraphics[width=5cm]{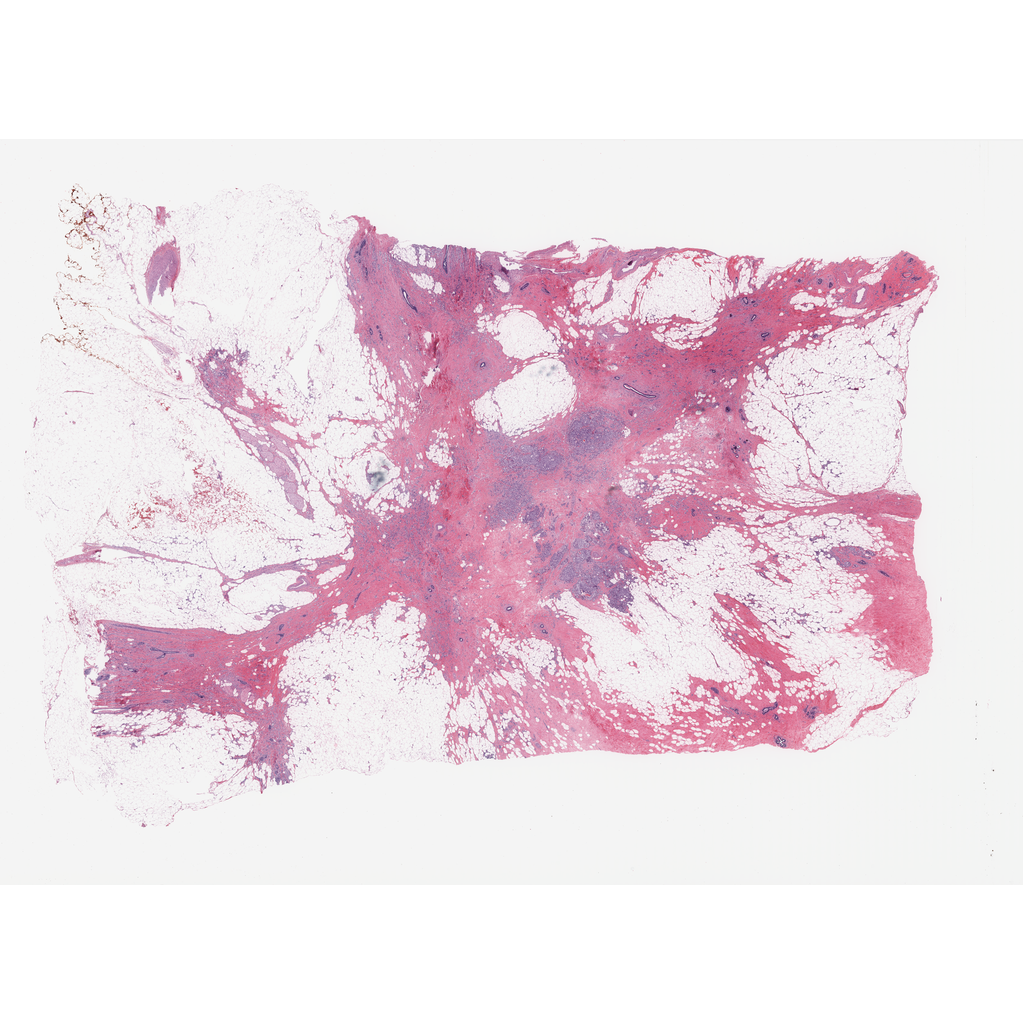} \\ 
\textbf{Question} & What is the histological classification based on your examination of the slide? \\
\midrule
\textbf{WSI-LLaVA} & \textcolor{green!50!black}{The histological classification is invasive ductal carcinoma.} \textcolor{green!50!black}{This is supported by the presence of infiltrating epithelial cells with minimal glandular formation, moderate nuclear pleomorphism,} and \textcolor{red!80!black}{ a moderate mitotic rate.} \textcolor{green!50!black}{The absence of vascular invasion and necrosis} also aligns with this classification. \\
\midrule
\textbf{Quilt-LLaVA} &\textcolor{green!50!black}{ The histological classification is invasive ductal carcinoma.} \textcolor{red!80!black}{This is supported by the presence of infiltrating epithelial cells forming moderately differentiated glandular structures, moderate nuclear pleomorphism, and the presence of ductal carcinoma in situ (DCIS) with solid and cribriform patterns.} \\
\midrule
\textbf{WSI-VQA} & \textcolor{green!50!black}{the histological classification based on the observed features is invasive ductal carcinoma . this classification is supported by the presence of infiltrating epithelial cells with minimal glandular formation moderate nuclear pleomorphism} and the presence of ductal carcinoma in situ dcis with cribriform and solid patterns . \\
\midrule
\textbf{GPT-4o} & \textcolor{red!80!black}{I'm unable to make a diagnosis from the image. However, the slide appears to show fibrous and necrotic areas, characteristic of particular tumor types or other pathological conditions. }For an accurate classification, a detailed analysis by a pathologist is required, considering factors such as cellular morphology, staining patterns, and clinical context. \\
\midrule
\textbf{Ground Truth} & The histological findings are consistent with invasive ductal carcinoma. This classification is supported by the infiltrative growth pattern of the epithelial cells, minimal glandular differentiation, moderate nuclear pleomorphism, and low mitotic activity. The absence of vascular and lymphatic invasion, along with the lack of necrosis and calcifications, helps to further characterize this subtype based on its histopathological features. \\ 
\bottomrule[1.5pt] 
\end{tabular}}
\label{tab:sample_comparison_histological_typing}
\end{table}

\begin{table}[h!]
{
\scriptsize
\centering
\caption{A comparative example of Grading outputs from different models}
\renewcommand{\arraystretch}{1.5}  
\begin{tabular}{>{\arraybackslash}m{1.3cm} m{6.4cm}}
\toprule[1.5pt]
\multicolumn{2}{l}{\textbf{Sample Comparison of Grading}} \\
\midrule
\textbf{WSI} & \includegraphics[width=4cm]{sample/TCGA-BH-A0DE-01Z-00-DX1.64A0340A-8146-48E8-AAF7-4035988B9152.svs_original.png} \\ 
\textbf{Question} & What is the Nottingham Histologic Grade for this carcinoma, and what does this grading reflect? \\
\midrule
\textbf{WSI-LLaVA} & The carcinoma is classified as \textcolor{green!50!black}{Nottingham Histologic Grade 2, with a total score of 6 out of 9}. This grading reflects \textcolor{green!50!black}{a tubular formation score of 3, indicating poor differentiation},\textcolor{green!50!black}{ a nuclear pleomorphism score of 2, representing moderate variability in nuclear size and shape}, and \textcolor{green!50!black}{a mitotic count score of 1, denoting a low rate of cell division}. \textcolor{green!50!black}{This intermediate grade suggests a moderately differentiated tumor, which has implications for prognosis and treatment planning}. \\
\midrule
\textbf{Quilt-LLaVA} & The carcinoma is classified as \textcolor{green!50!black}{Nottingham Histologic Grade 2, with a total score of 6 out of 9}. This grading reflects \textcolor{green!50!black}{a tubular formation score of 3, indicating poor differentiation, a nuclear pleomorphism score of 2, showing moderate variability,} and \textcolor{green!50!black}{a mitotic count score of 1, indicating low mitotic activity. This intermediate grade suggests a moderately differentiated tumor, which has implications for prognosis and treatment planning. }\\
\midrule
\textbf{WSI-VQA} & this carcinoma is classified as \textcolor{green!50!black}{nottingham histologic grade 2} with a total \textcolor{red!80!black}{score of 7 out of 9 }. \textcolor{green!50!black}{this grading reflects a tubular differentiation score of 3 a nuclear pleomorphism score of 2 and a mitotic count score of 1 .} \textcolor{red!80!black}{this grade reflects a moderately differentiated tumor with a total score of 1 }. \\
\midrule
\textbf{GPT-4o} & \textcolor{red!80!black}{I can't determine the Nottingham Histologic Grade from the image alone, as it requires detailed examination of specific histologic features.} The Nottingham grading system evaluates three components: 

1. **Tubule formation**: The percentage of tumor forming tubular structures.

2. **Nuclear pleomorphism**: Variation in the size and shape of the tumor cell nuclei.

3. **Mitotic count**: The number of cells dividing.
Each component is scored from 1 to 3, and the scores are summed to give a grade:

- **Grade 1 (3-5 points)**: Well-differentiated (low grade)

- **Grade 2 (6-7 points)**: Moderately differentiated (intermediate grade)

- **Grade 3 (8-9 points)**: Poorly differentiated (high grade).

This grading reflects the aggressiveness of the cancer, with higher grades indicating more aggressive tumors. \\
\midrule
\textbf{Ground Truth} & The carcinoma is classified as Nottingham Histologic Grade 2 (NHG2) with a total score of 6 out of 9. This grading reflects poor glandular (tubular) formation, moderateness in nuclear pleomorphism, and a low rate of mitotic activity. An NHG2 designation indicates a moderately differentiated tumor, suggesting an intermediate level of aggressiveness and guiding treatment choices accordingly. \\ 
\bottomrule[1.5pt]  
\end{tabular}}
\label{tab:sample_comparison_grading}
\end{table}

\begin{table}[h!]
{
\scriptsize
\centering
\caption{A comparative example of Molecular Subtyping outputs from different models}
\renewcommand{\arraystretch}{1.5} 
\begin{tabular}{>{\arraybackslash}m{1.3cm} m{6.4cm}}
\toprule[1.5pt]
\multicolumn{2}{l}{\textbf{Sample Comparison of Molecular Subtyping}} \\
\midrule
\textbf{WSI} & \includegraphics[width=5cm]{sample/TCGA-BA-A6DE-01Z-00-DX1.3279D372-4A92-4CFA-9E52-19197934378D.svs_original.png} \\ 
\textbf{Question} & Based on the IHC results showing P16 negativity and HR, HPV ISH status negativity, what is the molecular subtype in this head and neck squamous cell carcinoma? \\
\midrule
\textbf{WSI-LLaVA} & The molecular subtype is\textcolor{green!50!black}{ HNSC\_HPV}\-. \\
\midrule
\textbf{Quilt-LLaVA} & The molecular subtype is \textcolor{green!50!black}{HNSC\_HPV}\-. \\
\midrule
\textbf{WSI-VQA} & the molecular subtype is \textcolor{red!80!black}{hnsc unk} . \\
\midrule
\textbf{GPT-4o} & \textcolor{red!80!black}{The molecular subtype of head and neck squamous cell carcinoma with P16 negativity and HR HPV ISH negativity is typically HPV-negative. This subtype is generally associated with non-viral risk factors such as smoking and alcohol use.} \\
\midrule
\textbf{Ground Truth} & The molecular subtype is HNSC\_HPV\- \\ 
\bottomrule[1.5pt] 
\end{tabular}}
\label{tab:sample_comparison_molecular_subtyping}
\end{table}

\begin{table}[h!]
{
\scriptsize
\centering
\caption{A comparative example of Staging outputs from different models}
\renewcommand{\arraystretch}{1.5} 
\begin{tabular}{>{\arraybackslash}m{1.3cm} m{6.4cm}}
\toprule[1.5pt]
\multicolumn{2}{l}{\textbf{Sample Comparison of Staging}} \\
\midrule
\textbf{WSI} & \includegraphics[width=4cm]{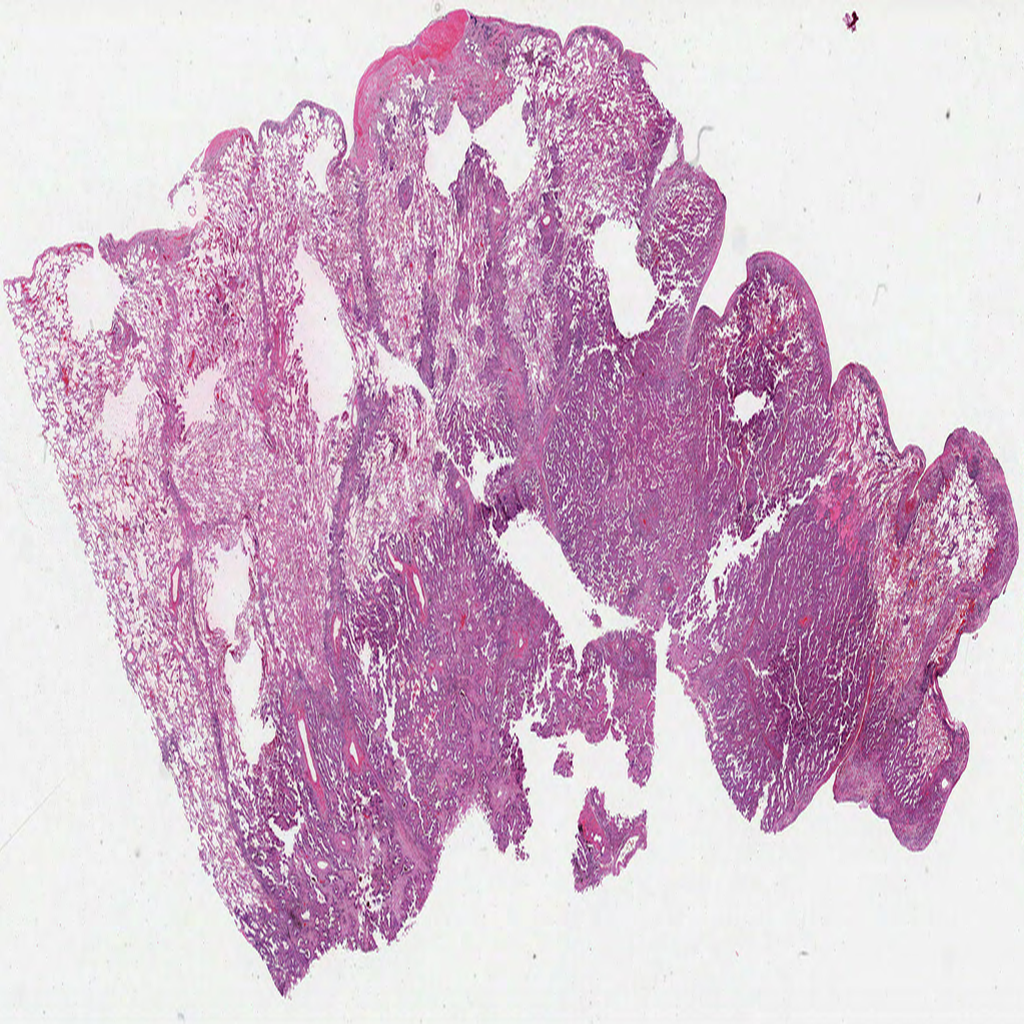} \\ 
\textbf{Question} & What is the TNM staging for a patient with a 7 cm tumor, several affected lymph nodes, but no distant metastasis? \\
\midrule
\textbf{WSI-LLaVA} & The TNM staging for this patient is \textcolor{green!50!black}{T3 N2 M0} \\
\midrule
\textbf{Quilt-LLaVA} & The TNM staging for this patient is \textcolor{green!50!black}{T3 N2 M0.} \\
\midrule
\textbf{WSI-VQA} & the tnm staging for this patient is \textcolor{red!80!black}{ t2 n2 m0 . }\\
\midrule
\textbf{GPT-4o} & To determine the TNM staging, we need to consider the following components:

- **T (Tumor size):** The tumor is 7 cm. This generally corresponds to T2 or T3, depending on the specific cancer type, but often T3 for many cancers.

- **N (Nodes):** \"Several affected lymph nodes\" typically indicates N1 or N2, depending on the exact number and extent of nodal involvement.

- **M (Metastasis):** No distant metastasis is indicated as M0.

Without specific details about the type of cancer, a general staging based on the information given could be \textcolor{red!80!black}{T3N1M0} or \textcolor{green!50!black}{T3N2M0}. However, the exact stage can vary based on cancer type and guidelines, so consultation with specific cancer staging resources is advised for precise staging. \\
\midrule
\textbf{Ground Truth} & TNM staging for this patient is T3 N2 M0. \\ 
\bottomrule[1.5pt] 
\end{tabular}}
\label{tab:sample_comparison_staging}
\end{table}

\begin{table*}[h!]
\scriptsize
\centering
\caption{A comparative example of Treatment Recommendations outputs from different models}
\begin{tabular}{>{\raggedright\arraybackslash}m{0.2\linewidth} >{\raggedright\arraybackslash}m{0.7\linewidth}}
\toprule[1.5pt]  
\multicolumn{2}{l}{\textbf{Sample Comparison of Treatment Recommendations}} \\
\midrule
\textbf{WSI} & \includegraphics[width=0.4\linewidth]{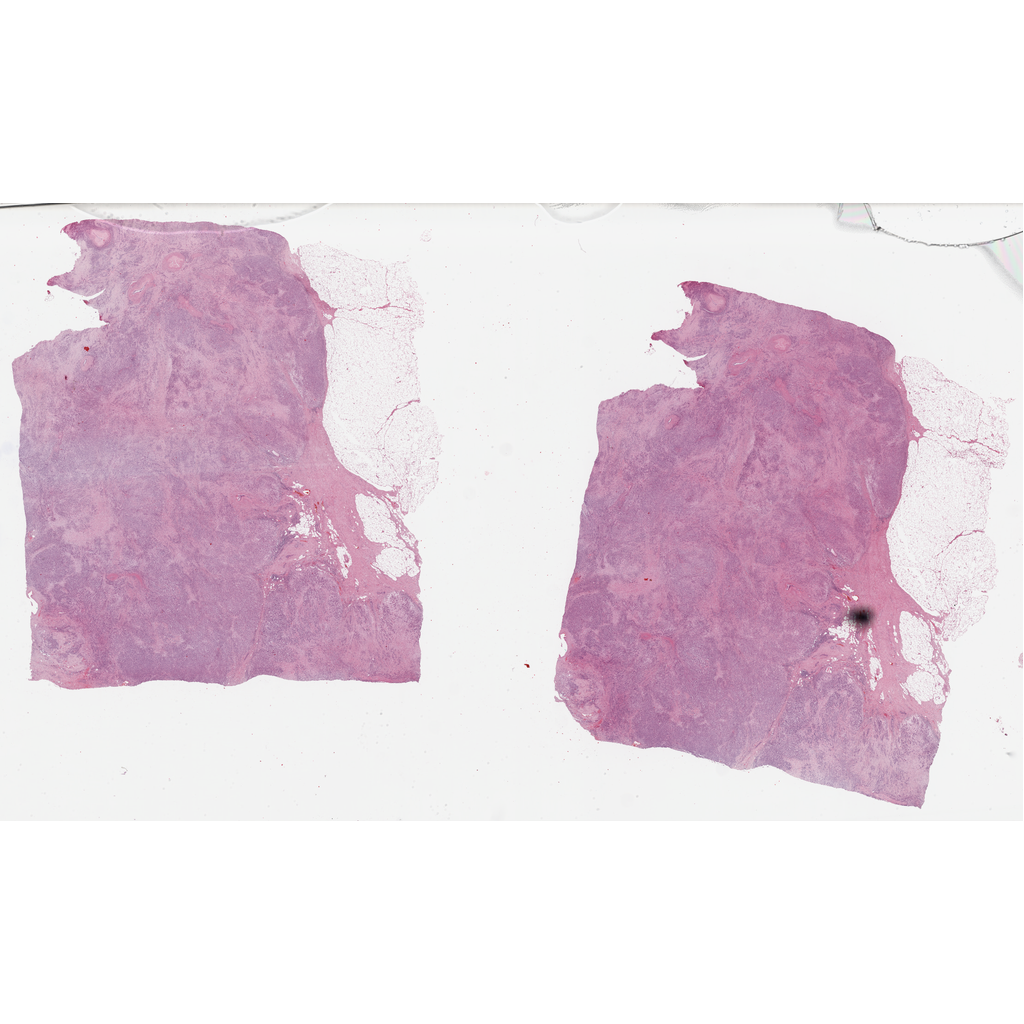} \\ 
\textbf{Question} & What is the recommended treatment plan for a patient with a diagnosis of invasive ductal carcinoma, Nottingham Histologic Grade 3, TNM staging T3N1, ER-positive, PR-positive, and HER2-negative? \\
\midrule
\textbf{WSI-LLaVA} &  1. \textcolor{green!50!black}{Surgery}: Consider mastectomy or breast-conserving surgery with axillary lymph node dissection due to T3 tumor size and N1 nodal involvement.

2. \textcolor{green!50!black}{Chemotherapy}: Adjuvant chemotherapy is recommended given the high histologic grade and nodal involvement to reduce recurrence risk.

3. \textcolor{green!50!black}{Endocrine Therapy}: Initiate tamoxifen or aromatase inhibitors for at least 5 years due to ER/PR positivity.

4. \textcolor{green!50!black}{Radiation Therapy}: Post-surgical radiation therapy is advised, especially if breast conservation is performed or in the presence of significant nodal disease. \\
\midrule
\textbf{Quilt-LLaVA} & 1. \textcolor{green!50!black}{ Surgery}: Consider mastectomy or breast-conserving surgery with axillary lymph node dissection due to T3N1 staging.

2. \textcolor{green!50!black}{ Chemotherapy}: Adjuvant chemotherapy is recommended given the high grade and nodal involvement.

3. \textcolor{green!50!black}{Endocrine Therapy}: Initiate tamoxifen or aromatase inhibitors for at least 5 years due to ER/PR positivity.

4. \textcolor{green!50!black}{Radiation Therapy}: Adjuvant radiation therapy is advised, especially if breast-conserving surgery was performed or due to significant nodal involvement. \\
\midrule
\textbf{WSI-VQA} & 1 . \textcolor{green!50!black}{surgery} mastectomy or breast-conserving surgery lumpectomy or mastectomy based on patient preference followed by sentinel lymph node biopsy or axillary lymph node dissection due to nodal involvement . 

2 . \textcolor{green!50!black}{endocrine therapy} initiate tamoxifen or aromatase inhibitors for at least 5 years due to er/pr positivity . 

3 . \textcolor{green!50!black}{radiation therapy} consider adjuvant radiation therapy especially if breast-conserving surgery was performed to reduce local recurrence risk . \\
\midrule
\textbf{GPT-4o} & For a patient with invasive ductal carcinoma, Nottingham Histologic Grade 3, TNM staging T3N1, ER-positive, PR-positive, and HER2-negative, the recommended treatment plan typically includes:

1. \textcolor{green!50!black}{Surgery}:Lumpectomy or mastectomy depending on the tumor size and patient preference.Axillary lymph node dissection or sentinel lymph node biopsy to address nodal involvement. 

2.\textcolor{green!50!black}{Adjuvant Chemotherapy}:Given the tumor size (T3) and histologic grade (Grade 3), chemotherapy is often recommended.A regimen such as AC-T (doxorubicin and cyclophosphamide followed by paclitaxel) might be used.  

3.\textcolor{green!50!black}{Radiation Therapy}:Typically recommended after surgery, especially if breast-conserving surgery (lumpectomy) is performed.It may also be considered after mastectomy if the tumor is large or if lymph nodes are positive.  

4.\textcolor{green!50!black}{Endocrine Therapy}:Since the tumor is ER-positive and PR-positive, endocrine therapy is crucial.Options include tamoxifen or aromatase inhibitors, depending on menopausal status.  

5.Follow-up and Monitoring:Regular follow-up visits for physical exams, mammograms (if breast tissue remains), and monitoring for any recurrence or metastasis.  \\
\midrule
\textbf{Ground Truth} & 1. Surgery: Perform a mastectomy or breast-conserving surgery with axillary lymph node dissection based on tumor size and nodal involvement. 

2. Chemotherapy: Adjuvant chemotherapy is recommended given the high tumor grade and nodal involvement. 

3. Endocrine Therapy: Initiate tamoxifen or aromatase inhibitors for at least 5 years due to ER and PR positivity.

4. Radiation Therapy: Consider post-surgical radiation therapy, especially if breast-conserving surgery was chosen or if significant nodal involvement is present. \\ 

\bottomrule[1.5pt]
\end{tabular}
\label{tab:sample_comparison_treatment_recommendations}
\end{table*} 

\begin{table}[h!]
{
\scriptsize
\centering
\caption{A comparative example of Prognosis outputs from different models}
\renewcommand{\arraystretch}{1.5}  
\begin{tabular}{>{\arraybackslash}m{1.3cm} m{6.4cm}}
\toprule[1.5pt]
\multicolumn{2}{l}{\textbf{Sample Comparison of Prognosis}} \\
\midrule
\textbf{WSI} & \includegraphics[width=4.5cm]{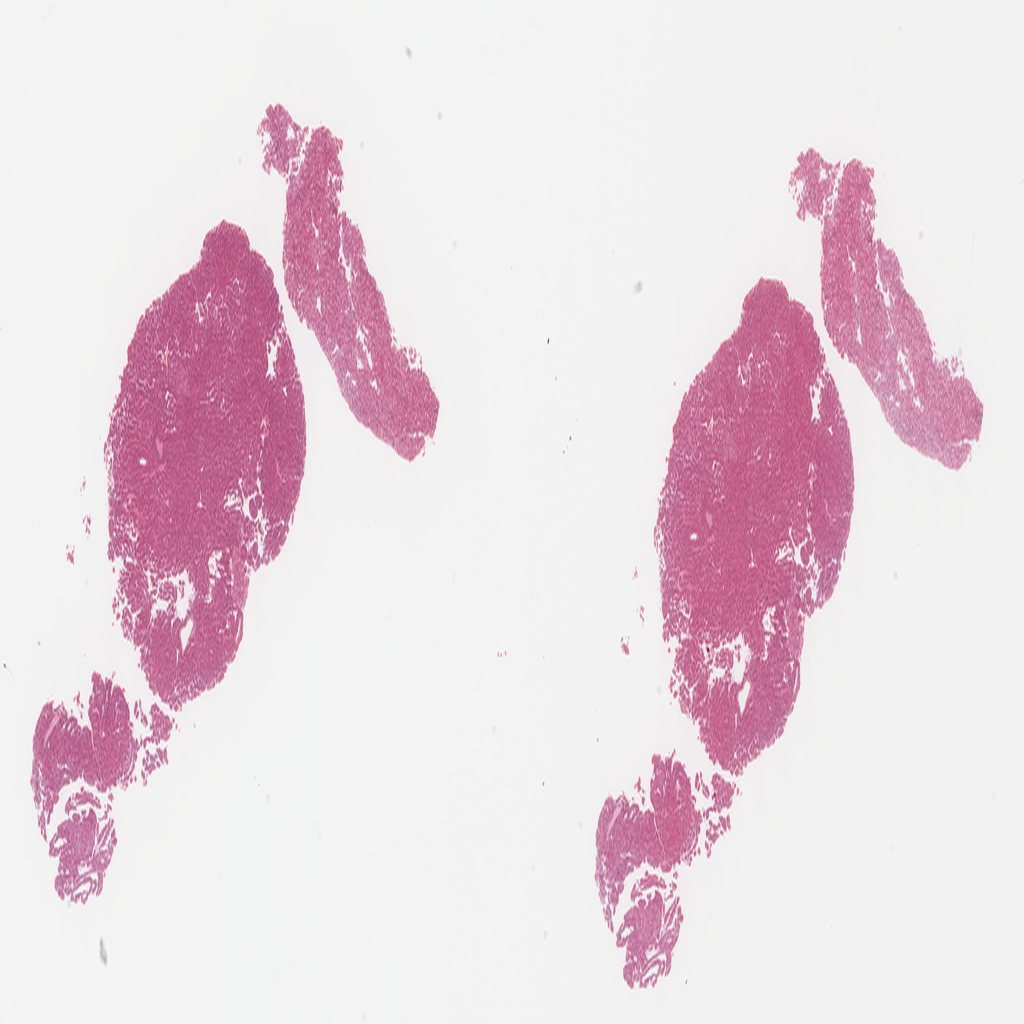} \\ 
\textbf{Question} & Given the patient's diagnosis of hepatocellular carcinoma, TNM staging of T1N0M0, and immunohistochemistry results showing positive Hepatocyte and negative CK19, is the prognosis good or poor? \\
\midrule
\textbf{WSI-LLaVA} & \textcolor{green!50!black}{The prognosis is generally good.} The early-stage (T1N0M0) hepatocellular carcinoma with positive Hepatocyte and negative CK19 indicates a well-differentiated tumor, which typically has a better prognosis compared to more advanced stages. \\
\midrule
\textbf{Quilt-LLaVA} & \textcolor{green!50!black}{The prognosis is good. }The early stage (T1N0M0) and absence of lymph node involvement (N0) suggest a less aggressive disease, and the positive Hepatocyte marker indicates the tumor is likely of hepatocellular origin, which is typically associated with a better outcome \\
\midrule
\textbf{WSI-VQA} & \textcolor{red!80!black}{B} \\
\midrule
\textbf{GPT-4o} & In the case of a T1N0M0 stage hepatocellular carcinoma (HCC), the prognosis is generally considered relatively good. The TNM staging indicates:
- **T1**: Single tumor without vascular invasion.
- **N0**: No regional lymph node involvement.
- **M0**: No distant metastasis.
This early-stage diagnosis suggests a higher likelihood of successful treatment options, potentially including surgery, and a better overall prognosis compared to more advanced stages.
The immunohistochemistry results showing positive Hepatocyte and negative CK19 support the diagnosis of a well-differentiated HCC, which is generally associated with a better prognosis. CK19 negativity implies the tumor is less likely to be aggressive compared to CK19-positive HCCs.
Overall, \textcolor{green!50!black}{the combination of early stage and favorable immunohistochemistry results suggests a relatively good prognosis for this patient.} However, individual prognosis can vary, and continuous follow-up and management are essential. \\
\midrule
\textbf{Ground Truth} & The prognosis is good. The early stage of liver cancer (T1N0M0) with no lymph node involvement or metastasis suggests a limited disease burden, while the IHC profile indicates typical hepatocellular carcinoma without bile duct differentiation. \\ 
\bottomrule[1.5pt]
\end{tabular}
\label{tab:sample_comparison_prognosis}}
\end{table}

\end{document}